\documentclass[letterpaper]{article} 
\usepackage{aaai2026}  
\usepackage{times}  
\usepackage{helvet}  
\usepackage{courier}  
\usepackage[hyphens]{url}  
\usepackage{graphicx} 
\usepackage{natbib}  
\usepackage{caption} 
\usepackage{algorithm}
\usepackage{algorithmic}

\usepackage{newfloat}
\usepackage{listings}
\DeclareCaptionStyle{ruled}{labelfont=normalfont,labelsep=colon,strut=off} 
\floatstyle{ruled}
\newfloat{listing}{tb}{lst}{}
\floatname{listing}{Listing}
\title{A Weighted Loss Approach to Robust Federated Learning under Data Heterogeneity}
\author{
Johan Erbani\textsuperscript{\rm 1}\quad
Sonia Ben Mokhtar\textsuperscript{\rm 1}\quad
Pierre-Edouard Portier\textsuperscript{\rm 2}\quad
Elod Egyed-Zsigmond\textsuperscript{\rm 1}\quad
Diana Nurbakova\textsuperscript{\rm 1}
}
\affiliations{
\textsuperscript{\rm 1}INSA Lyon, CNRS, LIRIS, UMR 5205, 69621 Villeurbanne, France\\
\textsuperscript{\rm 2}Caisse d’Epargne Rhône-Alpes, Tour Incity, 116 Cours Lafayette, 69003 Lyon, France\\
}

\usepackage{bibentry}

\usepackage{array}  
\usepackage{amsmath}      
\usepackage{amssymb}      
\usepackage{amsthm}
\usepackage{subcaption}
\usepackage{mathtools}    
\usepackage{color}          
\usepackage{booktabs}
\usepackage{multirow}

\definecolor{color1}{rgb}{0.265, 0.664, 0.538}
\definecolor{color2}{rgb}{0.983, 0.357, 0.115}
\definecolor{color3}{rgb}{0.368, 0.473, 0.712}
\definecolor{color4}{rgb}{0.857, 0.301, 0.642}
\definecolor{color5}{rgb}{0.542, 0.766, 0.175}
\definecolor{color6}{rgb}{0.947, 0.774, 0.000}
\definecolor{color7}{rgb}{0.840, 0.637, 0.343}
\definecolor{color8}{rgb}{0.562, 0.562, 0.562}
\definecolor{gold}{RGB}{245, 185, 0}
\definecolor{silver}{RGB}{5, 145, 250}
\definecolor{bronze}{RGB}{230, 57, 80}

\begin{document}

\maketitle

\begin{abstract}
Federated Learning (FL) allows multiple data holders (workers) to collaboratively train a machine learning model without sharing their private data. A central server coordinates the process by distributing the global model and receiving local gradients from workers. However, faulty or malicious (Byzantine) workers can compromise training. In heterogeneous environments, even honest gradients may differ substantially, making it challenging to distinguish them from Byzantine updates. We propose Worker Label Alignment Loss (WoLA), a weighted loss function that promotes alignment among honest gradients, thereby enhancing the detection of Byzantine behavior. Our method demonstrates superior performance compared to state-of-the-art techniques in empirical evaluations and is supported by theoretical analysis.
\end{abstract}

\begin{links}
    \link{Code}{https://github.com/anonymous14031992/WoLA}
\end{links}

\section{Introduction}
Federated learning (FL)~\cite{guerraoui2024fundamentals, guerraoui2024byzantine, wen2023survey} enables multiple machines (workers) to collaboratively train a model without sharing private data. A central server coordinates the training by distributing the current global model to all workers. Each worker computes an update based on its local data and returns it to the server. The server aggregates these updates, updates the global model, and repeats the process. Typically, these updates correspond to gradients computed on local minibatches~\cite{allouah2023fixing}.

Under ideal conditions, the server aggregates gradients using simple averaging. However, some workers may fail or behave maliciously, sending corrupted (Byzantine) updates. Robust training aims to achieve performance comparable to that without Byzantine failures. A common approach replaces averaging with robust aggregation methods on the server~\cite{guerraoui2024fundamentals, guerraoui2024byzantine}. These methods mitigate the impact of outliers, which are expected to include Byzantine updates. Their effectiveness, however, depends heavily on honest workers having similar data distributions~\cite{guerraoui2024byzantine, allouah2023fixing, peng2024mean, bao2024boba}. In practice, this assumption often fails~\cite{allouah2023fixing, karimireddy2020byzantine, li2022federated, fang2025we}, as each worker typically accesses only a small, potentially unrepresentative subset of the data.

In heterogeneous settings, variance among honest gradients increases, enabling Byzantine workers to better conceal their malicious updates~\cite{allouah2023fixing, guerraoui2024fundamentals, karimireddy2020byzantine}. To enhance robustness, many methods introduce a pre-aggregation step~\cite{guerraoui2024fundamentals, guerraoui2024byzantine}, such as Bucketing~\cite{karimireddy2020byzantine} or Nearest Neighbor Mean (NNM)~\cite{allouah2023fixing}. While effective when the proportion of Byzantine updates is low, we show that their performance deteriorates significantly in stronger adversarial scenarios.

We propose a novel approach for robust FL in heterogeneous and highly adversarial environments. Specifically, we focus on label distribution skew~\cite{li2022federated, ran2021dynamic, zhang2022federated}—a common form of data heterogeneity where label distributions vary across workers, but feature distributions within each class remain consistent. For example, in a mobile image recognition app, users of different ages may capture different subjects, leading to variation in label distributions across devices~\cite{li2022federated, ran2021dynamic, zhang2022federated}. Label skew causes inconsistent training, as each worker optimizes the model according to its local label distribution~\cite{karimireddy2020byzantine, zhang2022federated, bao2024boba}, producing highly divergent updates.

Instead of reducing update variance via pre-aggregation, we tackle the root cause: misaligned worker objectives. We introduce the \textbf{Worker Label Alignment (WoLA)} loss, a weighted loss function that aligns local training objectives across workers. WoLA reweights point-wise errors to generate gradients as if all local datasets shared the same label distribution. By steering honest gradients toward a common direction, WoLA reduces variance and improves resilience against Byzantine attacks.

Our contributions are as follows:
\begin{itemize}
\item We introduce WoLA, a weighted loss function designed to address label skew in highly adversarial FL settings.
\item We provide theoretical analysis supporting WoLA’s effectiveness, with proofs in the appendix.
\item We empirically demonstrate that WoLA significantly outperforms state-of-the-art methods across multiple models, datasets, aggregators, and Byzantine attack scenarios.
\end{itemize}

\section{Background}\label{Background Related Work}
This section outlines key FL concepts and defines our threat model. All notations are introduced at first use and summarized in Appendix Table~\ref{tab:notations}.

\subsection{Byzantine-free FL}
We consider a distributed FL setup with $n$ workers and a central server, tasked with a classification problem involving $C$ classes. Each worker $i$ holds a local dataset $\mathcal{D}_i$ of size $N_i$, consisting of samples $(x,y) \in \mathcal{X} \times [C]$, where $\mathcal{X}$ is the input space, $x$ an observation, and $y$ its label. Given a model $\Phi$, the local loss function for worker $i$ is:
$$
\mathcal{L}_i = \frac{1}{N_i} \sum_{(x,y)\in\mathcal{D}_i} \ell\big(y, \Phi(x)\big),
$$
where $\ell$ is a point-wise loss, assumed differentiable with respect to model parameters.

In each communication round, the server broadcasts the current model parameters to all workers. Each worker $i$ computes the gradient $\nabla \mathcal{L}_i$ on its local data $\mathcal{D}_i$ and sends this update back to the server.

\subsection{Threat Model}
We assume a Byzantine setting with a trusted server and fewer than half of the workers being malicious ($f < n/2$), following the threat model in~\cite{allouah2023fixing, karimireddy2020byzantine, bao2024boba}. Byzantine workers can send arbitrary updates and have full knowledge of all honest updates.

The server’s goal is to find a stationary point of the average loss over honest workers, i.e.,
$$
\nabla \mathcal{L} = 0,\quad\text{with }\mathcal{L} := \sum_{i \in \mathcal{H}} \frac{N_i}{N} \mathcal{L}_i,\quad N = \sum_{i \in \mathcal{H}} N_i,
$$
where $\mathcal{H}$ denotes the set of honest workers. Assuming such a stationary point exists, it can be approximated via gradient descent combined with robustness mechanisms.

\subsection{Global Dataset, Local Datasets \& Data Heterogeneity} 
We focus on a label skew scenario:  all data points are drawn from independent and identically distributed (i.i.d) random variables, forming the global dataset $\mathcal{D}$. These points are randomly assigned to honest workers, producing local datasets $\mathcal{D}_i$ for $i \in \mathcal{H}$, such that $\mathcal{D} = \cup_{i \in \mathcal{H}} \mathcal{D}_i$. Each worker receives data with a different label distribution, introducing data heterogeneity. Although Byzantine workers may also have their own data, we exclude them from the global dataset as they play no role in our analysis. 

We denote the global label distribution by $p$, corresponding to $\mathcal{D}$, and the local label distribution for worker $i$ by $p_i$, corresponding to $\mathcal{D}_i$.

\subsection{Gradient Dissimilarity}
Data heterogeneity increases divergence among honest gradients~\cite{guerraoui2024fundamentals, guerraoui2024byzantine, karimireddy2020byzantine}. This divergence can be measured by Gradient Dissimilarity~\cite{karimireddy2020byzantine}:
\begin{equation*} 
\text{Gradient Dissimilarity} = \frac{1}{\#\mathcal{H}} \sum_{i \in \mathcal{H}} \| \nabla \mathcal{L}_i - \nabla \mathcal{L} \|^2, 
\end{equation*} 
where $\#\mathcal{H}$ is the number of honest workers. 

Gradient Dissimilarity is a proxy for system robustness. Lower values indicate higher robustness.

\section{Related Work}
This section reviews state-of-the-art approaches relevant to our context.

\subsection{Robustness Strategies with Homogeneous Data}
Robust aggregation methods mitigate the impact of Byzantine failures during training~\cite{pillutla2022robust, guerraoui2024fundamentals}. Examples include coordinate-wise median (CWMed)\cite{yin2018byzantine}, geometric median (GM)\cite{small1990survey, pillutla2022robust}, coordinate-wise trimmed mean (CwTM)\cite{yin2018byzantine}, Krum\cite{blanchard2017machine}, Multi-Krum (MKrum)\cite{guerraoui2018hidden}, Minimum Diameter Averaging\cite{guerraoui2018hidden}, and Comparative Gradient Elimination~\cite{gupta2021byzantine}. These methods strongly rely on the assumption that honest workers have homogeneous data. Their robustness under data heterogeneity remains unclear~\cite{guerraoui2024byzantine, allouah2023fixing, peng2024mean}.

Beyond robust aggregators, additional strategies enhance robustness in homogeneous settings by leveraging a clean dataset held by the server to evaluate and filter worker updates~\cite{xie2019zeno, cao2020fltrust, fang2020local, regatti2022byzantine}.

\subsection{Robustness Strategies Under Data Heterogeneity}
To reduce gradient variance caused by data heterogeneity, a common approach applies a pre-aggregation step before the main aggregation rule, leading to a two-step process. Examples include Bucketing~\cite{karimireddy2020byzantine}, Nearest-Neighbor Mixing (NNM)\cite{allouah2023fixing}, and FoundationFL\cite{fang2025we}.

Bucketing~\cite{karimireddy2020byzantine} randomly partitions inputs into groups (buckets), averages the inputs within each bucket, and then feeds these averages into the aggregation rule. This random partitioning reduces the variance of honest updates in expectation. However, in some iterations, it may fail to reduce heterogeneity, giving Byzantine workers greater opportunity to disrupt training~\cite{guerraoui2024byzantine}.

In contrast, NNM~\cite{allouah2023fixing} is a deterministic pre-aggregation method that replaces each gradient with the average of its $n - f$ nearest neighbors. NNM achieves optimal probabilistic robustness guarantees but has a high computational cost, limiting its scalability in large distributed systems~\cite{guerraoui2024byzantine}.

FoundationFL~\cite{fang2025we} synthesizes additional updates at each training step. The server scores each update based on its distance to the coordinate-wise extremes and replicates the highest-scoring update approximately $n/2$ times\footnote{Given updates $g_1, \dots, g_n$, define $g_{\text{max}} = \max(g_1, \dots, g_n)$ and $g_{\text{min}} = \min(g_1, \dots, g_n)$ (coordinate-wise). The score $s_i$ for worker $i$ is then $s_i = \min(\|g_{\text{max}} - g_i\|, \|g_{\text{min}} - g_i\|)$. A high score means the update is far from the extremes.}. This scoring assumes that extreme values are Byzantine but becomes unreliable when data is heterogeneous~\cite{baruch2019little}.

Other methods, such as BOBA~\cite{bao2024boba} and RAGE~\cite{data2021byzantine}, adopt different strategies without relying on pre-aggregation. However, they require access to a server-side dataset and are computationally expensive, limiting their practicality in large-scale optimization~\cite{karimireddy2020byzantine}\footnote{At each round, BOBA requires computing several times the truncated SVD of the gradient matrix, which scales with both the number of clients and the gradient dimension.}.

\subsection{Our proposition}
To the best of our knowledge, employing a loss function to improve robustness under data heterogeneity is novel. WoLA is model-agnostic, requires no hyperparameter tuning, no server-side dataset, no knowledge of the number of Byzantine workers, and adds no computational overhead. It is straightforward to implement and can be combined with other robustness strategies. WoLA only requires a common training objective shared by the server once before training begins.

While many approaches address update heterogeneity~\cite{allouah2023fixing, karimireddy2020byzantine, bao2024boba, fang2025we, data2021byzantine}, WoLA tackles its root cause—label skewness—by aligning gradients from the start.

\section{Worker Label Alignement Loss}\label{WoLA}
This section introduces WoLA. We first present the rationale behind its design, then provide its formal definition and training objective. The final subsections describe the theoretical framework and the properties of WoLA. Proofs and additional details are provided in the Appendix.

\subsection{Rationale for WoLA} 
We begin by observing that local gradients are weighted averages of class-wise gradients:
\begin{equation*}
    \nabla \mathcal{L}_i = \frac{1}{N_i} \sum_{(x,y)\in\mathcal{D}_i} \nabla\ell\Big(y, \Phi(x)\Big) = \sum_{c=1}^C p_i^c \mu^c_i,
\end{equation*}
where $p^c_i \vcentcolon= N^c_i/N_i$ is the proportion of class $c$ in $\mathcal{D}_i$, and
\begin{align*}
\mu_i^c \vcentcolon=
\frac{\sum_{(x,y)\in\mathcal{D}_i} \mathbf{1}(y=c) \nabla\ell\Big(c, \Phi(x)\Big)}{N^c_i}
\end{align*}
is the average gradient for class $c$ in worker $i$'s local dataset.

With sufficient samples per class, local class-wise gradients $\mu_i^c$ approximate the global class-wise gradients $\mu^c$:
\begin{equation}\label{approximation}
    \nabla \mathcal{L}_i =  \sum_{c=1}^C p_i^c \mu^c_i \approx \sum_{c=1}^C p_i^c \mu^c,
\end{equation}
where the global class-wise gradient is defined as:
\begin{align*}
    \mu_c\vcentcolon= \frac{\sum_{(x,y)\in\mathcal{D}} \mathbf{1}(y=c) \nabla\ell\Big(c, \Phi(x)\Big)}{N^c}
\end{align*}
In general, global class-wise gradients are not equal~\cite{zhang2022federated, bao2024boba}; we illustrate this in the Appendix.

Assuming the above approximation in \eqref{approximation} holds, differences in local gradients primarily stem from differences in local label distributions (via $p_i^c$). Hence, aligning local gradients requires aligning local label distributions or simulating such alignment. WoLA is specifically designed to achieve this goal.

\subsection{Training Objective \& WoLA Definition}\label{Training Objective}
WoLA requires a training objective $q = (q^1, q^2, \ldots, q^C)$ with $\sum_{i \in [C]} q^i = 1$ and $q^i \geq 0$, which is shared with all workers once before training begins. This vector $q$ represents the simulated local label distribution of each worker when using WoLA. In practice, a higher $q^i$ increases the weight given to errors on class $i$ during local training.

The distribution $q$ may approximate the true global label distribution or be set arbitrarily. We present three settings for defining $q$:

\subsubsection*{Local label distributions are shared}
Each worker communicates its local label distribution to the server, which computes and broadcasts the global label distribution. However, this introduces a potential vulnerability: Byzantine workers can manipulate the training objective by reporting false distributions. We analyze this attack in the Appendix (see Section \textit{Training Objective Attack}).

\subsubsection*{Prior knowledge} The global label distribution is known in advance, based on expert knowledge or historical data. For example, in tumor detection on MRI scans or crop type recognition from satellite imagery, national statistics can serve as a reliable proxy for label distributions.

\subsubsection*{Consensus among workers on the training objective} All workers may agree on a predefined target distribution, regardless of the true label distribution. For example, in fair hiring systems, classes may be weighted equally to mitigate algorithmic bias.

WoLA adjusts point-wise error contributions to simulate a common label distribution $q$ across all workers.
\paragraph{Definition 1}\begin{itshape} The local WoLA loss for worker $i$, denoted $\mathcal{W}\!\mathcal{L}_i$, is defined as:
\begin{equation*}
\mathcal{W}\!\mathcal{L}_i = \frac{1}{N_i} \sum_{(x,y)\in \mathcal{D}_i} \frac{q^y}{p^y_i} \ell\Big(y, \Phi(x)\Big),
\end{equation*}
\end{itshape}%
\noindent A sample from class $c$ is up-weighted when class $c$ is underrepresented in the local dataset compared to the training objective $q$, and down-weighted otherwise. This formula is directly derived from Importance Sampling~\cite{tokdar2010importance}; further details are provided in the Appendix.

Before establishing the theoretical properties that explain WoLA's efficiency, we introduce the theoretical framework.

\subsection{Theoretical Framework}\label{Data Modeling}
Let $(X_1, Y_1, Z_1), (X_2, Y_2, Z_2), \ldots, (X_N, Y_N, Z_N)$ be independent and identically distributed (i.i.d.) samples from a distribution $\mathfrak{D}$ over $\mathcal{X} \times [C] \times \mathcal{H}$. Each pair $(X_i, Y_i)$ represents a datapoint in $\mathcal{D}$. For each $k \in [N]$, the random variable $Z_k$ takes values in $\mathcal{H}$ and indicates the worker to which the $k$-th datapoint is assigned. Let $(X, Y, Z)$ be a generic triplet with the same distribution $\mathfrak{D}$.

For any triplet $(X_k, Y_k, Z_k)$, we assume $X_k$ and $Z_k$ are conditionally independent given $Y_k$, i.e., $P(X_k, Z_k | Y_k) = P(X_k | Y_k) P(Z_k | Y_k)$. Given the label, features provide no information about assignments, and vice versa. This implies $P(X_k| Y_k, Z_k) = P(X_l| Y_l, Z_l)$ for any $k, l$, so the feature distribution is identical across workers.

\subsection{WoLA Properties}\label{Theoretical Results}
Let $L_i$ and $W_i$ be the random counterparts of $\nabla \mathcal{L}_i$ and $\nabla \mathcal{W}\!\mathcal{L}_i$, respectively. Let $L$ and $W$ denote the average gradients of honest workers for the standard loss and WoLA, respectively.

Without WoLA, local gradients remain different even with an infinite amount of data.
\paragraph{Proposition 1}\begin{itshape}Under the standard loss, the gradient of worker $i$ converges almost surely as $N \to \infty$:  
$$
L_i\underset{N\rightarrow \infty}{\overset{\text{a.s.}}{\longrightarrow}} \mathbb{E}\Big[\nabla \ell(Y, \Phi(X))\Big|Z=i\Big]
$$
Moreover, the average gradient satisfies:
$$
L\underset{N\rightarrow \infty}{\overset{\text{a.s.}}{\longrightarrow}}  \mathbb{E}\Big[\nabla \ell(Y, \Phi(X))\Big]
$$
\end{itshape}

\noindent Under label skewness, and since class-wise gradients differ, there exists at least one worker $i$ for which the expected gradient differs from the global expectation: 
$$
\mathbb{E}[\nabla \ell(Y, \Phi(X))|Z=i] \neq \mathbb{E}[\nabla \ell(Y, \Phi(X))]
$$
This implies $\lim L_i \overset{\text{a.s.}}{\neq} \lim L$ as $N\to \infty$, so the gradient dissimilarity remains strictly positive.

With WoLA, local gradients become asymptotically identical, regardless of the training objective shared by the server.
\paragraph{Proposition 2}\begin{itshape}Let $q$ be the training objective distribution shared by the server. Under WoLA, the gradient of worker $i$ converges almost surely as $N \to \infty$:  
$$
W_i\underset{N\rightarrow \infty}{\overset{\text{a.s.}}{\longrightarrow}} \sum_{c=1}^C 
q^c
\mathbb{E}\Big[\nabla \ell(Y, \Phi(X))\Big|Y = c\Big]
$$
Moreover, the average gradient also satisfies:
$$
W\underset{N\rightarrow \infty}{\overset{\text{a.s.}}{\longrightarrow}} \sum_{c=1}^C 
q^c
\mathbb{E}\Big[\nabla \ell(Y, \Phi(X))\Big|Y = c\Big]
$$
\end{itshape}
\noindent As a result, the gradient dissimilarity under WoLA vanishes as $N \to \infty$.

When the training objective matches the global label distribution (i.e., $q = p$), the WoLA and non-WoLA gradients become asymptotically identical.
\paragraph{Proposition 3}\begin{itshape} Let $q$ be the training objective and $p$ the global label distribution. If $q=p$, then:
$$
\lim_{N\to\infty} W \overset{\text{a.s.}}{=}\lim_{N\to\infty} L
$$
\end{itshape}

\noindent In the asymptotic regime, when the training objective matches the global label distribution, WoLA and non-WoLA gradients target the same model.

\definecolor{gold}{RGB}{245, 185, 0}
\definecolor{silver}{RGB}{5, 145, 250}
\definecolor{bronze}{RGB}{230, 57, 80}

\begin{table*}
  \caption{Test accuracy (± standard deviation) in \%, averaged over training, on Fashion MNIST and CIFAR10. Global results are averaged across aggregators (CWMed, CwTM, GM, MKrum) and attacks (ALIE, FOE, LF, Mimic, SF). Worst-case results correspond to the worst attack, averaged across aggregators. Colors rank values within columns: yellow (1st), blue (2nd), red (3rd). See full table in Appendix.
  }\label{main_worst_results_1}
  
  \begin{subtable}{\textwidth}
    \centering
    \vspace{3mm}
    \caption{Fashion MNIST (Global)}
    \vspace{-1mm}
    \fontsize{6.25}{6.25}\selectfont
    \setlength{\tabcolsep}{2pt}
\begin{tabular}{l|p{1.1cm}p{1.1cm}p{1.1cm}p{1.1cm}|p{1.1cm}p{1.1cm}p{1.1cm}p{1.1cm}|p{1.1cm}p{1.1cm}p{1.1cm}p{1.1cm}}
\midrule[1pt]
Strategy & \multicolumn{4}{c|}{$\alpha=3\quad f=2,4,6,8$} & \multicolumn{4}{c|}{$\alpha=1\quad f=2,4,6,8$} & \multicolumn{4}{c}{$\alpha=0.3\quad f=2,4,6,8$}\\
\midrule
\multicolumn{1}{c}{} & \multicolumn{12}{c}{Accuracy}\\
\midrule

- &
77.1 (0.7) & 74.0 (1.2) & 64.7 (3.1) & 45.3 (6.3) & 
76.6 (1.9) & 71.6 (2.9) & 59.2 (5.5) & 32.3 (7.7) & 
72.6 (2.6) & 60.3 (4.9) & 41.2 (7.4) & 16.7 (6.9) \\

BKT &
76.8 (0.7) & 74.5 (1.1) & 64.7 (3.1) & 45.3 (6.3) & 
77.2 (1.7) & 73.6 (2.5) & 59.2 (5.5) & 32.3 (7.7) & 
75.7 (2.8) & 67.4 (4.7) & 41.2 (7.4) & 16.7 (6.9) \\

FoundFL &
71.2 (3.1) & 62.2 (5.8) & 52.3 (3.9) & 44.7 (7.7) & 
63.6 (6.9) & 54.9 (7.0) & 47.7 (8.9) & 41.6 (9.5) & 
47.8 (10.6) & 41.7 (9.6) & 38.1 (9.1) & 31.2 (9.3) \\

NNM &
\textbf{\color{bronze}78.4 (0.6)} & 76.5 (0.9) & 72.3 (1.8) & 54.8 (5.9) & 
\textbf{\color{bronze}79.4 (1.3)} & 76.7 (1.7) & 69.9 (3.3) & 48.8 (7.4) & 
\textbf{\color{gold}79.4 (1.9)} & 74.0 (3.5) & 54.2 (9.2) & 30.5 (6.7) \\[0.8em]

WoLA &
\textbf{\color{gold}78.6 (0.6)} & \textbf{\color{silver}77.6 (0.6)} & \textbf{\color{bronze}75.8 (0.8)} & \textbf{\color{bronze}63.9 (1.4)} & 
\textbf{\color{silver}79.5 (1.3)} & \textbf{\color{silver}78.1 (1.6)} & \textbf{\color{silver}76.0 (1.7)} & \textbf{\color{silver}62.3 (4.0)} & 
78.0 (1.8) & \textbf{\color{silver}76.3 (2.3)} & \textbf{\color{silver}70.8 (4.8)} & \textbf{\color{silver}42.3 (7.3)} \\

WoLA+NNM &
\textbf{\color{gold}78.6 (0.6)} & \textbf{\color{gold}78.0 (0.6)} & \textbf{\color{gold}77.1 (0.7)} & \textbf{\color{gold}68.5 (1.0)} & 
\textbf{\color{gold}79.6 (1.3)} & \textbf{\color{gold}78.6 (1.5)} & \textbf{\color{gold}77.5 (1.4)} & \textbf{\color{gold}67.5 (4.0)} & 
\textbf{\color{silver}78.5 (1.8)} & \textbf{\color{gold}77.4 (2.3)} & \textbf{\color{gold}73.9 (5.0)} & \textbf{\color{gold}48.8 (8.6)} \\

WoLA$^{\dagger}$ &
\textbf{\color{silver}78.5 (0.5)} & \textbf{\color{bronze}77.3 (0.5)} & 74.5 (0.9) & 59.1 (2.3) & 
79.3 (1.5) & 76.9 (1.6) & 72.8 (2.0) & 55.0 (6.0) & 
77.7 (1.7) & 73.7 (2.4) & 64.9 (4.3) & 36.4 (11.2) \\

WoLA$^{\dagger}$+NNM &
\textbf{\color{silver}78.5 (0.6)} & \textbf{\color{silver}77.6 (0.4)} & \textbf{\color{silver}75.9 (0.8)} & \textbf{\color{silver}64.2 (2.1)} & 
\textbf{\color{gold}79.6 (1.5)} & \textbf{\color{bronze}77.3 (1.6)} & \textbf{\color{bronze}74.9 (1.5)} & \textbf{\color{bronze}55.7 (9.3)} & 
\textbf{\color{bronze}78.3 (1.6)} & \textbf{\color{bronze}75.0 (2.3)} & \textbf{\color{bronze}69.3 (4.8)} & \textbf{\color{bronze}38.9 (8.3)} \\

\midrule
\multicolumn{1}{c}{}& \multicolumn{12}{c}{Accuracy Gain Over Top Non-WoLA Methods}\\
\midrule
WoLA &
+0.2 & +1.1 & +3.6 & +9.1 & 
+0.1 & +1.4 & +6.1 & +13.4 & 
-1.4 & +2.2 & +16.6 & +11.1 \\

WoLA+NNM &
+0.2 & +1.5 & +4.8 & +13.7 & 
+0.2 & +1.9 & +7.6 & +18.7 & 
-0.9 & +3.4 & +19.8 & +17.6 \\

WoLA$^{\dagger}$ &
+0.1 & +0.7 & +2.2 & +4.3 & 
-0.1 & +0.2 & +2.9 & +6.2 & 
-1.7 & -0.4 & +10.7 & +5.3 \\

WoLA$^{\dagger}$+NNM &
+0.1 & +1.1 & +3.6 & +9.4 & 
+0.2 & +0.6 & +5.0 & +6.9 & 
-1.1 & +0.9 & +15.1 & +7.7 \\

\midrule[1pt]
\end{tabular}
\setlength{\tabcolsep}{6pt}

  \end{subtable}

  \begin{subtable}{\textwidth}
    \centering
    \vspace{3mm}
    \caption{CIFAR10 (Global)}
    \vspace{-1mm}
    \fontsize{6.25}{6.25}\selectfont
    \setlength{\tabcolsep}{2pt}
\begin{tabular}{l|p{1.1cm}p{1.1cm}p{1.1cm}p{1.1cm}|p{1.1cm}p{1.1cm}p{1.1cm}p{1.1cm}|p{1.1cm}p{1.1cm}p{1.1cm}p{1.1cm}}
\midrule[1pt]
Strategy & \multicolumn{4}{c|}{$\alpha=3\quad f=2,4,6,8$} & \multicolumn{4}{c|}{$\alpha=1\quad f=2,4,6,8$} & \multicolumn{4}{c}{$\alpha=0.3\quad f=2,4,6,8$}\\
\midrule
\multicolumn{1}{c}{} & \multicolumn{12}{c}{Accuracy}\\
\midrule

- &
69.4 (1.1) & 60.4 (2.1) & 41.4 (4.2) & 21.5 (4.2) & 
63.9 (1.7) & 48.1 (3.3) & 32.2 (3.6) & 16.1 (3.5) & 
51.9 (3.5) & 36.7 (4.5) & 24.5 (4.8) & 12.0 (3.5) \\

BKT &
67.9 (1.0) & 61.6 (2.0) & 41.4 (4.2) & 21.5 (4.2) & 
62.3 (1.8) & 51.4 (3.3) & 32.2 (3.6) & 16.1 (3.5) & 
52.8 (2.9) & 37.5 (3.7) & 24.5 (4.8) & 12.0 (3.5) \\

FoundFL &
45.1 (12.2) & 35.1 (9.9) & 32.6 (9.6) & 22.5 (8.6) & 
27.2 (6.8) & 20.5 (6.1) & 25.2 (5.8) & 19.3 (5.7) & 
20.0 (4.4) & 18.2 (4.6) & 18.3 (3.9) & 15.2 (4.6) \\

NNM &
\textbf{\color{silver}71.7 (0.8)} & 66.9 (1.5) & 56.7 (3.9) & 38.8 (7.0) & 
\textbf{\color{bronze}68.7 (1.2)} & 59.1 (3.4) & 40.6 (5.2) & 29.2 (6.5) & 
\textbf{\color{bronze}62.7 (2.4)} & 48.7 (3.7) & 32.5 (5.4) & \textbf{\color{silver}22.5 (3.9)} \\[0.8em]

WoLA &
\textbf{\color{bronze}71.6 (0.6)} & \textbf{\color{bronze}67.2 (0.8)} & \textbf{\color{bronze}58.8 (1.2)} & \textbf{\color{bronze}40.0 (2.2)} & 
\textbf{\color{bronze}68.7 (0.7)} & \textbf{\color{bronze}63.2 (1.1)} & \textbf{\color{bronze}51.9 (2.9)} & \textbf{\color{bronze}32.3 (3.9)} & 
62.5 (1.8) & \textbf{\color{bronze}54.0 (3.1)} & \textbf{\color{bronze}38.9 (4.7)} & 20.3 (5.0) \\

WoLA+NNM &
\textbf{\color{gold}72.7 (0.5)} & \textbf{\color{gold}70.0 (0.8)} & \textbf{\color{gold}64.2 (1.1)} & \textbf{\color{gold}51.6 (2.1)} & 
\textbf{\color{gold}70.3 (0.6)} & \textbf{\color{gold}66.9 (0.9)} & \textbf{\color{gold}59.1 (2.7)} & \textbf{\color{gold}45.0 (3.8)} & 
\textbf{\color{gold}65.4 (1.5)} & \textbf{\color{gold}60.1 (3.1)} & \textbf{\color{gold}49.0 (3.9)} & \textbf{\color{gold}34.2 (7.2)} \\

WoLA$^{\dagger}$ &
71.5 (0.6) & 66.4 (0.9) & 56.3 (1.6) & 34.4 (3.3) & 
68.4 (0.7) & 62.1 (1.5) & 46.9 (3.9) & 25.6 (6.4) & 
62.2 (1.9) & 52.4 (3.7) & 33.8 (5.5) & 16.3 (4.8) \\

WoLA$^{\dagger}$+NNM &
\textbf{\color{gold}72.7 (0.5)} & \textbf{\color{silver}69.4 (0.8)} & \textbf{\color{silver}62.5 (1.2)} & \textbf{\color{silver}45.4 (3.2)} & 
\textbf{\color{silver}69.9 (0.7)} & \textbf{\color{silver}66.2 (1.1)} & \textbf{\color{silver}55.4 (2.8)} & \textbf{\color{silver}37.5 (8.5)} & 
\textbf{\color{silver}65.0 (1.6)} & \textbf{\color{silver}58.2 (3.2)} & \textbf{\color{silver}42.7 (4.9)} & \textbf{\color{bronze}21.3 (7.5)} \\

\midrule
\multicolumn{1}{c}{}& \multicolumn{12}{c}{Accuracy Gain Over Top Non-WoLA Methods}\\
\midrule
WoLA &
-0.1 & +0.3 & +2.2 & +1.2 & 
-0.1 & +4.1 & +11.3 & +3.1 & 
-0.3 & +5.4 & +6.4 & -2.1 \\

WoLA+NNM &
+1.0 & +3.2 & +7.5 & +12.9 & 
+1.5 & +7.9 & +18.5 & +15.8 & 
+2.6 & +11.4 & +16.4 & +11.8 \\

WoLA$^{\dagger}$ &
-0.2 & -0.5 & -0.4 & -4.4 & 
-0.3 & +3.0 & +6.3 & -3.6 & 
-0.5 & +3.7 & +1.3 & -6.2 \\

WoLA$^{\dagger}$+NNM &
+0.9 & +2.5 & +5.9 & +6.7 & 
+1.2 & +7.1 & +14.7 & +8.3 & 
+2.2 & +9.5 & +10.2 & -1.2 \\

\midrule[1pt]
\end{tabular}
\setlength{\tabcolsep}{6pt}

  \end{subtable}

  \begin{subtable}{\textwidth}
    \centering
    \vspace{3mm}
    \caption{Fashion MNIST (Worst-case)}
    \vspace{-1mm}
    \fontsize{6.25}{6.25}\selectfont
    \setlength{\tabcolsep}{2pt}
\begin{tabular}{l|p{1.1cm}p{1.1cm}p{1.1cm}p{1.1cm}|p{1.1cm}p{1.1cm}p{1.1cm}p{1.1cm}|p{1.1cm}p{1.1cm}p{1.1cm}p{1.1cm}}
\midrule[1pt]
Strategy & \multicolumn{4}{c|}{$\alpha=3\quad f=2,4,6,8$} & \multicolumn{4}{c|}{$\alpha=1\quad f=2,4,6,8$} & \multicolumn{4}{c}{$\alpha=0.3\quad f=2,4,6,8$}\\
\midrule
\multicolumn{1}{c}{} & \multicolumn{12}{c}{Accuracy}\\
\midrule

- &
75.6 (0.8) & 69.6 (1.3) & 50.6 (7.7) & 18.3 (8.8) & 
73.7 (2.6) & 64.6 (4.0) & 40.8 (8.5) & 10.1 (4.7) & 
66.7 (2.6) & 45.2 (5.1) & 22.5 (6.1) & 4.6 (3.3) \\

BKT &
75.9 (0.8) & 71.4 (1.5) & 50.6 (7.7) & 18.3 (8.8) & 
76.2 (2.0) & 70.2 (2.8) & 40.8 (8.5) & 10.1 (4.7) & 
74.5 (3.3) & 59.2 (6.6) & 22.5 (6.1) & 4.6 (3.3) \\

FoundFL &
65.5 (4.5) & 35.9 (8.8) & 11.7 (1.6) & 12.2 (3.9) & 
53.6 (15.5) & 23.6 (10.3) & 20.6 (11.9) & 15.0 (5.2) & 
31.4 (12.0) & 25.7 (10.1) & 15.8 (5.9) & 14.8 (7.2) \\

NNM &
77.5 (0.5) & 73.3 (1.2) & 62.9 (4.1) & 25.3 (13.4) & 
77.9 (1.6) & 73.4 (2.3) & 54.5 (7.2) & 18.8 (8.0) & 
\textbf{\color{bronze}77.2 (2.2)} & 69.2 (4.7) & 36.6 (12.8) & 14.1 (2.7) \\[0.8em]

WoLA &
\textbf{\color{silver}77.9 (0.6)} & \textbf{\color{bronze}75.8 (0.5)} & \textbf{\color{bronze}71.8 (1.0)} & \textbf{\color{silver}51.5 (1.5)} & 
\textbf{\color{silver}78.9 (1.3)} & \textbf{\color{silver}76.2 (1.6)} & \textbf{\color{silver}71.6 (2.2)} & \textbf{\color{silver}46.5 (7.0)} & 
77.0 (1.9) & \textbf{\color{silver}73.4 (2.5)} & \textbf{\color{silver}58.9 (6.2)} & \textbf{\color{gold}24.7 (7.5)} \\

WoLA+NNM &
\textbf{\color{gold}78.0 (0.7)} & \textbf{\color{gold}76.6 (0.6)} & \textbf{\color{gold}74.8 (0.6)} & \textbf{\color{gold}54.8 (1.7)} & 
\textbf{\color{gold}79.1 (1.4)} & \textbf{\color{gold}76.9 (1.5)} & \textbf{\color{gold}75.0 (1.6)} & \textbf{\color{gold}49.5 (8.3)} & 
\textbf{\color{gold}77.9 (1.8)} & \textbf{\color{gold}74.9 (2.6)} & \textbf{\color{gold}61.6 (11.0)} & \textbf{\color{silver}18.9 (7.9)} \\

WoLA$^{\dagger}$ &
\textbf{\color{silver}77.9 (0.5)} & 75.4 (0.4) & 70.1 (1.0) & 48.1 (2.5) & 
\textbf{\color{bronze}78.7 (1.3)} & 75.1 (1.6) & 67.2 (3.4) & \textbf{\color{bronze}44.5 (5.1)} & 
76.7 (1.5) & 70.1 (2.8) & 47.7 (5.9) & \textbf{\color{bronze}18.0 (7.4)} \\

WoLA$^{\dagger}$+NNM &
\textbf{\color{bronze}77.8 (0.7)} & \textbf{\color{silver}75.9 (0.4)} & \textbf{\color{silver}73.2 (1.1)} & \textbf{\color{bronze}50.8 (2.4)} & 
\textbf{\color{gold}79.1 (1.6)} & \textbf{\color{bronze}75.7 (1.8)} & \textbf{\color{bronze}71.1 (2.0)} & 35.4 (18.6) & 
\textbf{\color{silver}77.8 (1.6)} & \textbf{\color{bronze}71.8 (2.2)} & \textbf{\color{bronze}57.9 (8.2)} & 11.3 (5.1) \\

\midrule
\multicolumn{1}{c}{}& \multicolumn{12}{c}{Accuracy Gain Over Top Non-WoLA Methods}\\
\midrule
WoLA &
+0.5 & +2.5 & +8.9 & +26.1 & 
+1.0 & +2.8 & +17.1 & +27.7 & 
-0.3 & +4.2 & +22.3 & +9.9 \\

WoLA+NNM &
+0.5 & +3.3 & +11.9 & +29.4 & 
+1.2 & +3.5 & +20.5 & +30.7 & 
+0.7 & +5.6 & +25.0 & +4.1 \\

WoLA$^{\dagger}$ &
+0.4 & +2.2 & +7.2 & +22.8 & 
+0.9 & +1.7 & +12.7 & +25.7 & 
-0.5 & +0.9 & +11.1 & +3.2 \\

WoLA$^{\dagger}$+NNM &
+0.3 & +2.6 & +10.3 & +25.4 & 
+1.2 & +2.3 & +16.5 & +16.7 & 
+0.6 & +2.5 & +21.4 & -3.5 \\

\midrule[1pt]
\end{tabular}
\setlength{\tabcolsep}{6pt}

  \end{subtable}

\begin{subtable}{\textwidth}
    \centering
    \vspace{3mm}
    \caption{CIFAR10 (Worst-case)}
    \vspace{-1mm}
    \fontsize{6.25}{6.25}\selectfont
    \setlength{\tabcolsep}{2pt}
\begin{tabular}{l|p{1.1cm}p{1.1cm}p{1.1cm}p{1.1cm}|p{1.1cm}p{1.1cm}p{1.1cm}p{1.1cm}|p{1.1cm}p{1.1cm}p{1.1cm}p{1.1cm}}
\midrule[1pt]
Strategy & \multicolumn{4}{c|}{$\alpha=3\quad f=2,4,6,8$} & \multicolumn{4}{c|}{$\alpha=1\quad f=2,4,6,8$} & \multicolumn{4}{c}{$\alpha=0.3\quad f=2,4,6,8$}\\
\midrule
\multicolumn{1}{c}{} & \multicolumn{12}{c}{Accuracy}\\
\midrule

- &
63.9 (2.9) & 50.1 (4.7) & 25.4 (8.0) & 14.4 (3.5) & 
55.3 (5.3) & 22.2 (5.0) & 16.3 (4.1) & 10.9 (3.1) & 
36.3 (5.4) & 17.7 (5.5) & 15.4 (5.2) & 8.2 (2.7) \\

BKT &
65.2 (1.6) & 55.6 (3.7) & 25.4 (8.0) & 14.4 (3.5) & 
58.3 (3.8) & 34.0 (7.3) & 16.3 (4.1) & 10.9 (3.1) & 
42.2 (6.3) & 18.0 (5.0) & 15.4 (5.2) & 8.2 (2.7) \\

FoundFL &
36.7 (13.2) & 14.0 (4.2) & 10.6 (1.8) & 11.0 (2.8) & 
21.6 (6.2) & 12.0 (2.5) & 11.0 (1.9) & 11.3 (3.3) & 
16.9 (4.0) & 12.6 (3.1) & 12.7 (2.8) & 9.6 (2.1) \\

NNM &
65.0 (2.3) & 53.3 (4.6) & 42.0 (8.6) & 19.2 (8.3) & 
58.8 (4.1) & 31.3 (8.9) & 15.2 (4.2) & 14.0 (3.0) & 
45.7 (5.8) & 17.9 (6.4) & 15.6 (3.7) & 11.8 (3.1) \\[0.8em]

WoLA &
\textbf{\color{bronze}69.7 (0.9)} & \textbf{\color{bronze}64.0 (1.1)} & \textbf{\color{bronze}49.0 (1.5)} & \textbf{\color{bronze}23.5 (3.0)} & 
\textbf{\color{bronze}66.0 (1.1)} & \textbf{\color{bronze}59.2 (2.1)} & \textbf{\color{bronze}39.7 (5.1)} & \textbf{\color{bronze}21.7 (2.1)} & 
\textbf{\color{bronze}57.0 (2.2)} & \textbf{\color{silver}46.9 (4.3)} & \textbf{\color{bronze}28.4 (3.8)} & \textbf{\color{bronze}11.9 (2.5)} \\

WoLA+NNM &
\textbf{\color{silver}70.5 (1.0)} & \textbf{\color{gold}65.2 (1.6)} & \textbf{\color{gold}57.7 (1.4)} & \textbf{\color{gold}34.6 (3.5)} & 
\textbf{\color{gold}67.1 (1.6)} & \textbf{\color{gold}60.3 (2.3)} & \textbf{\color{gold}49.2 (3.5)} & \textbf{\color{gold}27.3 (2.9)} & 
\textbf{\color{gold}59.2 (1.6)} & \textbf{\color{gold}49.7 (6.2)} & \textbf{\color{gold}37.1 (7.3)} & \textbf{\color{gold}22.0 (3.2)} \\

WoLA$^{\dagger}$ &
69.5 (0.9) & 62.9 (1.5) & 44.1 (2.5) & 18.9 (2.5) & 
65.6 (1.3) & 57.4 (2.1) & 33.5 (5.4) & 13.3 (2.6) & 
56.8 (2.2) & 45.1 (6.2) & 21.5 (5.5) & 8.4 (4.7) \\

WoLA$^{\dagger}$+NNM &
\textbf{\color{gold}70.6 (0.9)} & \textbf{\color{silver}64.3 (1.6)} & \textbf{\color{silver}55.0 (1.2)} & \textbf{\color{silver}34.0 (2.7)} & 
\textbf{\color{silver}66.3 (2.0)} & \textbf{\color{silver}59.5 (2.4)} & \textbf{\color{silver}44.2 (4.1)} & \textbf{\color{silver}26.5 (10.4)} & 
\textbf{\color{silver}58.7 (1.8)} & \textbf{\color{bronze}45.6 (6.0)} & \textbf{\color{silver}28.8 (7.2)} & \textbf{\color{silver}13.8 (5.7)} \\

\midrule
\multicolumn{1}{c}{}& \multicolumn{12}{c}{Accuracy Gain Over Top Non-WoLA Methods}\\
\midrule
WoLA &
+4.5 & +8.4 & +6.9 & +4.3 & 
+7.2 & +25.2 & +23.4 & +7.7 & 
+11.3 & +28.8 & +12.8 & +0.1 \\

WoLA+NNM &
+5.4 & +9.5 & +15.7 & +15.5 & 
+8.3 & +26.3 & +32.9 & +13.4 & 
+13.5 & +31.7 & +21.5 & +10.2 \\

WoLA$^{\dagger}$ &
+4.4 & +7.3 & +2.1 & -0.3 & 
+6.8 & +23.4 & +17.2 & -0.7 & 
+11.1 & +27.0 & +5.8 & -3.4 \\

WoLA$^{\dagger}$+NNM &
+5.4 & +8.6 & +13.0 & +14.8 & 
+7.5 & +25.4 & +28.0 & +12.5 & 
+13.0 & +27.5 & +13.1 & +2.0 \\

\midrule[1pt]
\end{tabular}
\setlength{\tabcolsep}{6pt}

  \end{subtable}
  
\end{table*}

\begin{figure}
    \centering
    \begin{subfigure}[b]{0.45\textwidth}
        \centering
        \includegraphics[width=\textwidth]{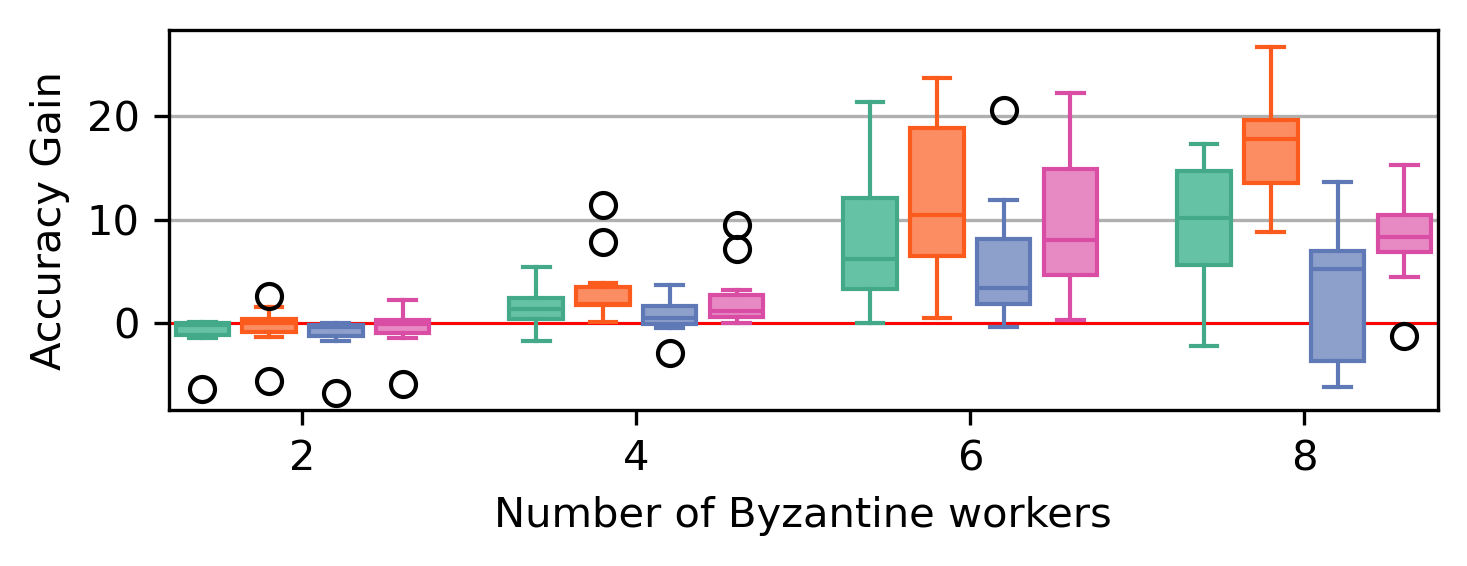}
        \caption{Boxplot of Global accuracy gains.}
        \label{subfig:Accuracy gains overall}
    \end{subfigure}
    \vskip\baselineskip
    \begin{subfigure}[b]{0.45\textwidth}
        \centering
        \includegraphics[width=\textwidth]{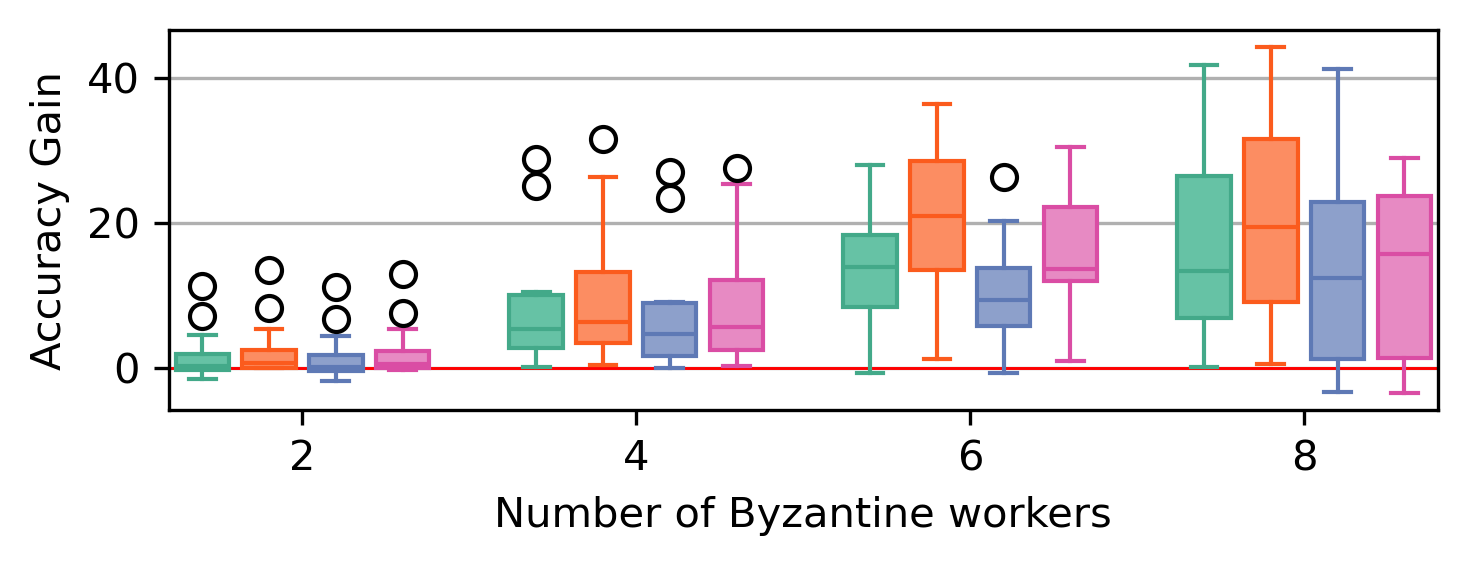}
        \caption{Boxplot of Worst-case accuracy gains.}
        \label{subfig:Accuracy gains worst}
    \end{subfigure}
    \caption{Accuracy gains over the best non-WoLA methods under low, medium, and high heterogeneity settings ($\alpha = 3$, $1$, and $0.3$, respectively), based on the Global and Worst-case subtables reported in Tables~\ref{main_worst_results_1} (Fashion MNIST \& CIFAR10) and~\ref{main_worst_results_2} (MNIST \& Purchase100). Legend:~\textcolor{color1}{$\;\bullet\!$}~WoLA \textcolor{color2}{$\;\bullet\!$}~WoLA+NNM \textcolor{color3}{$\;\bullet\!$}~WoLA$^\dagger$ \textcolor{color4}{$\;\bullet\!$}~WoLA$^\dagger$+NNM}
    \label{fig:Accuracy gains}
\end{figure}

\section{Experimental Setup}\label{Experimental Setup}
This section outlines our experimental setup, including the training objectives of WoLA, baseline methods, and robust aggregators. We describe the datasets and heterogeneity regimes, the distributed system configuration, Byzantine attack scenarios, evaluation metrics, optimization algorithms, models, and hyperparameters.

\subsection{Training Objective, Baselines \& Aggregators} 
We assume the goal is to learn a model that performs well on a test set matching the global label distribution. Accordingly, the ideal WoLA training objective aligns with this distribution, i.e., $q = p$.

We compare WoLA with four baselines: no pre-aggregation (-), Bucketing (BKT)~\cite{karimireddy2020byzantine}, Nearest-Neighbor Mixing (NNM)~\cite{allouah2023fixing}, and FoundationFL (FoundFL)~\cite{fang2025we}. We also evaluate WoLA combined with NNM. Methods requiring a server-side dataset or incurring substantial computational overhead\cite{karimireddy2020byzantine, data2021byzantine} are excluded.

To evaluate robustness to training objective attacks, we introduce WoLA$^\dagger$, a worst-case scenario where Byzantine workers submit highly misleading label distributions. A simple robust aggregator is used to counter the attack (see Appendix, Section \textit{Training Objective Attack}). We also assess WoLA$^\dagger$ combined with NNM.

All methods are tested with four robust aggregators: CWMed, CwTM, GM, and MKrum.

\subsection{Datasets \& Heterogeneity} We conduct experiments on four datasets: MNIST~\cite{lecun1998gradient}, Fashion-MNIST~\cite{xiao2017fashion}, CIFAR10~\cite{krizhevsky2009learning}, and Purchase100~\cite{shokri2017membership}. Due to space constraints, most results for MNIST and Purchase100 are provided in the Appendix.

To simulate data heterogeneity across workers, we sample from the global dataset using a Dirichlet distribution with concentration parameter $\alpha$, following the method in~\cite{allouah2023fixing, hsu2019measuring}. We consider three heterogeneity levels—low, medium, and high—corresponding to $\alpha \in \{3, 1, 0.3\}$. Each worker holds a local dataset of fixed size, equal to the global dataset size divided by the number of honest workers.

\subsection{Distributed system \& Byzantine attacks} 
We consider a distributed system with $n = 17$ workers, among which $f \in \{2, 4, 6, 8\}$ are Byzantine. We evaluate five state-of-the-art gradient attacks:  Fall of Empires (FOE)~\cite{xie2020fall}, A Little is Enough (ALIE)~\cite{baruch2019little}, Sign Flipping (SF)~\cite{allen2020byzantine}, Label Flipping (LF)~\cite{allen2020byzantine}, and Mimic~\cite{karimireddy2020byzantine}. 

In the Mimic attack, all Byzantine workers replicate a carefully selected honest update. Inspired by ALIE~\cite{baruch2019little}, we select the honest outlier most surrounded by other honest updates (see Appendix). This makes Byzantine updates harder to detect while remaining effective in heterogeneous settings. Tables~\ref{tab:AccuracyCIFAR103}–\ref{tab:AccuracyCIFAR100.3} in Appendix in Appendix show that Mimic is often the most harmful attack for baseline methods such as –, BKT, and NNM.

\subsection{Metrics \& Results}
We evaluate performance using two metrics averaged over training: Test Accuracy and Gradient Dissimilarity. 

Each experimental scenario—defined by dataset, heterogeneity level, defense method, and attack—is averaged over three runs with seeds $\{1,2,3\}$.

Comprehensive results are provided in Appendix and summarized in the following tables: Table~\ref{main_worst_results_1} presents Test Accuracy averaged over training for Fashion-MNIST and CIFAR10. The global accuracy corresponds to the mean performance across all aggregators (CWMed, CwTM, GM, MKrum) and all attacks (ALIE, FOE, LF, Mimic, SF). The worst-case accuracy refers to the lowest accuracy across attacks, averaged over aggregators. Table~\ref{main_worst_results_2} provides analogous results for MNIST and Purchase100.

Table~\ref{main_worst_results_1} and \ref{main_worst_results_2} include Accuracy Gain Over Top Non-WoLA Methods, computed as the difference between each WoLA variant and the best-performing non-WoLA baseline. These gains are also visualized as boxplots in Figures~\ref{subfig:Accuracy gains overall} and~\ref{subfig:Accuracy gains worst}.

Table~\ref{Gradient Dissimilarity averaged} shows Gradient Dissimilarity averaged over training, showing the mean across all aggregators and attacks.

Due to space constraints, Table~\ref{Gradient Dissimilarity averaged} and Table~\ref{main_worst_results_2} are provided in the Appendix.

\subsection{Optimization \& Models} 
WoLA relies on approximating the global class-wise gradients with local class-wise gradients. The more data per class available at each optimization step, the better this approximation. As a result, the best performance is likely achieved using non-stochastic gradient descent, where honest workers compute gradients over the entire dataset. However, in practice, stochastic methods are more common. To align with this setting, we follow~\cite{allouah2023fixing} and adopt the distributed stochastic heavy ball method (see Algorithm in Appendix). This method has shown strong robustness against Byzantine behavior in both heterogeneous~\cite{allouah2023fixing} and homogeneous settings~\cite{el2021distributed, karimireddy2021learning, farhadkhani2022byzantine}. We use momentum $\beta=0.9$ for MNIST and Fashion-MNIST, and $\beta=0.99$ for CIFAR10 and Purchase100. 

We use cross-entropy as point-wise loss, batch size $b=128$, learning rates that decay over time, gradient clipping at $5$, and $\ell_2$ regularization of $10^{-4}$.

We train a lightweight CNN for $T=800$ steps on MNIST and Fashion-MNIST. For CIFAR10, we use a deeper CNN, and for Purchase100, a fully connected network. Both are trained for $T=2000$ steps (details in Appendix).

\section{Experimental Results}
In this section, we present experiments to answer the following Research Questions (RQ). We use WoLA methods to refer collectively to WoLA, WoLA$^\dagger$, and their combinations with NNM.

\vspace{.2cm}\noindent\textbf{RQ\hspace{.1cm}1} Are WoLA and WoLA$^\dagger$ more robust overall than state-of-the-art strategies?

\vspace{.1cm}\noindent\textbf{RQ\hspace{.1cm}2} Are WoLA and WoLA$^\dagger$ more robust in worst-case scenarios than state-of-the-art strategies?

\vspace{.1cm}\noindent\textbf{RQ\hspace{.1cm}3} Does combining WoLA or WoLA$^\dagger$ with NNM improve robustness?

\noindent In Appendix, we also answer to questions:

\vspace{.2cm}\noindent\textbf{RQ\hspace{.1cm}A} Does WoLA guide training toward a better model?

\vspace{.1cm}\noindent\textbf{RQ\hspace{.1cm}B} What explains the robustness of WoLA methods?

\vspace{.1cm}\noindent\textbf{RQ\hspace{.1cm}C} How do batch size and momentum affect WoLA performance?

\subsection{Global Robustness of WoLA \& WoLA$^\dagger$ (RQ\hspace{.1cm}1)}
According to Tables (a) and (b) in Table~\ref{main_worst_results_1}, WoLA and WoLA$^\dagger$ significantly improve robustness. The performance gains over the best state-of-the-art methods are particularly significant when the number of Byzantines is large.

When $f = 2$: WoLA and WoLA$^\dagger$ show similar performance, slightly below the best non-WoLA strategies. Typical negative gains between $-1.1$ and $0$ points (Figure~\ref{subfig:Accuracy gains overall}). 

When $f = 4$: WoLA and WoLA$^\dagger$ still perform similarly and yield slightly better results than non-WoLA methods, with typical gains of a few points (Figure~\ref{subfig:Accuracy gains overall}). According to Table~\ref{main_worst_results_1}, WoLA achieves gains of up to $+5.4$ (CIFAR10, $\alpha = 0.3$).

When $f = 6$: WoLA slightly surpasses WoLA$^\dagger$, and both outperform other methods. WoLA shows typical gains between $+3.3$ and $+12.1$ points, while WoLA$^\dagger$ reaches between $+1.8$ and $+8.1$ (Figure~\ref{subfig:Accuracy gains overall}). Table~\ref{main_worst_results_1} reports maximum gains of $+16.6$ (WoLA) and $+10.3$ (WoLA$^\dagger$) on Fashion MNIST with $\alpha=0.3$.

When $f = 8$: WoLA is better than WoLA$^\dagger$, and both methods clearly outperform all others (Figure~\ref{subfig:Accuracy gains overall}). WoLA typically achieves gains between $+5.6$ and $+14.7$ points (median around $+10.1$, Figure~\ref{subfig:Accuracy gains overall}), while WoLA$^\dagger$ ranges from $-3.7$ to $+7$ (median around $+5.2$, Figure~\ref{subfig:Accuracy gains overall}). An exception occurs with CIFAR10, where both methods perform worse than NNM: $-2.1$ for WoLA and $-6.2$ for WoLA$^\dagger$.

Importantly, when $f > 2$, Table~\ref{main_worst_results_1} shows that WoLA almost always ranks in the top three regardless of the scenario (exceptions: Fashion MNIST $\alpha=0.3$, $f=2$; and CIFAR10 $\alpha=0.3$, $f=2,8$). While WoLA$^\dagger$ rarely ranks among the top values, it usually outperforms non-WoLA strategies.

\subsection{Worst-Case Robustness of WoLA \& WoLA$^\dagger$ (RQ\hspace{.1cm}2)}
According to Tables (c) and (d) in Table~\ref{main_worst_results_1}, Considering the strongest attack for each approach, WoLA and WoLA$^\dagger$ consistently enhance robustness, almost always outperforming other state-of-the-art methods, with gains reaching up to $+28.8$ points (CIFAR10, $\alpha=0.3$, $f=4$, WoLA). 

For $f = 2$: WoLA and WoLA$^\dagger$ perform similarly, slightly surpass non-WoLA methods with typical gains of a few points (Figure~\ref{subfig:Accuracy gains worst}), and up to $+11.3$ points on CIFAR10, $\alpha=0.3$ (Table~\ref{main_worst_results_1}).

For $f = 4$, WoLA and WoLA$^\dagger$ continue to perform similarly, both outperforming all baselines with median gains of approximately $+6.3$ and $+4.6$ points, respectively (Figure~\ref{subfig:Accuracy gains worst}). WoLA achieves gains of up to $+28.8$ (CIFAR-10, $\alpha = 0.3$), while WoLA$^\dagger$ reaches up to $+27.0$ under the same setting.

For $f = 6$ or $8$: WoLA outperforms WoLA$^\dagger$, and both clearly surpass all other methods (Figure~\ref{subfig:Accuracy gains worst}). WoLA achieves median gains around $+13.4$ points, WoLA$^\dagger$ around $+12.4$ (Figure~\ref{subfig:Accuracy gains worst}). Table~\ref{main_worst_results_1} shows that WoLA achieves gains of up to $+28.8$, and WoLA$^\dagger$ up to $+27.0$ (CIFAR-10, $\alpha = 0.3$, $f = 4$).

\subsection{Robustness Gains from Combining WoLA \& WoLA$^\dagger$ with NNM (RQ\hspace{.1cm}3)}
Combining WoLA or WoLA$^\dagger$ with NNM significantly improves robustness, consistently outperforming WoLA, WoLA$^\dagger$ or NNM alone. 

WoLA+NNM consistently achieves the top or top-three performance in Table~\ref{main_worst_results_1}, reflecting strong robustness across settings. Gains over the best non-WoLA methods can reach around $20$ points. For example, on Fashion MNIST with $\alpha=0.3$ and $f=6$, $+19.8$; and on CIFAR10 with $\alpha=1$ and $f=6$, $+18.5$. The median gain is $+10.4$ when $f=6$ and $+17.8$ when $f=8$ (Figure~\ref{subfig:Accuracy gains overall}). Considering worst attacks, improvements are even larger, with median gains around $+20$ for both $f=6$ and $f=8$ (Figure~\ref{subfig:Accuracy gains worst}).

WoLA$^\dagger$+NNM also regularly reaches top-three values in Tables~\ref{main_worst_results_1}, confirming its robustness. It performs similarly to WoLA+NNM when $f \leq 4$, and slightly worse when $f > 4$, but still outperforms WoLA$^\dagger$ alone and all non-WoLA strategies (Figure~\ref{subfig:Accuracy gains overall},~\ref{subfig:Accuracy gains worst}). Importantly, it turns nearly all negative gains of WoLA$^\dagger$ alone into positive ones (Table~\ref{main_worst_results_1}). For instance, on CIFAR10 with $\alpha=3$ and $f=8$, the score improves from $-4.4$ to $+6.7$; with $\alpha=1$ and $f=8$, from $-3.6$ to $+8.3$; and with $\alpha=0.3$ and $f=2$, from $-0.5$ to $+2.2$. Overall, median gains over non-WoLA methods are around $+8$ when $f = 6$ or $8$ (Figure~\ref{subfig:Accuracy gains overall}). Considering worst attacks, the median gain is around $+13$ for both $f=6$ and $f=8$ (Figure~\ref{subfig:Accuracy gains worst}).

\section{Conclusion and Future Work}\label{Conclusion and Future Work}
We proposed WoLA, a simple and effective loss reweighting method to improve robustness in federated learning under label skew. Unlike existing approaches that act after gradient computation, WoLA addresses the root cause of gradient divergence by aligning worker objectives through a shared training distribution. It can be combined with existing defenses like NNM for additional gains.

Empirically, WoLA significantly outperforms state-of-the-art methods: on CIFAR-10 with medium heterogeneity ($\alpha = 1$, $f=6$), it improves accuracy by up to $+11.3$ points over the best non-WoLA baseline. On Fashion MNIST ($\alpha = 0.3$, $f=6$), the gain is $+16.6$ points. Combined with NNM, WoLA+NNM achieves up to $+18.5$ on CIFAR-10 ($\alpha = 1$, $f=6$), and $+19.8$ points more on Fashion MNIST ($\alpha = 0.1$, $f=6$). Worst-case scenarios show even larger gains.

Theoretically, we prove that WoLA aligns honest gradients asymptotically, reducing dissimilarity to zero even under label skew. This holds regardless of the training objective.

Future work includes improving robustness further by refining the weighting scheme—e.g., at the mini-batch level—and designing stronger defenses against manipulations of the training objective.

\bibliography{bib}

\newpage 
\appendix 
\section{Notations}\label{Notations}
All notations used in the paper are summarized in Table~\ref{tab:notations}.

\begin{table}
\caption{Notation—In (c) and (d), superscript indices denote classes, while subscripts denote workers.}\label{tab:notations}

\subfloat[Mathematical]{
\begin{tabular}{|>{\centering\arraybackslash}p{13mm}|p{64mm}|}
\hline
\textbf{Notation} & \textbf{Meaning} \\
\hline
$[k]$ & Set of integers from $1$ to $k$, with $k \in \mathbb{N}$ \\
$\#$ & Cardinality of a set \\
$\mathbf{1}$ & Indicator function\\
$\nabla$ & Gradient \\
$\|\cdot\|$ & $\ell_2$-norm \\
$\mathbb{E}[X]$ & Expected value of a random variable $X$\\
$\mathbb{E}[X|\cdot]$ & Conditional expectation of $X$\\
i.i.d.&	Independent and identically distributed\\
a.s. & Probabilistic property known as almost surely\\
$\vcentcolon=$ & Defined as\\
\hline
\end{tabular}}

\vspace{.4cm}

\subfloat[Federated Learning]{
\begin{tabular}{|>{\centering\arraybackslash}p{13mm}|p{64mm}|}
\hline
\textbf{Notation} & \textbf{Meaning} \\
\hline
$\Phi$ & Model \\
$n$ & Number of workers \\
$\mathcal{H}$ & Set of honest workers, with $\mathcal{H} \subset [n]$ \\
$f$ & Number of Byzantine workers, $f = n - \#\mathcal{H}$ \\
\hline
\end{tabular}}

\vspace{.4cm}

\subfloat[Dataset]{
\begin{tabular}{|>{\centering\arraybackslash}p{13mm}|p{64mm}|}
\hline
\textbf{Notation} & \textbf{Meaning} \\
\hline
$C$ & Number of classes \\
$\mathcal{D}$ & Global training set composed of observation-label pairs $(x, y) \in \mathcal{X} \times [C]$\\
$\mathcal{D}_i$ & Local dataset of worker $i$, with $\mathcal{D} = \cup_{i\in \mathcal{H}} \mathcal{D}_i$ \\
$N$ & Size of the global dataset: $N = \#\mathcal{D}$ \\
$N_i$ & Size of worker $i$’s local dataset: $N_i = \#\mathcal{D}_i$\\
$N^c$ & Number of class-$c$ instances in the global dataset, with $N = \sum_{c=1}^C N^c$ \\
$N_i^c$ & Number of class-$c$ instances in worker $i$’s dataset, with $N_i = \sum_{c=1}^C N_i^c$ \\
$p$ & Label distribution of $\mathcal{D}$: $p=(p^1, p^2, \ldots, p^C)$ with $p^c=N^c/N$\\
$p_i$ & Label distribution of $\mathcal{D}_i$: $p_i=(p_i^1, p_i^2, \ldots, p_i^C)$ with $p^c_i=N_i^c/N_i$\\
\hline
\end{tabular}}

\vspace{.4cm}

\subfloat[Loss]{
\begin{tabular}{|>{\centering\arraybackslash}p{13mm}|p{64mm}|}
\hline
\textbf{Notation} & \textbf{Meaning} \\
\hline
$\mathcal{L}$ & Loss\\
$\ell(y, \Phi(x))$ & Pointwise loss of prediction $\Phi(x)$ and label $y$\\
$\mathcal{L}_i$ & Worker $i$ loss: $\mathcal{L}_i = \sum_{(x,y)\in\mathcal{D}_i} \frac{\ell(y, \Phi(x))}{N_i}$, with $\sum_{i \in \mathcal{H}} \mathcal{L}_i\ N_i/N = \mathcal{L}$  \\
$\mu^c$& Class-wise gradient of class $c$ in $\mathcal{D}$, i.e., $\sum_{(x, y) \in \mathcal{D}} \mathbf{1}(y=c) \nabla \ell(c, \Phi(x))/N^c$\\
$\mu^c_i$& Class-wise gradient of class $c$ in $\mathcal{D}_i$, i.e., $\sum_{(x, y) \in \mathcal{D}_i} \mathbf{1}(y=c) \nabla \ell(c, \Phi(x))/N^c_i$\\
$\mathcal{WL}$ & Worker Label Alignement Loss (WoLA) \\
$\mathcal{W}\!\mathcal{L}_i$ & Worker $i$ WoLA, $\sum_{i \in \mathcal{H}}\mathcal{W}\!\mathcal{L}_i \ N_i/N = \mathcal{WL}$ \\
$q$ & Shared training objective: a probability distribution $q = (q^1, q^2, \ldots, q^C)$, i.e., $\sum_{i \in [C]} q^i = 1$ and $q^i\geq 0$.\\
\hline
\end{tabular}}
\end{table}

\section{Difference in Class-Wise Gradients}
This section illustrates that class-wise gradients differ. Similar observations have been reported in previous studies~\cite{zhang2022federated, bao2024boba}.

\begin{figure}
    \centering
    \includegraphics[width=0.45\textwidth]{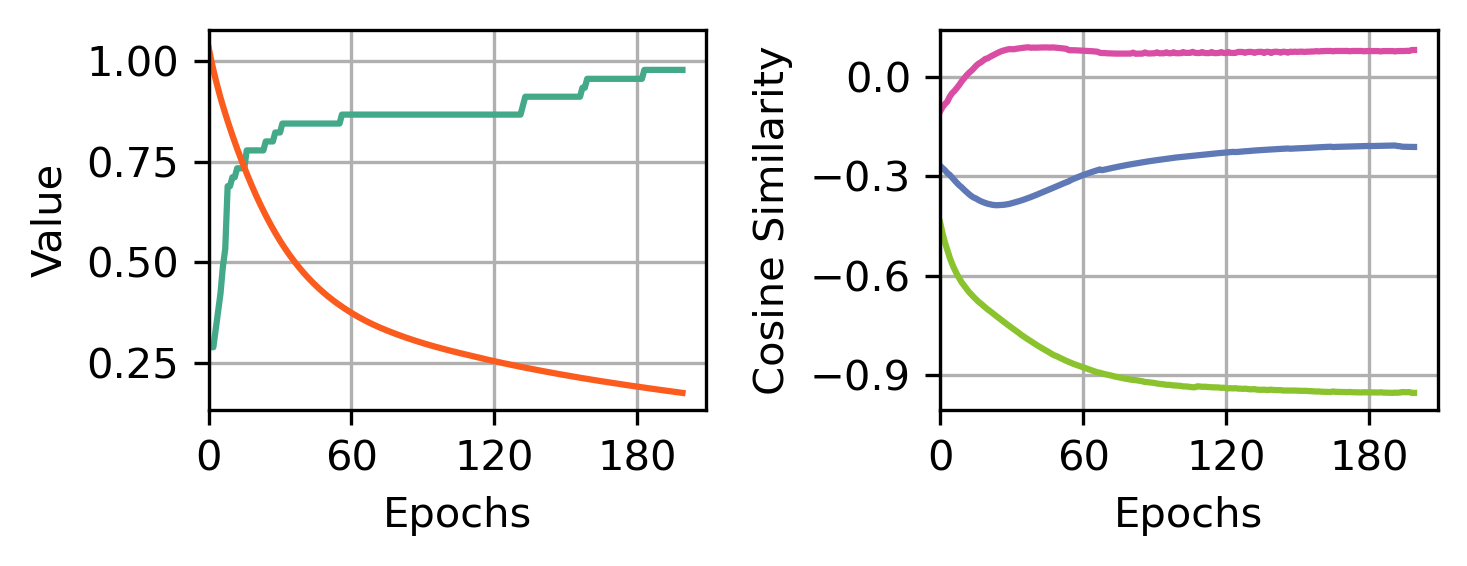} 
    \caption{Training a shallow neural network on the Iris dataset using batch gradient descent. The dataset is split into 70\% training and 30\% testing. While the model successfully learns Iris patterns, the class gradient directions vary significantly between classes. 
    Legend: 
    \textcolor{color2}{$\;\bullet\!$}~Loss 
    \textcolor{color1}{$\;\bullet\!$}~Accuracy 
    Gradient Cosine Similarity —
    \textcolor{color3}{$\,\bullet\!$}~Setosa vs.$\!$ Versicolour 
    \textcolor{color4}{$\;\bullet\!$}~Setosa vs.$\!$ Virginica 
    \textcolor{color5}{$\;\bullet\!$}~Versicolour vs.$\!$ Virginica} 
    \label{fig:example}
\end{figure}

Figure~\ref{fig:example} shows standard (non-FL) training of a shallow neural network on the Iris dataset~\cite{fisher1936use}. The loss decreases and accuracy improves over epochs, indicating successful model convergence. Independently of the optimization, at each step we compute the class-wise gradients Setosa, Versicolour, and Virginica, and track the cosine similarity between these gradients over training steps.

The figure shows that angles between class-wise gradients often exceed $\pi/2$, meaning the gradients point in very different directions. For Versicolour and Virginica, the gradients are nearly opposite. This highlights that different classes can push the model in distinct—and sometimes conflicting—directions during training. 

\section{Importance Sampling}
The WoLA formula is derived from Importance Sampling, a Monte Carlo method that approximates an expectation under a target distribution by a weighted average of samples drawn from a different distribution \cite{tokdar2010importance}. In the following, we provide the detailed derivation of WoLA based on Importance Sampling.

Let $(X, Y)$ be random variables with joint density $f$, where realizations represent points in the global dataset $\mathcal{D}$. Let $q$ be the marginal distribution of $Y$. Similarly, let $(X_i, Y_i)$ be random variables with joint density $f_i$, representing points in the local dataset $\mathcal{D}_i$, and $p_i$ the marginal distribution of $Y_i$. The label distribution skewness assumption states that the conditional distributions of features given labels are identical across workers, i.e., $X|Y$ and $X_i|Y_i$ follow the same law.

From this, we have:
\begin{align*}
&\frac{f(x,y)}{\int f(x,y) dx}=\frac{f_i(x,y)}{\int f_i(x,y) dx}\\
\Leftrightarrow&\frac{f(x,y)}{q(y)}=\frac{f_i(x,y)}{p_i(y)}\\
\Leftrightarrow&\frac{f(x,y)}{f_i(x,y)} = \frac{q(y)}{p_i(y)},
\end{align*}
which leads to:
\begin{align*}
\mathbb{E}\Big(\nabla\ell(Y, \Phi(X))\Big)
&=\int\nabla\ell(y, \Phi(x))f(x,y)dxdy\\
&=\int\nabla\ell(y, \Phi(x))\frac{f(x,y)}{f_i(x,y)}f_i(x,y)dxdy\\
&=\int\nabla\ell(y, \Phi(x))\frac{q(y)}{p_i(y)}f_i(x,y)dxdy\\
&=\mathbb{E}\Big(\nabla\ell(Y_i, \Phi(X_i))\frac{q(Y_i)}{p_i(Y_i)}\Big)\\
\end{align*}
As a result, to estimate $\mathbb{E}\big(\nabla \ell(Y, \Phi(X))\big)$ using data drawn from the distribution of $(X_i, Y_i)$, we have to estimate $\mathbb{E}\big(\nabla \ell(Y_i, \Phi(X_i)) {q(Y_i)}/{p_i(Y_i)}\big)$. This corresponds to the probabilistic form of the WoLA formula.

\section{Additional Research Questions}
This section addresses additional research questions.

\begin{table*}
  \caption{Gradient Dissimilarity (s.d.), averaged over training. Results are averaged across aggregators (CWMed, CwTM, GM, MKrum) and attacks (ALIE, FOE, LF, Mimic, SF). Colors rank values within columns: yellow (1st), blue (2nd), red (3rd).}
  \label{Gradient Dissimilarity averaged}
  
  \begin{subtable}{\textwidth}
    \centering
    \caption{MNIST}
    \vspace{-1mm}
    \fontsize{6.25}{6.25}\selectfont
    \setlength{\tabcolsep}{2pt}
\begin{tabular}{l|p{1.1cm}p{1.1cm}p{1.1cm}p{1.1cm}|p{1.1cm}p{1.1cm}p{1.1cm}p{1.1cm}|p{1.1cm}p{1.1cm}p{1.1cm}p{1.1cm}}
\midrule[1pt]
Strategy & \multicolumn{4}{c|}{$\alpha=3\quad f=2,4,6,8$} & \multicolumn{4}{c|}{$\alpha=1\quad f=2,4,6,8$} & \multicolumn{4}{c}{$\alpha=0.3\quad f=2,4,6,8$}\\
\midrule
\multicolumn{1}{c}{} & \multicolumn{12}{c}{Accuracy}\\
\midrule

- &
\textbf{\color{bronze}0.1 (0.03)} & 0.33 (0.09) & 1.09 (0.3) & 2.07 (0.35) & 
0.69 (0.15) & 1.8 (0.36) & 3.27 (0.85) & 4.28 (1.18) & 
4.46 (0.94) & 7.04 (1.35) & 8.91 (1.55) & 4.56 (0.98) \\

BKT &
\textbf{\color{silver}0.06 (0.01)} & 0.12 (0.03) & 1.09 (0.3) & 2.07 (0.35) & 
\textbf{\color{bronze}0.13 (0.02)} & \textbf{\color{bronze}0.36 (0.14)} & 3.27 (0.85) & 4.28 (1.18) & 
0.55 (0.18) & 2.79 (0.67) & 8.91 (1.55) & 4.56 (0.98) \\

FoundFL &
0.37 (0.13) & 0.84 (0.32) & 0.59 (0.19) & 0.52 (0.27) & 
0.84 (0.45) & 1.28 (0.58) & 0.88 (0.46) & 0.89 (0.47) & 
4.54 (1.83) & 3.86 (1.83) & 3.61 (1.78) & 3.45 (1.86) \\

NNM &
\textbf{\color{silver}0.06 (0.01)} & \textbf{\color{bronze}0.08 (0.02)} & 0.13 (0.03) & 0.83 (0.19) & 
\textbf{\color{silver}0.12 (0.02)} & \textbf{\color{bronze}0.36 (0.24)} & 0.71 (0.31) & 1.92 (0.81) & 
\textbf{\color{bronze}0.38 (0.08)} & 1.22 (0.35) & 4.88 (1.38) & 4.74 (1.02) \\[0.8em]

WoLA &
\textbf{\color{gold}0.03 (0.0)} & \textbf{\color{gold}0.03 (0.0)} & \textbf{\color{bronze}0.06 (0.02)} & 0.11 (0.02) & 
\textbf{\color{gold}0.04 (0.0)} & \textbf{\color{silver}0.06 (0.01)} & \textbf{\color{silver}0.11 (0.03)} & \textbf{\color{silver}0.33 (0.18)} & 
\textbf{\color{silver}0.07 (0.01)} & \textbf{\color{bronze}0.16 (0.05)} & 0.69 (0.35) & \textbf{\color{bronze}1.68 (0.42)} \\

WoLA+NNM &
\textbf{\color{gold}0.03 (0.0)} & \textbf{\color{gold}0.03 (0.0)} & \textbf{\color{silver}0.04 (0.01)} & \textbf{\color{gold}0.05 (0.01)} & 
\textbf{\color{gold}0.04 (0.0)} & \textbf{\color{gold}0.05 (0.01)} & \textbf{\color{gold}0.06 (0.01)} & \textbf{\color{gold}0.19 (0.12)} & 
\textbf{\color{gold}0.05 (0.01)} & \textbf{\color{gold}0.07 (0.01)} & \textbf{\color{silver}0.46 (0.33)} & \textbf{\color{gold}0.81 (0.18)} \\

WoLA$^{\dagger}$ &
\textbf{\color{gold}0.03 (0.0)} & \textbf{\color{silver}0.04 (0.0)} & \textbf{\color{bronze}0.06 (0.01)} & \textbf{\color{bronze}0.1 (0.03)} & 
\textbf{\color{gold}0.04 (0.01)} & \textbf{\color{gold}0.05 (0.01)} & \textbf{\color{bronze}0.13 (0.06)} & 0.68 (0.68) & 
\textbf{\color{silver}0.07 (0.01)} & \textbf{\color{silver}0.15 (0.03)} & \textbf{\color{bronze}0.55 (0.26)} & 2.24 (0.8) \\

WoLA$^{\dagger}$+NNM &
\textbf{\color{gold}0.03 (0.0)} & \textbf{\color{gold}0.03 (0.0)} & \textbf{\color{gold}0.03 (0.01)} & \textbf{\color{silver}0.06 (0.01)} & 
\textbf{\color{gold}0.04 (0.0)} & \textbf{\color{gold}0.05 (0.01)} & \textbf{\color{gold}0.06 (0.02)} & \textbf{\color{bronze}0.36 (0.29)} & 
\textbf{\color{gold}0.05 (0.01)} & \textbf{\color{gold}0.07 (0.01)} & \textbf{\color{gold}0.37 (0.25)} & \textbf{\color{silver}1.5 (0.87)} \\

\midrule[1pt]
\end{tabular}
\setlength{\tabcolsep}{6pt}

  \end{subtable}

  \begin{subtable}{\textwidth}
    \centering
    \vspace{3mm}
    \caption{Fashion MNIST}
    \vspace{-1mm}
    \fontsize{6.25}{6.25}\selectfont
    \setlength{\tabcolsep}{2pt}
\begin{tabular}{l|p{1.1cm}p{1.1cm}p{1.1cm}p{1.1cm}|p{1.1cm}p{1.1cm}p{1.1cm}p{1.1cm}|p{1.1cm}p{1.1cm}p{1.1cm}p{1.1cm}}
\midrule[1pt]
Strategy & \multicolumn{4}{c|}{$\alpha=3\quad f=2,4,6,8$} & \multicolumn{4}{c|}{$\alpha=1\quad f=2,4,6,8$} & \multicolumn{4}{c}{$\alpha=0.3\quad f=2,4,6,8$}\\
\midrule
\multicolumn{1}{c}{} & \multicolumn{12}{c}{Accuracy}\\
\midrule

- &
3.84 (0.56) & 4.2 (0.83) & 4.22 (0.76) & 2.84 (0.62) & 
8.15 (1.76) & 7.67 (1.56) & 7.32 (1.5) & 4.15 (1.39) & 
14.01 (1.24) & 13.47 (1.37) & 11.15 (1.69) & 4.55 (1.5) \\

BKT &
1.26 (0.11) & 1.82 (0.23) & 4.22 (0.76) & 2.84 (0.62) & 
2.45 (0.56) & 3.8 (1.17) & 7.32 (1.5) & 4.15 (1.39) & 
5.5 (0.84) & 8.82 (1.23) & 11.15 (1.69) & 4.55 (1.5) \\

FoundFL &
1.28 (0.69) & 1.31 (0.68) & 1.08 (0.45) & 0.76 (0.32) & 
3.01 (1.32) & 2.18 (1.18) & 2.24 (1.28) & 1.92 (1.01) & 
6.18 (2.86) & 5.93 (2.26) & 5.7 (2.17) & 4.02 (1.96) \\

NNM &
2.54 (0.26) & 2.54 (0.34) & 2.92 (0.48) & 1.92 (0.67) & 
5.11 (1.32) & 5.0 (1.02) & 5.26 (1.52) & 3.23 (1.12) & 
9.94 (1.62) & 10.1 (1.87) & 9.48 (2.09) & 6.0 (1.29) \\[0.8em]

WoLA &
\textbf{\color{silver}0.18 (0.02)} & \textbf{\color{silver}0.17 (0.02)} & \textbf{\color{bronze}0.17 (0.03)} & \textbf{\color{bronze}0.14 (0.02)} & 
\textbf{\color{silver}0.41 (0.17)} & \textbf{\color{silver}0.38 (0.15)} & \textbf{\color{bronze}0.43 (0.17)} & \textbf{\color{silver}0.29 (0.12)} & 
\textbf{\color{bronze}0.89 (0.15)} & \textbf{\color{bronze}1.16 (0.24)} & 1.57 (0.48) & \textbf{\color{bronze}1.51 (0.7)} \\

WoLA+NNM &
\textbf{\color{gold}0.17 (0.02)} & \textbf{\color{gold}0.15 (0.02)} & \textbf{\color{gold}0.13 (0.02)} & \textbf{\color{gold}0.11 (0.02)} & 
\textbf{\color{gold}0.38 (0.16)} & \textbf{\color{gold}0.29 (0.11)} & \textbf{\color{gold}0.28 (0.11)} & \textbf{\color{gold}0.28 (0.14)} & 
\textbf{\color{gold}0.78 (0.12)} & \textbf{\color{gold}0.72 (0.12)} & \textbf{\color{silver}0.97 (0.38)} & \textbf{\color{gold}0.99 (0.33)} \\

WoLA$^{\dagger}$ &
\textbf{\color{bronze}0.19 (0.02)} & \textbf{\color{bronze}0.19 (0.03)} & 0.18 (0.05) & 0.15 (0.04) & 
0.51 (0.2) & 0.51 (0.21) & 0.51 (0.25) & 0.73 (0.7) & 
0.95 (0.17) & 1.35 (0.31) & \textbf{\color{bronze}1.53 (0.27)} & 2.17 (0.88) \\

WoLA$^{\dagger}$+NNM &
\textbf{\color{silver}0.18 (0.02)} & \textbf{\color{gold}0.15 (0.02)} & \textbf{\color{silver}0.14 (0.04)} & \textbf{\color{silver}0.12 (0.03)} & 
\textbf{\color{bronze}0.45 (0.18)} & \textbf{\color{bronze}0.4 (0.16)} & \textbf{\color{silver}0.35 (0.16)} & \textbf{\color{bronze}0.4 (0.29)} & 
\textbf{\color{silver}0.82 (0.1)} & \textbf{\color{silver}0.79 (0.13)} & \textbf{\color{gold}0.91 (0.23)} & \textbf{\color{silver}1.46 (0.75)} \\

\midrule[1pt]
\end{tabular}
\setlength{\tabcolsep}{6pt}

  \end{subtable}

  \begin{subtable}{\textwidth}
    \centering
    \vspace{3mm}
    \caption{Purchase100}
    \vspace{-1mm}
    \fontsize{6.25}{6.25}\selectfont
    \setlength{\tabcolsep}{2pt}
\begin{tabular}{l|p{1.1cm}p{1.1cm}p{1.1cm}p{1.1cm}|p{1.1cm}p{1.1cm}p{1.1cm}p{1.1cm}|p{1.1cm}p{1.1cm}p{1.1cm}p{1.1cm}}
\midrule[1pt]
Strategy & \multicolumn{4}{c|}{$\alpha=3\quad f=2,4,6,8$} & \multicolumn{4}{c|}{$\alpha=1\quad f=2,4,6,8$} & \multicolumn{4}{c}{$\alpha=0.3\quad f=2,4,6,8$}\\
\midrule
\multicolumn{1}{c}{} & \multicolumn{12}{c}{Accuracy}\\
\midrule

- &
0.14 (0.01) & 0.2 (0.02) & \textbf{\color{bronze}0.27 (0.02)} & 0.42 (0.05) & 
0.55 (0.08) & 0.67 (0.1) & 0.99 (0.11) & 1.55 (0.23) & 
2.14 (0.29) & 2.84 (0.33) & 3.59 (0.43) & 3.74 (0.42) \\

BKT &
\textbf{\color{bronze}0.13 (0.01)} & 0.17 (0.01) & \textbf{\color{bronze}0.27 (0.02)} & 0.42 (0.05) & 
\textbf{\color{bronze}0.39 (0.04)} & 0.53 (0.05) & 0.99 (0.11) & 1.55 (0.23) & 
1.29 (0.16) & 2.12 (0.28) & 3.59 (0.43) & 3.74 (0.42) \\

FoundFL &
0.24 (0.02) & 0.32 (0.03) & 0.28 (0.02) & 0.32 (0.03) & 
0.98 (0.17) & 1.01 (0.12) & 0.98 (0.15) & 1.04 (0.17) & 
4.29 (0.73) & 3.79 (0.68) & 3.4 (0.46) & 2.94 (0.54) \\

NNM &
\textbf{\color{silver}0.12 (0.01)} & \textbf{\color{bronze}0.15 (0.01)} & \textbf{\color{silver}0.16 (0.02)} & 0.24 (0.02) & 
\textbf{\color{silver}0.34 (0.04)} & \textbf{\color{bronze}0.45 (0.05)} & \textbf{\color{bronze}0.6 (0.1)} & 0.99 (0.19) & 
\textbf{\color{bronze}1.03 (0.14)} & 1.76 (0.36) & 3.03 (0.52) & 3.45 (0.49) \\[0.8em]

WoLA &
\textbf{\color{gold}0.02 (0.0)} & \textbf{\color{silver}0.02 (0.0)} & \textbf{\color{gold}0.02 (0.0)} & \textbf{\color{silver}0.02 (0.0)} & 
\textbf{\color{gold}0.03 (0.0)} & \textbf{\color{silver}0.04 (0.0)} & \textbf{\color{gold}0.04 (0.0)} & \textbf{\color{silver}0.05 (0.01)} & 
\textbf{\color{silver}0.09 (0.01)} & \textbf{\color{bronze}0.12 (0.01)} & \textbf{\color{silver}0.15 (0.02)} & \textbf{\color{silver}0.24 (0.03)} \\

WoLA+NNM &
\textbf{\color{gold}0.02 (0.0)} & \textbf{\color{gold}0.01 (0.0)} & \textbf{\color{gold}0.02 (0.0)} & \textbf{\color{gold}0.01 (0.0)} & 
\textbf{\color{gold}0.03 (0.0)} & \textbf{\color{gold}0.03 (0.0)} & \textbf{\color{gold}0.04 (0.0)} & \textbf{\color{gold}0.03 (0.0)} & 
\textbf{\color{gold}0.08 (0.0)} & \textbf{\color{gold}0.09 (0.01)} & \textbf{\color{gold}0.11 (0.01)} & \textbf{\color{gold}0.16 (0.02)} \\

WoLA$^{\dagger}$ &
\textbf{\color{gold}0.02 (0.0)} & \textbf{\color{silver}0.02 (0.0)} & \textbf{\color{gold}0.02 (0.0)} & 0.05 (0.01) & 
\textbf{\color{gold}0.03 (0.0)} & \textbf{\color{silver}0.04 (0.0)} & \textbf{\color{silver}0.06 (0.01)} & 0.13 (0.05) & 
\textbf{\color{silver}0.09 (0.0)} & 0.14 (0.03) & 0.26 (0.09) & 0.44 (0.19) \\

WoLA$^{\dagger}$+NNM &
\textbf{\color{gold}0.02 (0.0)} & \textbf{\color{silver}0.02 (0.0)} & \textbf{\color{gold}0.02 (0.0)} & \textbf{\color{bronze}0.03 (0.01)} & 
\textbf{\color{gold}0.03 (0.0)} & \textbf{\color{gold}0.03 (0.0)} & \textbf{\color{gold}0.04 (0.01)} & \textbf{\color{bronze}0.11 (0.06)} & 
\textbf{\color{gold}0.08 (0.0)} & \textbf{\color{silver}0.11 (0.02)} & \textbf{\color{bronze}0.23 (0.09)} & \textbf{\color{bronze}0.41 (0.19)} \\

\midrule[1pt]
\end{tabular}
\setlength{\tabcolsep}{6pt}

  \end{subtable}

\begin{subtable}{\textwidth}
    \centering
    \vspace{3mm}
    \caption{CIFAR10}
    \vspace{-1mm}
    \fontsize{6.25}{6.25}\selectfont
    \setlength{\tabcolsep}{2pt}
\begin{tabular}{l|p{1.1cm}p{1.1cm}p{1.1cm}p{1.1cm}|p{1.1cm}p{1.1cm}p{1.1cm}p{1.1cm}|p{1.1cm}p{1.1cm}p{1.1cm}p{1.1cm}}
\midrule[1pt]
Strategy & \multicolumn{4}{c|}{$\alpha=3\quad f=2,4,6,8$} & \multicolumn{4}{c|}{$\alpha=1\quad f=2,4,6,8$} & \multicolumn{4}{c}{$\alpha=0.3\quad f=2,4,6,8$}\\
\midrule
\multicolumn{1}{c}{} & \multicolumn{12}{c}{Accuracy}\\
\midrule

- &
1.81 (0.27) & 2.49 (0.26) & 2.45 (0.33) & 1.67 (0.48) & 
5.06 (0.8) & 5.73 (0.88) & 4.68 (0.83) & 2.12 (0.96) & 
8.56 (0.7) & 9.14 (0.86) & 7.11 (1.04) & 2.93 (1.25) \\

BKT &
\textbf{\color{bronze}0.8 (0.06)} & \textbf{\color{bronze}1.09 (0.08)} & 2.45 (0.33) & 1.67 (0.48) & 
\textbf{\color{bronze}1.57 (0.2)} & 2.26 (0.49) & 4.68 (0.83) & 2.12 (0.96) & 
2.84 (0.26) & 4.12 (0.64) & 7.11 (1.04) & 2.93 (1.25) \\

FoundFL &
1.36 (0.56) & 1.24 (0.48) & 1.46 (0.56) & 1.08 (0.63) & 
2.45 (1.0) & 1.91 (1.22) & 3.39 (1.24) & 3.08 (1.34) & 
3.84 (1.68) & 3.6 (1.37) & 4.11 (2.21) & 2.93 (1.91) \\

NNM &
1.13 (0.11) & 1.49 (0.23) & \textbf{\color{bronze}1.36 (0.34)} & 1.28 (0.4) & 
2.24 (0.29) & 3.11 (0.74) & 2.29 (0.76) & 1.96 (0.78) & 
3.39 (0.24) & 5.01 (0.92) & 3.43 (0.73) & 3.07 (0.94) \\[0.8em]

WoLA &
\textbf{\color{silver}0.47 (0.03)} & \textbf{\color{silver}0.44 (0.03)} & \textbf{\color{silver}0.38 (0.04)} & \textbf{\color{bronze}0.36 (0.11)} & 
\textbf{\color{silver}0.72 (0.05)} & 0.78 (0.1) & \textbf{\color{bronze}0.68 (0.11)} & \textbf{\color{silver}0.54 (0.14)} & 
1.01 (0.08) & 1.23 (0.24) & \textbf{\color{bronze}1.16 (0.38)} & \textbf{\color{gold}0.87 (0.35)} \\

WoLA+NNM &
\textbf{\color{gold}0.44 (0.02)} & \textbf{\color{gold}0.39 (0.02)} & \textbf{\color{gold}0.37 (0.03)} & 0.43 (0.11) & 
\textbf{\color{gold}0.63 (0.04)} & \textbf{\color{silver}0.63 (0.06)} & \textbf{\color{silver}0.65 (0.1)} & \textbf{\color{bronze}0.61 (0.13)} & 
\textbf{\color{silver}0.84 (0.06)} & \textbf{\color{silver}0.99 (0.14)} & \textbf{\color{silver}1.01 (0.23)} & \textbf{\color{silver}0.94 (0.28)} \\

WoLA$^{\dagger}$ &
\textbf{\color{silver}0.47 (0.03)} & \textbf{\color{silver}0.44 (0.03)} & \textbf{\color{silver}0.38 (0.04)} & \textbf{\color{gold}0.32 (0.05)} & 
\textbf{\color{silver}0.72 (0.05)} & \textbf{\color{bronze}0.76 (0.1)} & 0.69 (0.15) & \textbf{\color{silver}0.54 (0.25)} & 
\textbf{\color{bronze}0.99 (0.09)} & \textbf{\color{bronze}1.18 (0.19)} & 1.18 (0.25) & 1.09 (0.47) \\

WoLA$^{\dagger}$+NNM &
\textbf{\color{gold}0.44 (0.02)} & \textbf{\color{gold}0.39 (0.02)} & \textbf{\color{gold}0.37 (0.04)} & \textbf{\color{silver}0.35 (0.05)} & 
\textbf{\color{gold}0.63 (0.04)} & \textbf{\color{gold}0.62 (0.05)} & \textbf{\color{gold}0.6 (0.08)} & \textbf{\color{gold}0.53 (0.2)} & 
\textbf{\color{gold}0.82 (0.07)} & \textbf{\color{gold}0.96 (0.15)} & \textbf{\color{gold}0.92 (0.22)} & \textbf{\color{bronze}1.07 (0.47)} \\

\midrule[1pt]
\end{tabular}
\setlength{\tabcolsep}{6pt}

  \end{subtable}
  
\end{table*}

\begin{figure}
    \centering
    \includegraphics[width=0.48\textwidth]{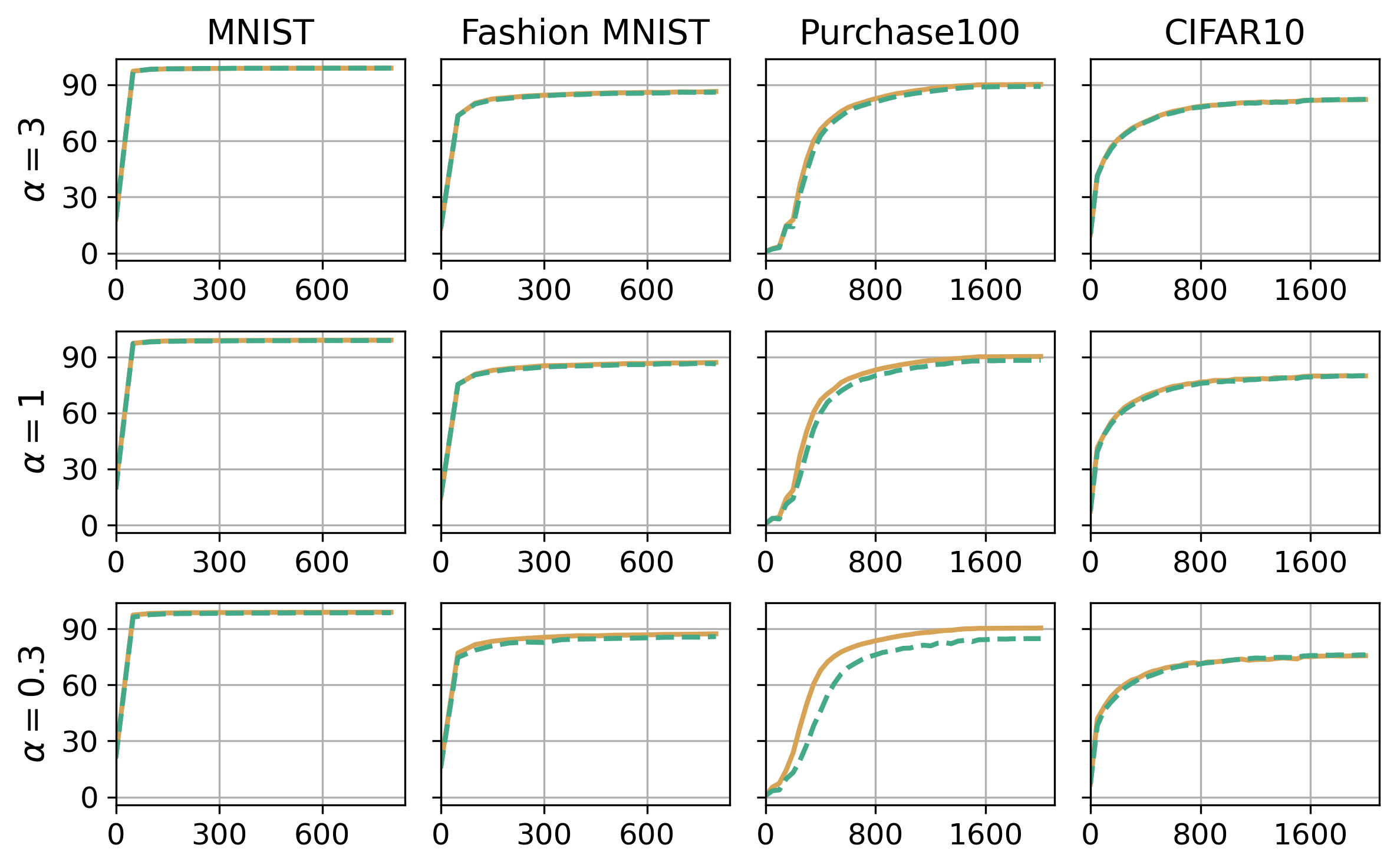}
    \caption{Test Accuracy (averaged over three seeds) without Byzantine workers ($f = 0$), using the mean aggregator on MNIST, Fashion MNIST, Purchase100, and CIFAR10, under low, medium, and high heterogeneity ($\alpha = 3, 1$, and $0.3$). Each column represents a dataset, and each row a heterogeneity level.
 Legend:~\textcolor{color1}{$\;\bullet\!$}~WoLA 
    \textcolor{color7}{$\;\bullet\!$}~-}
    \label{fig:mean}
\end{figure}

\begin{figure}
    \centering
    \begin{subfigure}[b]{0.45\textwidth}
        \centering
        \includegraphics[width=\textwidth]{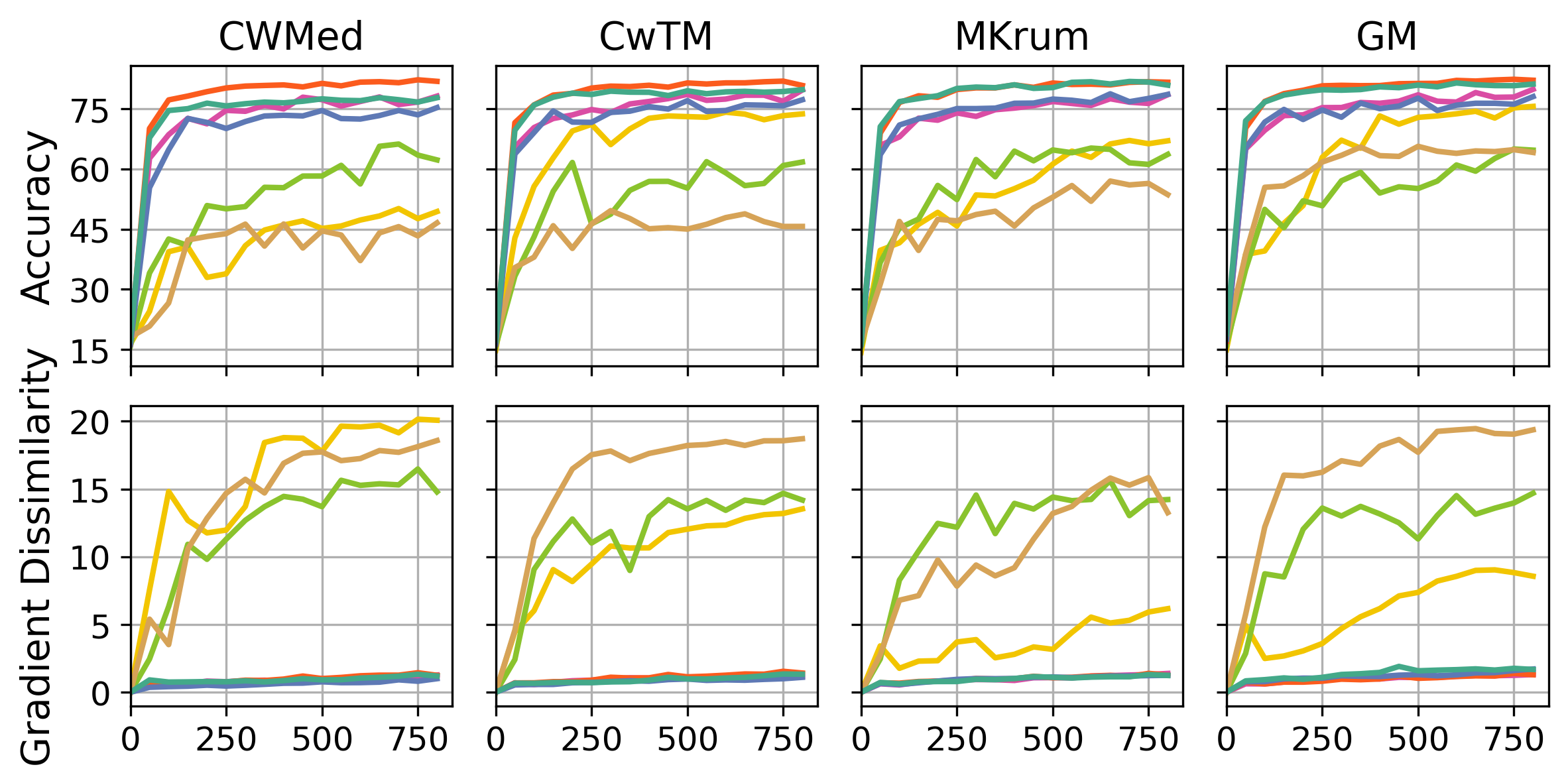}
        \caption{Results under ALIE attack.}
        \label{subfig:alie}
    \end{subfigure}
    \vskip\baselineskip
    \begin{subfigure}[b]{0.45\textwidth}
        \centering
        \includegraphics[width=\textwidth]{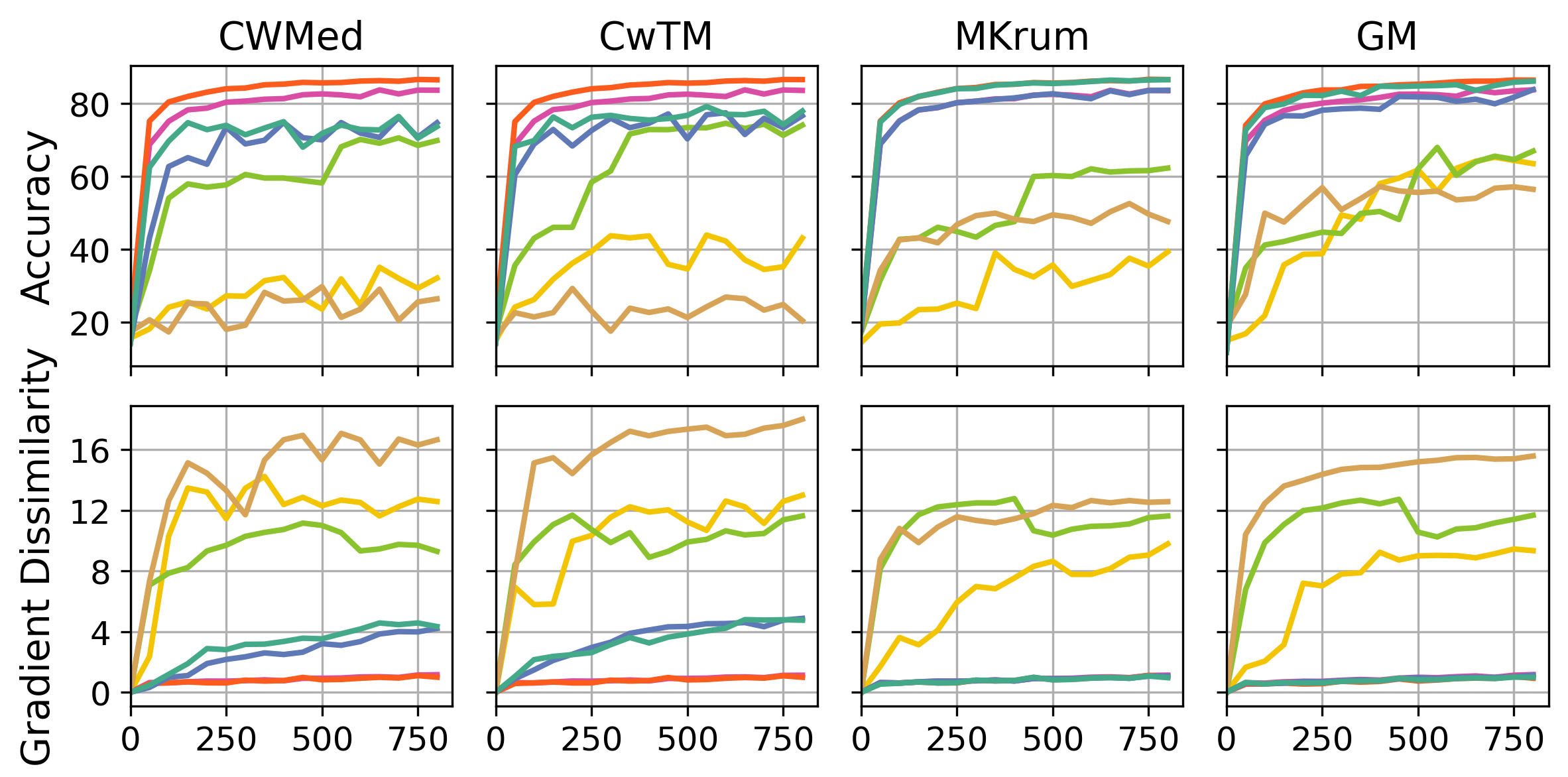}
        \caption{Results under LF attack.}
        \label{subfig:lf}
    \end{subfigure}
    \caption{
    Test Accuracy and Gradient Dissimilarity (averaged over three seeds) under ALIE and LF attacks on Fashion MNIST with high heterogeneity ($\alpha = 0.3$) and $6$ Byzantine workers ($f = 6$). Each column corresponds to a robust aggregator, and each row to an evaluation metric.
 Legend:~\textcolor{color1}{$\;\bullet\!$}~WoLA 
    \textcolor{color2}{$\;\bullet\!$}~WoLA+NNM 
    \textcolor{color3}{$\;\bullet\!$}~WoLA$^\dagger$ 
    \textcolor{color4}{$\;\bullet\!$}~WoLA$^\dagger$+NNM
    \textcolor{color5}{$\;\bullet\!$}~NNM
    \textcolor{color6}{$\;\bullet\!$}~FoundFL
    \textcolor{color7}{$\;\bullet\!$}~- and BKT}
    \label{fig:particular training}
\end{figure}

\begin{figure}
    \centering
    \begin{subfigure}[b]{0.48\textwidth}
        \centering
        \includegraphics[width=\textwidth]{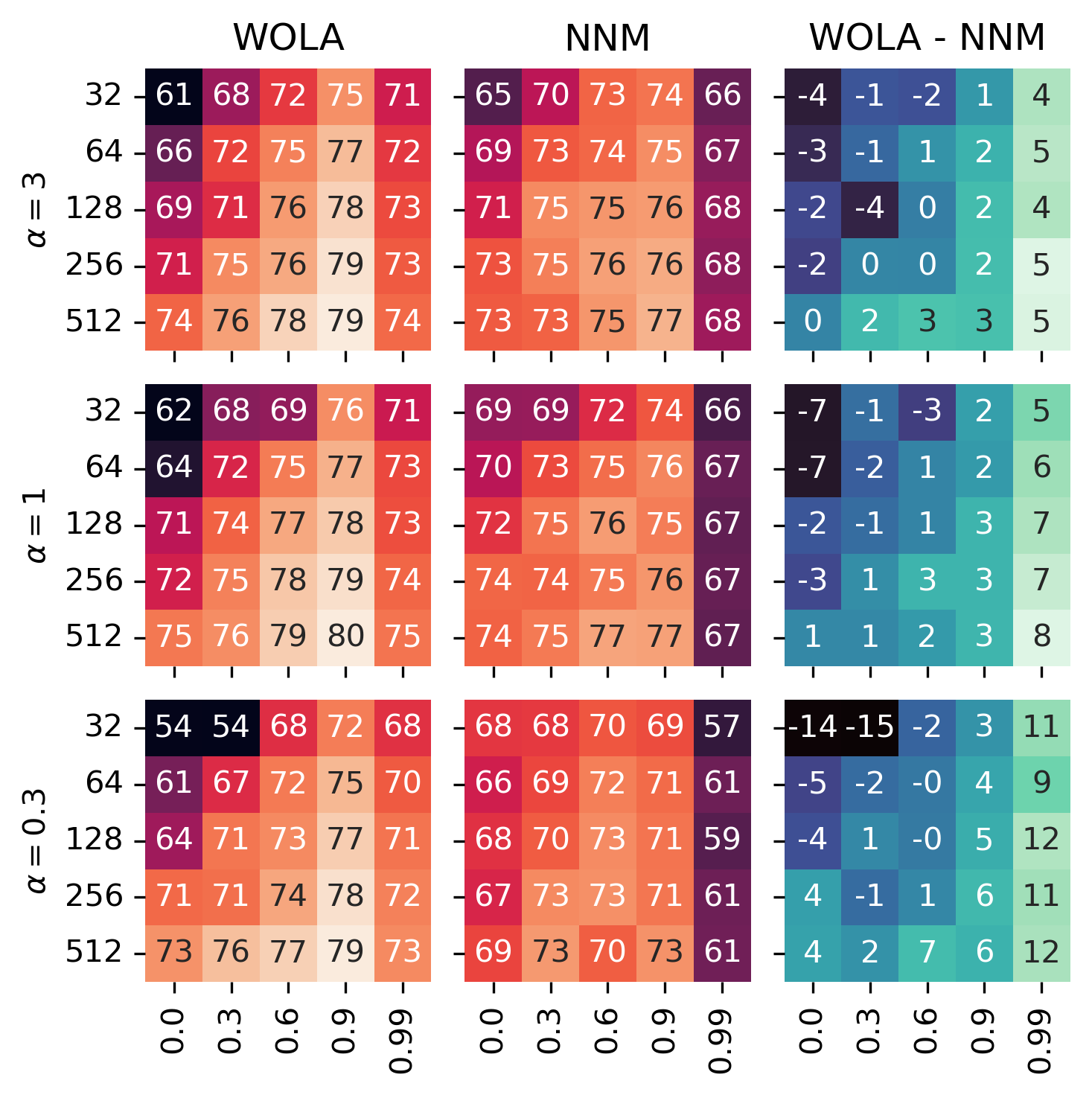}
        \caption{Fashion MNIST}
        \label{subfig:Fashion MNIST heatmaps}
    \end{subfigure}
    \vskip\baselineskip
    \begin{subfigure}[b]{0.48\textwidth}
        \centering
        \includegraphics[width=\textwidth]{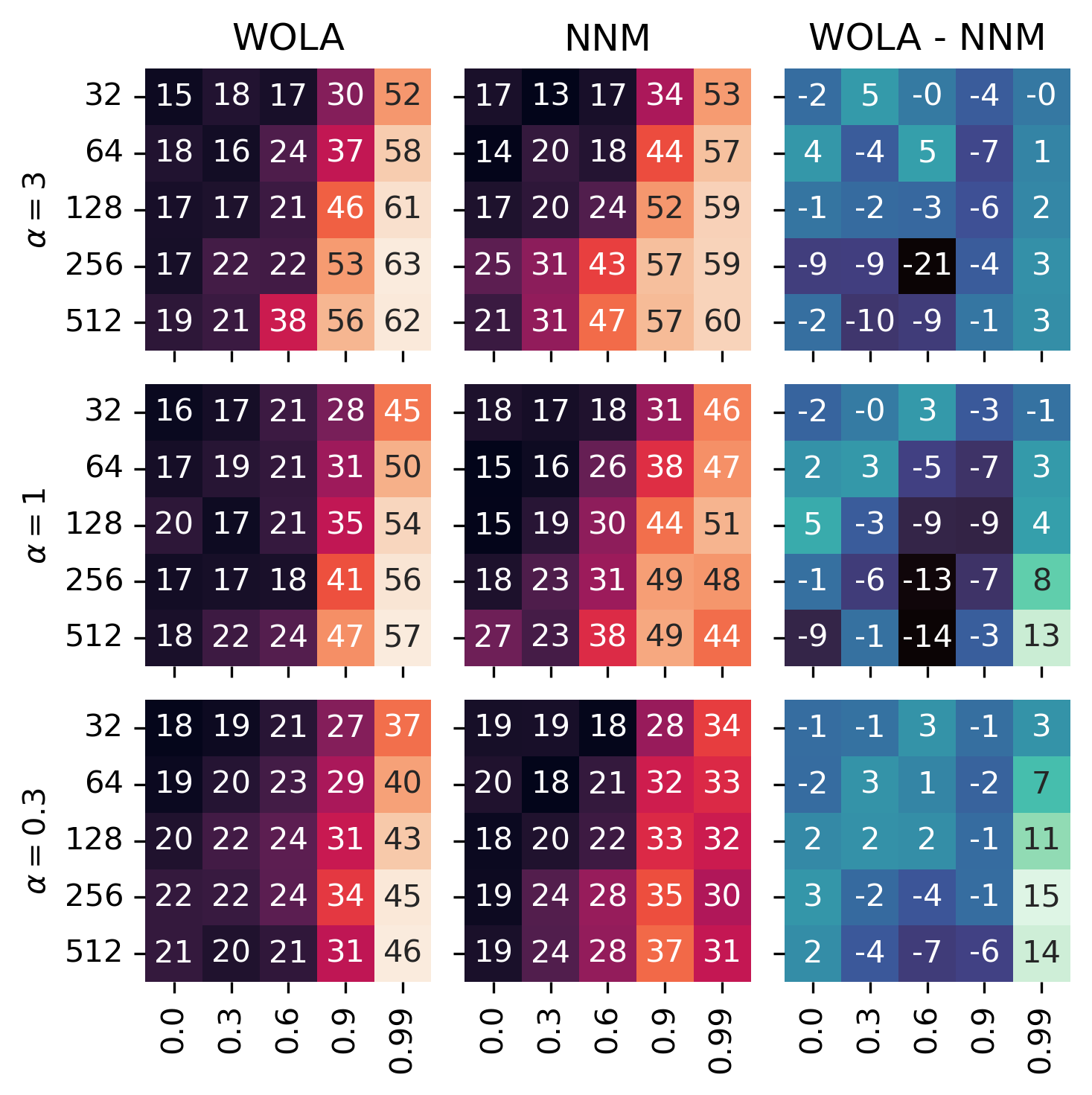}
        \caption{CIFAR10}
        \label{subfig:CIFAR10 heatmaps}
    \end{subfigure}
    \caption{Heatmaps of mean Test Accuracy over training (averaged over three seeds). Each row corresponds to a different level of data heterogeneity: low ($\alpha = 3$), medium ($\alpha = 1$), and high ($\alpha = 0.3$). Each column shows: WoLA accuracy, NNM accuracy, and WoLA’s accuracy gain over NNM. In each heatmap, the vertical axis represents batch size, and the horizontal axis represents momentum. Both scenarios involve $17$ workers ($n=17$), including $5$ Byzantine workers ($f = 5$), under the ALIE attack mitigated by the CwTM aggregator. Accuracy is reported in \%, with a shared color scale for WoLA and NNM accuracy within each row.
}\label{fig:heatmaps}
\end{figure}

\subsection{Targeted Model with WoLA (RQ\hspace{.1cm}A)}
As expected from \textit{Proposition 3} in Subsection WoLA Properties, WoLA does not target a better model during training.

Figure~\ref{fig:mean} shows Test Accuracy over a training without Byzantine workers, using simple mean aggregation. The model trained with WoLA performs similarly to the unweighted loss, though slightly worse on Purchase100 under high heterogeneity.

\subsection{Sources of Robustness in WoLA Methods (RQ\hspace{.1cm}B)}
Gradient dissimilarity is a key indicator of system robustness~\citep{karimireddy2020byzantine, allouah2023fixing}. As expected from \textit{Proposition 2} in Subsection WoLA Properties, WoLA methods significantly reduce gradient dissimilarity, thereby enhancing robustness against Byzantine attacks.

Table~\ref{Gradient Dissimilarity averaged} shows that WoLA and its variants consistently achieve much lower gradient dissimilarity across all scenarios. Interestingly, combining NNM with WoLA further reduces gradient dissimilarity, which may explain its greater robustness over WoLA alone.

Figure~\ref{fig:particular training} shows the evolution of Test Accuracy and Gradient Dissimilarity during training on Fashion-MNIST with high heterogeneity ($\alpha=0.3$) and six Byzantine workers ($f=6$) under two attacks (ALIE and LF). We also observe that WoLA's methods maintain lower gradient dissimilarity than other state-of-the-art approaches.

\subsection{Impact of Batch Size \& Momentum on WoLA (RQ\hspace{.1cm}C)}
Figure~\ref{fig:heatmaps} reports the mean Test Accuracy on Fashion MNIST and CIFAR-10 achieved by WoLA under three heterogeneity levels ($\alpha = 3$, $1$, and $0.3$), with $17$ workers including $5$ Byzantine nodes ($f = 5$), using the ALIE attack and the CwTM aggregator. The figure also shows the performance of NNM and the accuracy gain of WoLA over NNM.

For Fashion MNIST, regardless of the heterogeneity level or momentum value, increasing the batch size consistently improves WoLA's performance. Similarly, increasing the momentum coefficient improves performance up to $0.9$.

For CIFAR-10, under high heterogeneity ($\alpha = 0.3$), increasing the batch size (up to $256$) or momentum coefficient improves performance. Under low and medium heterogeneity ($\alpha = 3$ and $1$), the effects are less consistent. However, with high momentum ($0.9$ or $0.99$), increasing the batch size still tends to help. The best performance is observed with a high momentum coefficient ($0.99$).

Comparing WoLA and NNM, we find that the best results are obtained with moderately large batch sizes and high momentum. These conclusions are specific to the given setting and are not a substitute for a full-scale study, but they provide useful insights into parameter sensitivity.

\section{Training Objective Attack}\label{WoLA under Training Objective Attack}
Our experiments include scenarios where workers share local label distributions with the server, while Byzantine machines manipulate the training objective by sending malicious distributions. We describe this attack setting below.

\subsection{WoLA$^\dagger$ Scenario}
Similar to how Byzantine workers can send arbitrary gradient updates with full knowledge of honest gradients, we assume they can also send arbitrary label distributions with full knowledge of the honest workers’ local label distributions. Rather than enumerating all possible attacks, we focus on a worst-case attack: Byzantine machines submit a zero vector except for a single entry set to $1$, corresponding to the minority class in the global dataset (proofs are provided in the next subsection).

To defend against such attacks, we adopt the geometric median as the aggregation method. This choice offers two main advantages: (i) Unlike methods such as Trimmed Mean or Krum, which exclude outliers, the geometric median treats all updates equally. This is crucial since outlier distributions are not necessarily malicious; excluding them risks discarding honest data and harming performance. (ii) When aggregating probability distributions, the geometric median produces a valid probability distribution—unlike the coordinate-wise median—thus preserving the training objective’s expected form.

Our aim is to show that even a worst-case attack has limited impact on model performance when countered by a simple defense. We denote our method under this attack scenario as WoLA$^\dagger$.

\subsection{Proof that WoLA$^\dagger$ is a Worst-Case Attack}\label{Worst Training Objective Attack}
We show that the training objective $q$ is maximally deviated by a Byzantine attack that concentrates all its mass on the class with minimal weight among honest workers, under the $L_1$ norm denoted by $\|\cdot\|_1$.

Let $u$ and $v$ be two probability distributions over $C$ classes, with $u = (u^1, u^2, \ldots, u^C)$ such that $u^i \in [0,1]$ and $\sum_{i=1}^C u^i = 1$, and similarly for $v$. Let $u^{\min}$ and $u^{\max}$ denote the minimum and maximum components of $u$, and $v^{\max}$ the maximum of $v$.

We first recall the identity:
\begin{equation*}
\begin{aligned}
    \|u - v\|_1 = \sum_{i=1}^C |u^i - v^i| &= \sum_{i=1}^C u^i + v^i - 2\min(u^i, v^i) \\&= 2 - 2\sum_{i=1}^C \min(u^i, v^i).
\end{aligned}
\end{equation*}

Observe that $v^{\max} \geq u^{\min}$. Otherwise, we would have $v^i < u^i$ for all $i$, which implies $\sum_i v^i < \sum_i u^i = 1$, contradicting the assumption that $v$ is a probability distribution. It follows that:
\begin{equation*}
\begin{aligned}
u^{\min} \leq \min(u^{\min}, v^{\max}) &\leq \min(u^{\max}, v^{\max}) \\&\leq \sum_{i=1}^C \min(u^i, v^i),
\end{aligned}
\end{equation*}
which gives the upper bound:
\begin{equation*}
\|u - v\|_1 \leq 2 - 2u^{\min}.
\end{equation*}

Now, consider the training objective $q = \sum_{i=1}^n q_i$, where each $q_i$ is a probability distribution sent by worker $i$. Let $u = \frac{1}{n-f} \sum_{i \in \mathcal{H}} q_i$ be the average contribution from honest workers, and $v = \frac{1}{f} \sum_{i \notin \mathcal{H}} q_i$ the average contribution from Byzantine workers. Then:
\begin{equation*}
\|q - u\|_1 = \frac{f}{n} \|v - u\|_1 \leq \frac{f}{n}(2 - 2u^{\min}).
\end{equation*}

This bound is tight: it is achieved by setting each $q_i$ for $i \notin \mathcal{H}$ to put all mass on the class where $u$ is minimal, completing the proof.

\section{Theoretical Proofs}\label{Appendix Theoretical Results}
In this section, we define the random counterparts of the local updates and provide all proofs of the theoretical results given in Section~\ref{Theoretical Results}.

\subsection{Gradients Modeling}
In this subsection, we define the random counterparts of $\nabla \mathcal{L}_i$ and $\nabla \mathcal{W}\!\mathcal{L}_i$. Let the global dataset be $\mathcal{D} = \{(x_1, y_1), (x_2, y_2), \ldots, (x_N, y_N)\}$ with $(x_i, y_i) \in \mathcal{X} \times [C]$, and let $z_1, z_2, \ldots, z_N$ be integers in $\mathcal{H}$ such that $z_k = i$ if $(x_k, y_k) \in \mathcal{D}_i$. 

Since $N_i = \sum_{l=1}^N \mathbf{1}(z_l = i)$, the following holds for $\nabla \mathcal{L}_i$:
\begin{align*}
    &\nabla \mathcal{L}_i
    =\frac{1}{N_i} \sum_{(x,y)\in\mathcal{D}_i} \ell\left(y, \Phi\left(x\right)\right)\\
    =&\frac{N}{N_i} \frac{\sum_{k=1}^{N} \mathbf{1}(z_k=i)\ell\left(y_{k}, \Phi\left(x_{k}\right)\right)}{N}\\
    =&\Bigg(\frac{\sum_{l=1}^N \mathbf{1}(z_l = i)}{N} \Bigg)^{-1}\frac{\sum_{k=1}^N 
    \mathbf{1}(z_k=i)\nabla \ell(y_k, \Phi(x_k))}{N}
\end{align*}

Let $L_i$ be the random counterpart of $\nabla \mathcal{L}_i$. Replacing realizations with random variables (as defined in Subsection~\ref{Data Modeling}), $L_i$ is given by:
\begin{equation}\label{Li}
\begin{aligned}
    L_i=&
    \Bigg(\frac{\sum_{l=1}^N \mathbf{1}(Z_l = i)}{N} \Bigg)^{-1}\\
    &\quad\quad\frac{\sum_{k=1}^N 
    \mathbf{1}(Z_k=i)\nabla \ell(Y_k, \Phi(X_k))}{N}
\end{aligned}
\end{equation}

We now turn to $\nabla \mathcal{W}\!\mathcal{L}_i$:
\begin{align*}
&\nabla \mathcal{W}\!\mathcal{L}_i = \frac{1}{N_i} \sum_{(x,y)\in\mathcal{D}_i}  \frac{q^y}{p^y_i}\nabla\ell\Big(y, \Phi(x)\Big)\\
=& \frac{1}{N_i} \sum_{k=1}^{N} \mathbf{1}(z_k=i) \frac{q^{y_k}}{p^{y_k}_i} \nabla \ell\Big(y_k, \Phi(x_k)\Big)\\
=& \frac{1}{N_i} \sum_{k=1}^{N}\mathbf{1}(z_k=i) \sum_{c=1}^{C}\mathbf{1}(y_k=c) \frac{q^c}{N_i^c / N_i} \nabla\ell\Big(y_k, \Phi(x_k)\Big)\\
=& \sum_{c=1}^Cq^c \frac{N}{N_i^c} \frac{\sum_{k=1}^{N}\mathbf{1}(y_k=c)\mathbf{1}(z_k=i)\nabla\ell\Big(y_k, \Phi(x_k)\Big)}{N}
\end{align*}
Since $N^c_i = \sum_{l=1}^N \mathbf{1}(y_l = c)\mathbf{1}(z_l = i)$, we have:
\begin{align*}
\nabla \mathcal{W}\!\mathcal{L}_i =&\sum_{c=1}^C 
    q^c
    \Bigg(\frac{\sum_{l=1}^N \mathbf{1}(y_l = c) \mathbf{1}(z_l = i)}{N}\Bigg)^{-1}\\
    &\quad\Bigg(\frac{\sum_{k=1}^N\mathbf{1}(y_k = c)\mathbf{1}(z_k = i)\nabla \ell(y_k, \Phi(x_k))}{N}\Bigg)
\end{align*}

Let $W_i$ be the random counterpart of $\nabla \mathcal{W}\!\mathcal{L}_i$. Replacing realizations with random variables (as defined in Subsection~\ref{Data Modeling}), $W_i$ is given by:
\begin{equation}\label{Si}
\begin{aligned}
W_i=& \sum_{c=1}^C 
    q^c
    \Bigg(\frac{\sum_{l=1}^N \mathbf{1}(Y_l = c) \mathbf{1}(Z_l = i)}{N}\Bigg)^{-1}\\
    &\quad\Bigg(\frac{\sum_{k=1}^N\mathbf{1}(Y_k = c)\mathbf{1}(Z_k = i)\nabla \ell(Y_k, \Phi(X_k))}{N}\Bigg)
\end{aligned}
\end{equation}

\subsection{Lemmas}\label{Lemmas}
In this subsection, we give two lemmas that will be used in the subsequent theoretical proofs.

\paragraph{Lemma 1}\begin{itshape}
Let $X$, $Y$, and $Z$ be random variables such that $X \mid Y$ is independent of $Z \mid Y$. Let $f$ be a function defined on the space of realizations of $X$. Then, for all $y$ and $z$ such that $\mathbb{P}(Y = y) > 0$ and $\mathbb{P}(Z = z) > 0$, we have:
\begin{equation*}
\mathbb{E}\Big[f(X) \Big| Y = y, Z = z\Big] = \mathbb{E}\Big[f(X) \Big| Y = y\Big]
\end{equation*}
\end{itshape}

\begin{proof}
\begin{align*}
&\mathbb{E}\Big[f(X) \Big| Y = y, Z = z\Big]
=\frac{\mathbb{E}\Big[\mathbf{1}(Y = y)\mathbf{1}(Z = z)f(X)\Big]}{\mathbb{P}(Y = y, Z = z)}\\
=&\mathbb{E}\Big[\mathbf{1}(Z = z)f(X)\Big| Y = y \Big]\frac{\mathbb{P}(Y = y)}{\mathbb{P}(Y = y, Z = z)}\\
=&\mathbb{E}\Big[\mathbf{1}(Z = z)\Big| Y = y \Big]\mathbb{E}\Big[f(X)\Big| Y = y \Big]\frac{\mathbb{P}(Y = y)}{\mathbb{P}(Y = y, Z = z)}\\
=&\frac{\mathbb{E}\Big[\mathbf{1}(Z = z)\mathbf{1}(Y = y) \Big]}{\mathbb{P}(Y = y)}\mathbb{E}\Big[f(X)\Big| Y = y \Big]\frac{\mathbb{P}(Y = y)}{\mathbb{P}(Y = y, Z = z)}\\
=& \mathbb{E}\Big[f(X)\Big| Y = y \Big]
\end{align*}
\end{proof}

\paragraph{Lemma 2}\begin{itshape}
The following expression converges almost surely (\textit{a.s.}) as $N \to \infty$:
\begin{align*}
&\Bigg(\frac{\sum_{l=1}^N \mathbf{1}(Y_l = c) \mathbf{1}(Z_l = i)}{N}\Bigg)^{-1} \\
&\quad\quad\Bigg(\frac{\sum_{k=1}^N\mathbf{1}(Y_k = c)\mathbf{1}(Z_k = i)\nabla \ell(Y_k, \Phi(X_k))}{N}\Bigg)\\
&\quad\quad\quad\quad\underset{N\rightarrow \infty}{\overset{\text{a.s.}}{\longrightarrow}} \mathbb{E}\Big[\nabla \ell(Y, \Phi(X))\Big|Y = c\Big]
\end{align*}
\end{itshape}
\begin{proof}
By the continuity of the involved functions and the fact that the product of limits equals the limit of the product, it holds:
\begin{align*}
    &\lim_{N\to\infty} \Bigg(\frac{\sum_{l=1}^N \mathbf{1}(Y_l = c) \mathbf{1}(Z_l = i)}{N}\Bigg)^{-1}\\ &\quad\quad\quad\quad\Bigg(\frac{\sum_{k=1}^N\mathbf{1}(Y_k = c)\mathbf{1}(Z_k = i)\nabla \ell(Y_k, \Phi(X_k))}{N}\Bigg) \\
    &=\Bigg(\lim_{N\to\infty}\frac{\sum_{l=1}^N \mathbf{1}(Y_l = c) \mathbf{1}(Z_l = i)}{N}\Bigg)^{-1}\\
    &\quad\quad\quad\quad\Bigg(\lim_{N\to\infty}\frac{\sum_{k=1}^N\mathbf{1}(Y_k = c)\mathbf{1}(Z_k = i)\nabla \ell(Y_k, \Phi(X_k))}{N}\Bigg)
\end{align*}  
By the \textit{Law of Large Numbers}, as $N \to \infty$, the expressions converge almost surely to:
\begin{align*}
    &\mathbb{E}\Big[\mathbf{1}(Y = c)\mathbf{1}(Z = i)\Big]^{-1}\\
    &\quad\quad\quad\quad\mathbb{E}\Big[\mathbf{1}(Y = c)\mathbf{1}(Z = i)\nabla \ell(Y, \Phi(X))\Big]\\
    =&  
    \frac{\mathbb{E}\Big[\mathbf{1}(Y = c)\mathbf{1}(Z = i)\nabla \ell(Y, \Phi(X))\Big]}{\mathbb{P}(Y = c, Z = i)}\\
    =&
    \mathbb{E}\Big[\nabla \ell(Y, \Phi(X))\Big|Y = c, Z = i\Big]
\end{align*}
According to Lemma 1:
\begin{equation*}
    \mathbb{E}\Big[\nabla \ell(Y, \Phi(X))\Big|Y = c, Z = i\Big]=\mathbb{E}\Big[\nabla \ell(Y, \Phi(X))\Big|Y = c\Big]
\end{equation*}
\end{proof}

\subsection{Proof of Proposition 1}
We begin by calculating the limit of $L_i$. By definition, we have:
    \begin{align*}
        L_i=\Bigg(\frac{\sum_{l=1}^N \mathbf{1}(Z_l = i)}{N} \Bigg)^{-1}\frac{\sum_{k=1}^N 
    \mathbf{1}(Z_k=i)\nabla \ell(Y_k, \Phi(X_k))}{N}
    \end{align*}
By the \textit{Law of Large Numbers}, it holds:
\begin{align*}
&\frac{\sum_{l=1}^N \mathbf{1}(Z_l=i)}{N}\underset{N\rightarrow \infty}{\overset{\text{a.s.}}{\longrightarrow}}\mathbb{E}\Big[\mathbf{1}(Z=i)\Big]=\mathbb{P}(Z=i),\\\\
&\frac{\sum_{k=1}^N 
    \mathbf{1}(Z_k=i)\nabla \ell(Y_k, \Phi(X_k))}{N}\\
    &\quad\quad\quad\quad\underset{N\rightarrow \infty}{\overset{\text{a.s.}}{\longrightarrow}}\mathbb{E}[\mathbf{1}(Z=i)\nabla \ell(Y, \Phi(X))].
\end{align*}
By the continuity of the involved functions and the fact that the product of limits equals the limit of the product, it holds:
\begin{align*}
&\lim_{N\to\infty} L_i
    =\lim_{N\to\infty} \Bigg(\frac{\sum_{l=1}^N \mathbf{1}(Z_l = i)}{N} \Bigg)^{-1}\\
    &\quad\quad\quad\quad\frac{\sum_{k=1}^N 
    \mathbf{1}(Z_k=i)\nabla \ell(Y_k, \Phi(X_k))}{N} \\
    =&\Bigg(\lim_{N\to\infty}\frac{\sum_{l=1}^N \mathbf{1}(Z_l = i)}{N} \Bigg)^{-1}\\
    &\quad\quad\quad\quad\Bigg(\lim_{N\to\infty}\frac{\sum_{k=1}^N 
    \mathbf{1}(Z_k=i)\nabla \ell(Y_k, \Phi(X_k))}{N}\Bigg)\\
    \overset{\text{a.s.}}{=}&\frac{\mathbb{E}[\mathbf{1}(Z=i)\nabla \ell(Y, \Phi(X))]}{\mathbb{P}(Z=i)}
    =\mathbb{E}[\nabla \ell(Y, \Phi(X))|Z=i]
\end{align*} 

Similarly, for $L$:
\begin{align*}
   \lim_{N\to\infty} L=&\lim_{N\to\infty}\sum_{i\in \mathcal{H}} L_i\frac{\sum_{l=1}^N \mathbf{1}(Z_l=i)}{N}\\
   =&\sum_{i\in \mathcal{H}} \lim_{N\to\infty}L_i\lim_{N\to\infty}\frac{\sum_{l=1}^N \mathbf{1}(Z_l=i)}{N}\\
    \overset{\text{a.s.}}{=}& \sum_{i\in \mathcal{H}} \mathbb{E}[\nabla \ell(Y, \Phi(X))|Z=i] \mathbb{P}(Z=i)
\end{align*}
By the \textit{Law of total expectation}, it holds:
\begin{equation*}
    \lim_{N\to\infty} L \overset{\text{a.s.}}{=}\mathbb{E}\Big[\nabla \ell(Y, \Phi(X))\Big]
\end{equation*}

\subsection{Proof of Proposition 2}
We begin by calculating the limit of $W_i$. By the continuity of the involved functions, it holds:
\begin{align*}
    &\lim_{N\to\infty} W_i \\
    =&  \lim_{N\to\infty}
    \sum_{c=1}^C 
    q^c
    \Bigg(\frac{\sum_{l=1}^N \mathbf{1}(Y_l = c) \mathbf{1}(Z_l = i)}{N}\Bigg)^{-1}\\
    &\quad\quad\quad\quad\Bigg(\frac{\sum_{k=1}^N\mathbf{1}(Y_k = c)\mathbf{1}(Z_k = i)\nabla \ell(Y_k, \Phi(X_k))}{N}\Bigg)\\
    =&
    \sum_{c=1}^C 
    q^c\lim_{N\to\infty}
    \Bigg(\frac{\sum_{l=1}^N \mathbf{1}(Y_l = c) \mathbf{1}(Z_l = i)}{N}\Bigg)^{-1}\\
    &\quad\quad\quad\quad\Bigg(\frac{\sum_{k=1}^N\mathbf{1}(Y_k = c)\mathbf{1}(Z_k = i)\nabla \ell(Y_k, \Phi(X_k))}{N}\Bigg)
\end{align*}
Using Lemma 2, as $N \to \infty$, the expression converge almost surely to:
\begin{equation*}
    \lim_{N\to\infty} W_i \overset{\text{a.s.}}{=}
    \sum_{c=1}^C 
    q^c
    \mathbb{E}\Big[\nabla \ell(Y, \Phi(X))\Big|Y = c\Big]
\end{equation*}

Similarly, for $W$:
\begin{equation*}
\begin{aligned}
\lim_{N\to\infty}W=&\lim_{N\to\infty} \sum_{i\in \mathcal{H}} W_i \frac{\sum_{l=1}^N \mathbf{1}(Z_l=i)}{N}\\=&\sum_{i\in \mathcal{H}} \lim_{N\to\infty}W_i \lim_{N\to\infty}\frac{\sum_{l=1}^N \mathbf{1}(Z_l=i)}{N}
\end{aligned}
\end{equation*}
By the \textit{Law of Large Numbers}, it holds:
$$
\lim_{N\to\infty}\frac{\sum_{l=1}^N \mathbf{1}(Z_l=i)}{N}\overset{\text{a.s.}}{=}\mathbb{E}\Big[\mathbf{1}(Z=i)\Big]=\mathbb{P}(Z=i),
$$
Using above results, it concludes the lemma:
\begin{align*}
&\lim_{N\to\infty}W\overset{\text{a.s.}}{=}\sum_{i\in \mathcal{H}} \sum_{c=1}^C 
q^c
\mathbb{E}\Big[\nabla \ell(Y, \Phi(X))\Big|Y = c\Big]
\mathbb{P}(Z=i)\\
=&\sum_{c=1}^C 
q^c
\mathbb{E}\Big[\nabla \ell(Y, \Phi(X))\Big|Y = c\Big]
\end{align*}

\subsection{Proof of Gradient Dissimilarity Vanishing}\label{Gradient Dissimilarity Vanishing}
According to Proposition 2, by the continuity of the gradient dissimilarity formula, we obtain:
\begin{equation*}
\begin{aligned}
&\lim_{N\to\infty} \frac{1}{\#\mathcal{H}} \sum_{i\in \mathcal{H}}\left\|W_i-W\right\|^2\\
=&\frac{1}{\#\mathcal{H}} \sum_{i\in \mathcal{H}}\left\|\lim_{N\to\infty} W_i - \lim_{N\to\infty} W\right\|^2
\overset{\text{a.s.}}{=}0
\end{aligned}
\end{equation*}

\subsection{Proof of Proposition 3}
According to Proposition 2, it holds:
\begin{equation*}
W\underset{N\rightarrow \infty}{\overset{\text{a.s.}}{\longrightarrow}} \sum_{c=1}^C 
q^c
\mathbb{E}\Big[\nabla \ell(Y, \Phi(X))\Big|Y = c\Big]
\end{equation*}
By assumption, it holds $\tilde{p}_c=\mathbb{P}(Y=c)$ for all $c\in[C]$. Then, using the \textit{Law of total expectation}, it follows:
\begin{equation*}
\begin{aligned}
W\underset{N\rightarrow \infty}{\overset{\text{a.s.}}{\longrightarrow}}& \sum_{c=1}^C 
\mathbb{P}(Y=c)
\mathbb{E}\Big[\nabla \ell(Y, \Phi(X))\Big|Y = c\Big]\\
&=\mathbb{E}\Big[\nabla \ell(Y, \Phi(X))\Big]
\end{aligned}
\end{equation*}

\section{Experimental Details}
\subsection{Robust Distributed Stochastic Heavy Ball Algorithm}\label{Robust Distributed Stochastic Heavy Ball Algorithm}
We present here the robust distributed stochastic Heavy Ball algorithm. More details can be found in~\cite{allouah2023fixing}. We use superscript in brackets to denote the iteration index, i.e., $^{(t)}$, and subscripts to indicate the worker index.

\begin{algorithm}
\caption{Robust Distributed Stochastic Heavy Ball}
\label{algo}
\begin{algorithmic}[1]
\REQUIRE Initial model $\theta^{(0)}$, initial momentum $m_i^{(0)} = 0$ for honest workers, robust aggregation $F$, learning rate $\gamma$, momentum coefficient $\beta$, and number of steps $T$
\FOR{$t = 1$ to $T$}
    \STATE Server broadcasts $\theta^{(t-1)}$ to all workers
    \FOR{every honest worker $i$}
        \STATE Compute a stochastic gradient $g^{(t)}_{i}$
        \STATE Update local momentum: $m_i^{(t)} \leftarrow \beta m_i^{(t-1)} + (1 - \beta) g_i^{(t)}$
        \STATE Send $m_i^{(t)}$ to the server
    \ENDFOR
    \STATE Server aggregates the momenta:
    \STATE \hspace{1em} $R^{(t)} = F(m_1^{(t)}, \ldots, m_n^{(t)})$
    \STATE Server updates the model:
    \STATE \hspace{1em} $\theta^{(t)} = \theta^{(t-1)} - \gamma R^{(t)}$
\ENDFOR
\end{algorithmic}
\end{algorithm}

\subsection{Model Architectures}\label{Model Architectures}
We adopt the following abreviations : L(\#outputs) denotes a fully connected linear layer; R is a ReLU activation; T is a Tanh activation; C(\#channels) represents a 2D convolutional layer with kernel size $5$, padding $0$, and stride $1$; M denotes 2D max-pooling with kernel size $2$; B stands for batch normalization; and D represents dropout with a fixed probability of $0.25$.

The model architectures are as follows:
\begin{itemize}
    \item MNIST and Fashion MNIST (CNN from~\cite{allouah2023fixing}): C(20)-R-M–C(20)-R-M–L(500)-R–L(10)
    \item CIFAR10 (CNN from~\cite{allouah2023fixing}):(3,32×32)-C(64)-R-B-C(64)-R-B-M-D-C(128)-R-B-C(128)-R-B-M-D-L(128)-R-D-L(10)
    \item Purchase100 (fully connected neural networks from~\cite{nasr2018machine}): L(1024)-L(512)-T-L(256)-T-L(100)
\end{itemize}

The learning rates are defined as:
\begin{itemize}
    \item MNIST and Fashion MNIST (from~\cite{allouah2023fixing}): $\gamma_t=0.75/(1+\left\lfloor\frac{t}{50}\right\rfloor)$
    \item CIFAR10 (from~\cite{allouah2023fixing}) and Purchase100: $\gamma_t= 0.25$ if $t \leq 1500$, $0.025$ otherwise.
\end{itemize}

\subsection{Mimic with Most Surrounded Outlier Selection}
In the Mimic attack, we choose the honest outlier that is most surrounded by other honest updates. The selection procedure is detailed below. We first introduce the operators $\operatorname{dist}$ and $\operatorname{rank}$ used in this process.

Let $\mathbf{v}_1, \ldots, \mathbf{v}_h$ be the honest update vectors. For each vector $\mathbf{v}_i$, define $\sigma(i, \cdot)$ as the permutation that orders the vectors in decreasing order of their Euclidean distance from $\mathbf{v}_i$:
$$
\operatorname{dist}(\mathbf{v}_i, \mathbf{v}_{\sigma(i,1)}) \geq \cdots \geq \operatorname{dist}(\mathbf{v}_i, \mathbf{v}_{\sigma(i,h)}),
$$
where $\operatorname{dist}$ denotes the Euclidean distance. For any vector $\mathbf{v}_j$, there exists an index $r$ such that $\sigma(i, r) = j$, meaning that $\mathbf{v}_j$ is the $r$‑th farthest vector from $\mathbf{v}_i$. Note that $\sigma(i, h) = i$. We define the operator $\operatorname{rank}$ as:
$$
\operatorname{rank}(\mathbf{v}_i, \mathbf{v}_j) = r,
$$
where $j = \sigma(i, r)$.

The operators $\operatorname{dist}$ and $\operatorname{rank}$ allow us to compute a score for each update based on how surrounded a vector is and how far it is from the mean of the honest updates. The update to mimic is the one with the highest score, see Algorithm~\ref{algo2}.

\begin{algorithm}
\caption{Mimic with Most Surrounded Outlier Selection}
\label{algo2}
\begin{algorithmic}[1]
\REQUIRE Honest updates $\mathbf{v}_1, \ldots, \mathbf{v}_h \in \mathbb{R}^d$, number of Byzantine workers $f$
\ENSURE Update $\mathbf{v}_{\text{argmax}}$ with highest score

\FOR{$i = 1$ to $h$}
    \STATE $s_i \leftarrow 0$
\ENDFOR

\STATE $\textit{center} \leftarrow \frac{1}{h} \sum_{i=1}^{h} \mathbf{v}_i$

\FOR{$i = 1$ to $h$}
    \FOR{$j = 1$ to $h$}
        \STATE $r \leftarrow \operatorname{rank}(\mathbf{v}_i, \mathbf{v}_j)$
        \STATE $s_j \leftarrow s_j + \min(f, r) \cdot \operatorname{dist}(\mathbf{v}_j, \textit{center})$
    \ENDFOR
\ENDFOR

\STATE $k \leftarrow \arg\max_j s_j$
\RETURN $\mathbf{v}_k$
\end{algorithmic}
\end{algorithm}

\section{Full and Additional Results}\label{Comprehensive Results}
We report summarized performance in Table~\ref{main_worst_results_2} for MNIST and Purchase100, which show similar method behaviors as Table~\ref{main_worst_results_1}. Additionally, full experimental results are provided. For MNIST, see Tables~\ref{tab:AccuracyMNIST3},~\ref{tab:AccuracyMNIST1}, and~\ref{tab:AccuracyMNIST0.3}; for Fashion MNIST, see Tables~\ref{tab:AccuracyFashion MNIST3},~\ref{tab:AccuracyFashion MNIST1}, and~\ref{tab:AccuracyFashion MNIST0.3}; for Purchase100, see Tables~\ref{tab:AccuracyPurchase1003},~\ref{tab:AccuracyPurchase1001}, and~\ref{tab:AccuracyPurchase1000.3}; and for CIFAR10, see Tables~\ref{tab:AccuracyCIFAR103},~\ref{tab:AccuracyCIFAR101}, and~\ref{tab:AccuracyCIFAR100.3}. Each set of three tables corresponds to low, medium, and high heterogeneity levels ($\alpha = 3$, $1$, and $0.3$).

\definecolor{gold}{RGB}{245, 185, 0}
\definecolor{silver}{RGB}{5, 145, 250}
\definecolor{bronze}{RGB}{230, 57, 80}

\begin{table*}
  \caption{Test accuracy (± standard deviation) in \%, averaged over training, on MNIST and Purchase100. Global results are averaged across aggregators (CWMed, CwTM, GM, MKrum) and attacks (ALIE, FOE, LF, Mimic, SF). Worst-case results correspond to the worst attack, averaged across aggregators. Colors rank values within columns: yellow (1st), blue (2nd), red (3rd). See full table in Appendix.
  }\label{main_worst_results_2}
  
  \begin{subtable}{\textwidth}
    \centering
    \vspace{3mm}
    \caption{MNIST (Global)}
    \vspace{-1mm}
    \fontsize{6.25}{6.25}\selectfont
    \setlength{\tabcolsep}{2pt}
\begin{tabular}{l|p{1.1cm}p{1.1cm}p{1.1cm}p{1.1cm}|p{1.1cm}p{1.1cm}p{1.1cm}p{1.1cm}|p{1.1cm}p{1.1cm}p{1.1cm}p{1.1cm}}
\midrule[1pt]
Strategy & \multicolumn{4}{c|}{$\alpha=3\quad f=2,4,6,8$} & \multicolumn{4}{c|}{$\alpha=1\quad f=2,4,6,8$} & \multicolumn{4}{c}{$\alpha=0.3\quad f=2,4,6,8$}\\
\midrule
\multicolumn{1}{c}{} & \multicolumn{12}{c}{Accuracy}\\
\midrule

- &
\textbf{\color{silver}93.6 (0.3)} & 92.9 (0.4) & 86.1 (1.6) & 66.3 (4.5) & 
93.6 (0.3) & 92.2 (0.6) & 82.3 (3.3) & 43.5 (11.7) & 
91.9 (1.8) & 84.6 (4.0) & 58.8 (12.1) & 17.3 (7.3) \\

BKT &
\textbf{\color{bronze}93.5 (0.3)} & 92.8 (0.5) & 86.1 (1.6) & 66.3 (4.5) & 
93.6 (0.4) & 92.5 (0.7) & 82.3 (3.3) & 43.5 (11.7) & 
93.3 (0.7) & 89.6 (1.9) & 58.8 (12.1) & 17.3 (7.3) \\

FoundFL &
89.6 (1.8) & 74.4 (4.2) & 56.8 (4.6) & 47.5 (5.0) & 
85.4 (5.2) & 69.6 (7.3) & 57.7 (6.7) & 50.4 (6.8) & 
61.6 (15.9) & 50.4 (12.8) & 43.0 (12.1) & 38.0 (11.8) \\

NNM &
\textbf{\color{gold}93.7 (0.2)} & \textbf{\color{bronze}93.3 (0.3)} & \textbf{\color{bronze}92.7 (0.4)} & 75.5 (6.9) & 
\textbf{\color{silver}93.9 (0.3)} & 93.0 (0.6) & 90.1 (2.5) & 63.2 (9.0) & 
\textbf{\color{gold}93.9 (0.6)} & 91.5 (1.4) & 67.5 (14.5) & 35.2 (8.4) \\[0.8em]

WoLA &
\textbf{\color{gold}93.7 (0.3)} & \textbf{\color{silver}93.4 (0.3)} & \textbf{\color{bronze}92.7 (0.3)} & \textbf{\color{bronze}81.9 (0.6)} & 
\textbf{\color{silver}93.9 (0.3)} & \textbf{\color{silver}93.5 (0.4)} & \textbf{\color{bronze}92.4 (0.4)} & \textbf{\color{silver}80.5 (2.2)} & 
\textbf{\color{bronze}93.6 (0.5)} & \textbf{\color{bronze}93.0 (0.7)} & \textbf{\color{bronze}88.9 (2.6)} & \textbf{\color{silver}54.5 (8.0)} \\
\end{tabular}
\setlength{\tabcolsep}{6pt}

  \end{subtable}

  \begin{subtable}{\textwidth}
    \centering
    \vspace{3mm}
    \caption{Purchase100 (Global)}
    \vspace{-1mm}
    \fontsize{6.25}{6.25}\selectfont
    \setlength{\tabcolsep}{2pt}
\begin{tabular}{l|p{1.1cm}p{1.1cm}p{1.1cm}p{1.1cm}|p{1.1cm}p{1.1cm}p{1.1cm}p{1.1cm}|p{1.1cm}p{1.1cm}p{1.1cm}p{1.1cm}}
\midrule[1pt]
Strategy & \multicolumn{4}{c|}{$\alpha=3\quad f=2,4,6,8$} & \multicolumn{4}{c|}{$\alpha=1\quad f=2,4,6,8$} & \multicolumn{4}{c}{$\alpha=0.3\quad f=2,4,6,8$}\\
\midrule
\multicolumn{1}{c}{} & \multicolumn{12}{c}{Accuracy}\\
\midrule

- &
70.4 (0.7) & 65.3 (1.1) & 50.5 (1.9) & 24.8 (2.8) & 
67.9 (1.0) & 58.0 (2.0) & 37.1 (3.1) & 8.0 (2.1) & 
62.2 (1.8) & 44.4 (3.3) & 15.9 (2.3) & 2.6 (1.1) \\

BKT &
70.1 (0.9) & 64.6 (1.2) & 50.5 (1.9) & 24.8 (2.8) & 
68.2 (1.1) & 59.3 (1.9) & 37.1 (3.1) & 8.0 (2.1) & 
\textbf{\color{silver}64.5 (1.9)} & 49.3 (3.5) & 15.9 (2.3) & 2.6 (1.1) \\

FoundFL &
52.2 (3.6) & 36.5 (3.1) & 28.4 (2.6) & 22.4 (2.4) & 
38.8 (5.9) & 28.9 (4.1) & 20.7 (4.5) & 13.3 (4.9) & 
24.6 (8.2) & 15.6 (6.6) & 10.3 (4.3) & 6.4 (2.8) \\

NNM &
\textbf{\color{gold}71.6 (0.7)} & 67.9 (0.9) & 61.3 (1.8) & 38.1 (2.3) & 
\textbf{\color{gold}70.6 (0.9)} & 63.9 (1.6) & 52.0 (2.8) & 28.3 (4.0) & 
\textbf{\color{gold}69.6 (1.2)} & \textbf{\color{bronze}58.8 (3.2)} & 28.9 (5.6) & 15.1 (3.2) \\[0.8em]

WoLA &
70.6 (0.6) & \textbf{\color{bronze}69.2 (0.6)} & \textbf{\color{bronze}66.7 (0.7)} & \textbf{\color{bronze}52.7 (1.0)} & 
\textbf{\color{bronze}69.2 (0.8)} & \textbf{\color{bronze}67.0 (0.8)} & \textbf{\color{bronze}61.3 (1.2)} & \textbf{\color{silver}43.1 (1.7)} & 
63.2 (1.1) & 57.1 (2.1) & \textbf{\color{bronze}43.6 (2.6)} & \textbf{\color{silver}21.8 (2.5)} \\

WoLA+NNM &
\textbf{\color{silver}70.9 (0.6)} & \textbf{\color{gold}69.8 (0.6)} & \textbf{\color{gold}68.3 (0.7)} & \textbf{\color{gold}57.6 (0.9)} & 
\textbf{\color{silver}69.3 (0.8)} & \textbf{\color{gold}67.8 (0.7)} & \textbf{\color{gold}65.3 (1.1)} & \textbf{\color{gold}53.5 (1.7)} & 
\textbf{\color{bronze}64.0 (1.1)} & \textbf{\color{gold}60.6 (1.7)} & \textbf{\color{gold}52.6 (2.5)} & \textbf{\color{gold}33.0 (3.7)} \\
\end{tabular}
\setlength{\tabcolsep}{6pt}

  \end{subtable}

\begin{subtable}{\textwidth}
    \centering
    \vspace{3mm}
    \caption{MNIST (Worst-case)}
    \vspace{-1mm}
    \fontsize{6.25}{6.25}\selectfont
    \setlength{\tabcolsep}{2pt}
\begin{tabular}{l|p{1.1cm}p{1.1cm}p{1.1cm}p{1.1cm}|p{1.1cm}p{1.1cm}p{1.1cm}p{1.1cm}|p{1.1cm}p{1.1cm}p{1.1cm}p{1.1cm}}
\midrule[1pt]
Strategy & \multicolumn{4}{c|}{$\alpha=3\quad f=2,4,6,8$} & \multicolumn{4}{c|}{$\alpha=1\quad f=2,4,6,8$} & \multicolumn{4}{c}{$\alpha=0.3\quad f=2,4,6,8$}\\
\midrule
\multicolumn{1}{c}{} & \multicolumn{12}{c}{Accuracy}\\
\midrule

- &
\textbf{\color{bronze}93.4 (0.3)} & 91.9 (0.5) & 73.6 (5.1) & 27.5 (5.8) & 
93.3 (0.4) & 90.1 (1.3) & 62.8 (8.1) & 12.6 (5.7) & 
89.3 (3.2) & 70.7 (8.6) & 29.5 (11.5) & 7.1 (3.4) \\

BKT &
93.2 (0.5) & 91.5 (0.9) & 73.6 (5.1) & 27.5 (5.8) & 
93.2 (0.5) & 90.8 (1.5) & 62.8 (8.1) & 12.6 (5.7) & 
92.6 (1.1) & 84.6 (4.0) & 29.5 (11.5) & 7.1 (3.4) \\

FoundFL &
80.4 (4.4) & 36.5 (8.2) & 10.5 (1.2) & 10.9 (1.7) & 
76.4 (15.1) & 27.1 (8.7) & 10.9 (1.2) & 13.5 (3.3) & 
48.5 (22.4) & 20.5 (6.3) & 14.7 (4.0) & 16.9 (7.6) \\

NNM &
\textbf{\color{gold}93.6 (0.2)} & 92.8 (0.3) & \textbf{\color{bronze}91.3 (0.6)} & 47.0 (27.5) & 
\textbf{\color{silver}93.8 (0.3)} & 91.6 (1.7) & 83.9 (8.9) & 22.5 (11.7) & 
\textbf{\color{gold}93.7 (0.6)} & 85.6 (3.7) & 49.6 (24.0) & 18.1 (5.9) \\[0.8em]

WoLA &
\textbf{\color{silver}93.5 (0.3)} & \textbf{\color{bronze}92.9 (0.3)} & 90.6 (0.4) & 66.7 (0.7) & 
\textbf{\color{silver}93.8 (0.3)} & \textbf{\color{bronze}92.9 (0.3)} & \textbf{\color{bronze}89.9 (0.7)} & \textbf{\color{silver}64.3 (3.0)} & 
\textbf{\color{bronze}93.3 (0.5)} & \textbf{\color{silver}92.1 (0.8)} & \textbf{\color{bronze}77.6 (8.1)} & \textbf{\color{gold}28.8 (6.2)} \\

WoLA+NNM &
\textbf{\color{gold}93.6 (0.3)} & \textbf{\color{gold}93.2 (0.3)} & \textbf{\color{gold}92.6 (0.2)} & \textbf{\color{gold}70.5 (1.6)} & 
\textbf{\color{gold}93.9 (0.3)} & \textbf{\color{gold}93.2 (0.3)} & \textbf{\color{gold}92.3 (0.4)} & \textbf{\color{gold}66.7 (4.7)} & 
\textbf{\color{silver}93.6 (0.5)} & \textbf{\color{gold}92.6 (0.7)} & \textbf{\color{silver}78.9 (11.2)} & \textbf{\color{bronze}24.1 (11.4)} \\
\end{tabular}
\setlength{\tabcolsep}{6pt}

  \end{subtable}
  
\begin{subtable}{\textwidth}
    \centering
    \vspace{3mm}
    \caption{Purchase100 (Worst-case)}
    \vspace{-1mm}
    \fontsize{6.25}{6.25}\selectfont
    \setlength{\tabcolsep}{2pt}
\begin{tabular}{l|p{1.1cm}p{1.1cm}p{1.1cm}p{1.1cm}|p{1.1cm}p{1.1cm}p{1.1cm}p{1.1cm}|p{1.1cm}p{1.1cm}p{1.1cm}p{1.1cm}}
\midrule[1pt]
Strategy & \multicolumn{4}{c|}{$\alpha=3\quad f=2,4,6,8$} & \multicolumn{4}{c|}{$\alpha=1\quad f=2,4,6,8$} & \multicolumn{4}{c}{$\alpha=0.3\quad f=2,4,6,8$}\\
\midrule
\multicolumn{1}{c}{} & \multicolumn{12}{c}{Accuracy}\\
\midrule

- &
68.2 (0.6) & 59.2 (0.8) & 29.0 (1.2) & 8.7 (2.3) & 
64.0 (1.5) & 48.5 (2.9) & 18.5 (3.1) & 0.9 (0.3) & 
54.9 (2.5) & 28.5 (3.8) & 2.1 (0.8) & 0.9 (0.3) \\

BKT &
68.6 (0.8) & 58.1 (0.9) & 29.0 (1.2) & 8.7 (2.3) & 
65.8 (1.6) & 50.3 (3.1) & 18.5 (3.1) & 0.9 (0.3) & 
58.2 (3.4) & 30.0 (7.1) & 2.1 (0.8) & 0.9 (0.3) \\

FoundFL &
33.6 (1.9) & 4.2 (1.7) & 0.6 (0.3) & 0.4 (0.1) & 
21.5 (6.8) & 2.6 (1.3) & 1.6 (0.9) & 1.6 (0.9) & 
13.2 (4.7) & 3.2 (1.1) & 2.9 (0.3) & 2.5 (1.3) \\

NNM &
\textbf{\color{gold}70.4 (0.5)} & 63.3 (1.5) & 52.2 (1.4) & 6.4 (1.0) & 
67.1 (1.7) & 53.8 (4.3) & 33.6 (4.4) & 3.2 (1.8) & 
\textbf{\color{gold}61.8 (3.1)} & 36.9 (8.1) & 3.2 (2.3) & 1.0 (0.4) \\[0.8em]

WoLA &
69.7 (0.6) & \textbf{\color{bronze}66.7 (0.6)} & \textbf{\color{bronze}61.4 (0.7)} & \textbf{\color{silver}37.7 (1.0)} & 
\textbf{\color{bronze}68.0 (0.8)} & \textbf{\color{bronze}63.8 (0.8)} & \textbf{\color{bronze}49.0 (1.4)} & 19.3 (1.2) & 
60.3 (1.0) & \textbf{\color{bronze}47.3 (2.3)} & \textbf{\color{bronze}18.2 (2.6)} & \textbf{\color{gold}6.1 (2.8)} \\

WoLA+NNM &
\textbf{\color{silver}70.3 (0.6)} & \textbf{\color{gold}68.0 (0.6)} & \textbf{\color{gold}66.3 (0.6)} & \textbf{\color{gold}44.2 (0.8)} & 
\textbf{\color{gold}68.7 (0.8)} & \textbf{\color{gold}65.8 (0.6)} & \textbf{\color{gold}62.0 (1.5)} & \textbf{\color{gold}37.6 (1.9)} & 
\textbf{\color{silver}61.7 (1.0)} & \textbf{\color{gold}53.5 (1.5)} & \textbf{\color{gold}39.7 (3.3)} & \textbf{\color{bronze}3.0 (1.8)} \\
\end{tabular}
\setlength{\tabcolsep}{6pt}

  \end{subtable}
  
\end{table*}

\renewcommand{\arraystretch}{1.2}
\setlength{\tabcolsep}{1pt}
\begin{table*}
\centering
\tiny
\caption{Accuracy (s.d.) \% on MNIST with $\alpha=3$ and $f=2,4,6,8$. Colors rank values within sub-columns: yellow (1st), blue (2nd), and red (3rd).}
\label{tab:AccuracyMNIST3}
\begin{tabular}{l|l|p{.48cm}p{.48cm}p{.48cm}p{.48cm}|p{.48cm}p{.48cm}p{.48cm}p{.48cm}|p{.48cm}p{.48cm}p{.48cm}p{.48cm}|p{.48cm}p{.48cm}p{.48cm}p{.48cm}|p{.48cm}p{.48cm}p{.48cm}p{.48cm}|p{.48cm}p{.48cm}p{.48cm}p{.48cm}|p{.48cm}p{.48cm}p{.48cm}p{.48cm}}
\midrule[1pt]
\textbf{Agg.} & \textbf{Other Strategies} & \multicolumn{4}{c|}{\textbf{ALIE}} & \multicolumn{4}{c|}{\textbf{FOE}} & \multicolumn{4}{c|}{\textbf{LF}} & \multicolumn{4}{c|}{\textbf{Mimic}} & \multicolumn{4}{c|}{\textbf{SF}} & \multicolumn{4}{c|}{\textbf{Average of Attacks}} & \multicolumn{4}{c}{\textbf{Worst Case}}\\
\midrule

\multirow{16}{*}[.035cm]{CWMed} & - &
\textbf{\color{silver}93.7\newline(0.3)} & 93.2\newline(0.4) & 92.4\newline(0.3) & 84.8\newline(3.2) & 
93.3\newline(0.3) & 92.5\newline(0.5) & 91.4\newline(0.6) & 85.5\newline(2.2) & 
\textbf{\color{bronze}93.6\newline(0.3)} & 92.8\newline(0.3) & 88.9\newline(1.3) & 52.7\newline(11.5) & 
\textbf{\color{bronze}93.6\newline(0.3)} & 93.1\newline(0.5) & \textbf{\color{bronze}92.6\newline(0.4)} & 91.2\newline(0.8) & 
93.2\newline(0.3) & 92.5\newline(0.4) & 91.6\newline(0.8) & 80.5\newline(4.4) & 
\textbf{\color{bronze}93.5\newline(0.3)} & 92.8\newline(0.4) & 91.4\newline(0.7) & 78.9\newline(4.4) & 
93.2\newline(0.3) & 92.5\newline(0.5) & 88.9\newline(1.3) & 52.7\newline(11.5) \\
\end{tabular}
\end{table*}
\setlength{\tabcolsep}{6pt}
\renewcommand{\arraystretch}{1}

\renewcommand{\arraystretch}{1.2}
\setlength{\tabcolsep}{1pt}
\begin{table*}
\centering
\tiny
\caption{Accuracy (s.d.) \% on MNIST with $\alpha=1$ and $f=2,4,6,8$. Colors rank values within sub-columns: yellow (1st), blue (2nd), and red (3rd).}
\label{tab:AccuracyMNIST1}
\begin{tabular}{l|l|p{.48cm}p{.48cm}p{.48cm}p{.48cm}|p{.48cm}p{.48cm}p{.48cm}p{.48cm}|p{.48cm}p{.48cm}p{.48cm}p{.48cm}|p{.48cm}p{.48cm}p{.48cm}p{.48cm}|p{.48cm}p{.48cm}p{.48cm}p{.48cm}|p{.48cm}p{.48cm}p{.48cm}p{.48cm}|p{.48cm}p{.48cm}p{.48cm}p{.48cm}}
\midrule[1pt]
\textbf{Agg.} & \textbf{Other Strategies} & \multicolumn{4}{c|}{\textbf{ALIE}} & \multicolumn{4}{c|}{\textbf{FOE}} & \multicolumn{4}{c|}{\textbf{LF}} & \multicolumn{4}{c|}{\textbf{Mimic}} & \multicolumn{4}{c|}{\textbf{SF}} & \multicolumn{4}{c|}{\textbf{Average of Attacks}} & \multicolumn{4}{c}{\textbf{Worst Case}}\\
\midrule

\multirow{16}{*}[.035cm]{CWMed} & - &
\textbf{\color{bronze}93.9\newline(0.3)} & 93.1\newline(0.3) & 90.6\newline(1.4) & 55.4\newline(19.1) & 
93.2\newline(0.4) & 91.7\newline(0.5) & 86.7\newline(3.0) & 62.4\newline(10.7) & 
93.2\newline(0.3) & 91.0\newline(0.7) & 68.4\newline(13.6) & 21.5\newline(9.5) & 
93.2\newline(0.5) & 92.6\newline(0.7) & 89.8\newline(1.7) & 79.9\newline(7.2) & 
92.6\newline(0.5) & 90.9\newline(0.8) & 86.1\newline(4.2) & 35.1\newline(14.8) & 
93.2\newline(0.4) & 91.8\newline(0.6) & 84.3\newline(4.8) & 50.9\newline(12.3) & 
92.6\newline(0.5) & 90.9\newline(0.8) & 68.4\newline(13.6) & 21.5\newline(9.5) \\

 & BKT &
\textbf{\color{bronze}93.9\newline(0.4)} & \textbf{\color{silver}93.6\newline(0.3)} & 90.6\newline(1.4) & 55.4\newline(19.1) & 
\textbf{\color{silver}93.8\newline(0.4)} & \textbf{\color{bronze}93.3\newline(0.3)} & 86.7\newline(3.0) & 62.4\newline(10.7) & 
93.2\newline(0.5) & 91.5\newline(0.9) & 68.4\newline(13.6) & 21.5\newline(9.5) & 
93.8\newline(0.4) & 92.8\newline(0.5) & 89.8\newline(1.7) & 79.9\newline(7.2) & 
93.6\newline(0.5) & 92.6\newline(0.4) & 86.1\newline(4.2) & 35.1\newline(14.8) & 
93.7\newline(0.4) & 92.8\newline(0.5) & 84.3\newline(4.8) & 50.9\newline(12.3) & 
93.2\newline(0.5) & 91.5\newline(0.9) & 68.4\newline(13.6) & 21.5\newline(9.5) \\
\end{tabular}
\end{table*}
\setlength{\tabcolsep}{6pt}
\renewcommand{\arraystretch}{1}

\renewcommand{\arraystretch}{1.2}
\setlength{\tabcolsep}{1pt}
\begin{table*}
\centering
\tiny
\caption{Accuracy (s.d.) \% on MNIST with $\alpha=0.3$ and $f=2,4,6,8$. Colors rank values within sub-columns: yellow (1st), blue (2nd), and red (3rd).}
\label{tab:AccuracyMNIST0.3}
\begin{tabular}{l|l|p{.48cm}p{.48cm}p{.48cm}p{.48cm}|p{.48cm}p{.48cm}p{.48cm}p{.48cm}|p{.48cm}p{.48cm}p{.48cm}p{.48cm}|p{.48cm}p{.48cm}p{.48cm}p{.48cm}|p{.48cm}p{.48cm}p{.48cm}p{.48cm}|p{.48cm}p{.48cm}p{.48cm}p{.48cm}|p{.48cm}p{.48cm}p{.48cm}p{.48cm}}
\midrule[1pt]
\textbf{Agg.} & \textbf{Other Strategies} & \multicolumn{4}{c|}{\textbf{ALIE}} & \multicolumn{4}{c|}{\textbf{FOE}} & \multicolumn{4}{c|}{\textbf{LF}} & \multicolumn{4}{c|}{\textbf{Mimic}} & \multicolumn{4}{c|}{\textbf{SF}} & \multicolumn{4}{c|}{\textbf{Average of Attacks}} & \multicolumn{4}{c}{\textbf{Worst Case}}\\
\midrule

\multirow{16}{*}[.035cm]{CWMed} & - &
93.1\newline(0.8) & 91.1\newline(1.2) & 65.2\newline(18.0) & 15.0\newline(3.7) & 
90.0\newline(2.1) & 85.8\newline(6.3) & 70.4\newline(7.4) & 16.0\newline(3.7) & 
82.9\newline(9.9) & 68.9\newline(7.2) & 40.1\newline(9.9) & 8.4\newline(3.7) & 
89.3\newline(3.2) & 76.8\newline(12.6) & 55.5\newline(29.5) & 12.6\newline(2.7) & 
80.5\newline(9.9) & 60.5\newline(14.0) & 50.3\newline(9.7) & 16.2\newline(2.9) & 
87.2\newline(5.2) & 76.6\newline(8.3) & 56.3\newline(14.9) & 13.6\newline(3.3) & 
80.5\newline(9.9) & 60.5\newline(14.0) & 40.1\newline(9.9) & 8.4\newline(3.7) \\

 & BKT &
\textbf{\color{silver}93.8\newline(0.6)} & 93.0\newline(0.9) & 65.2\newline(18.0) & 15.0\newline(3.7) & 
\textbf{\color{bronze}93.6\newline(0.6)} & \textbf{\color{bronze}92.4\newline(1.2)} & 70.4\newline(7.4) & 16.0\newline(3.7) & 
92.8\newline(1.0) & 87.6\newline(2.8) & 40.1\newline(9.9) & 8.4\newline(3.7) & 
\textbf{\color{silver}93.6\newline(0.6)} & 90.7\newline(2.0) & 55.5\newline(29.5) & 12.6\newline(2.7) & 
93.4\newline(0.6) & 86.9\newline(4.1) & 50.3\newline(9.7) & 16.2\newline(2.9) & 
\textbf{\color{bronze}93.4\newline(0.7)} & 90.1\newline(2.2) & 56.3\newline(14.9) & 13.6\newline(3.3) & 
92.8\newline(1.0) & 86.9\newline(4.1) & 40.1\newline(9.9) & 8.4\newline(3.7) \\
\end{tabular}
\end{table*}
\setlength{\tabcolsep}{6pt}
\renewcommand{\arraystretch}{1}

\renewcommand{\arraystretch}{1.2}
\setlength{\tabcolsep}{1pt}
\begin{table*}
\centering
\tiny
\caption{Accuracy (s.d.) \% on Fashion MNIST with $\alpha=3$ and $f=2,4,6,8$. Colors rank values within sub-columns: yellow (1st), blue (2nd), and red (3rd).}
\label{tab:AccuracyFashion MNIST3}
\begin{tabular}{l|l|p{.48cm}p{.48cm}p{.48cm}p{.48cm}|p{.48cm}p{.48cm}p{.48cm}p{.48cm}|p{.48cm}p{.48cm}p{.48cm}p{.48cm}|p{.48cm}p{.48cm}p{.48cm}p{.48cm}|p{.48cm}p{.48cm}p{.48cm}p{.48cm}|p{.48cm}p{.48cm}p{.48cm}p{.48cm}|p{.48cm}p{.48cm}p{.48cm}p{.48cm}}
\midrule[1pt]
\textbf{Agg.} & \textbf{Other Strategies} & \multicolumn{4}{c|}{\textbf{ALIE}} & \multicolumn{4}{c|}{\textbf{FOE}} & \multicolumn{4}{c|}{\textbf{LF}} & \multicolumn{4}{c|}{\textbf{Mimic}} & \multicolumn{4}{c|}{\textbf{SF}} & \multicolumn{4}{c|}{\textbf{Average of Attacks}} & \multicolumn{4}{c}{\textbf{Worst Case}}\\
\midrule

\multirow{16}{*}[.035cm]{CWMed} & - &
78.0\newline(0.8) & 75.1\newline(1.2) & 70.5\newline(2.2) & 61.4\newline(4.7) & 
74.2\newline(0.9) & 70.6\newline(1.4) & 65.3\newline(2.4) & 55.3\newline(6.8) & 
75.5\newline(0.9) & 71.3\newline(1.6) & 61.3\newline(3.4) & 32.2\newline(12.3) & 
75.9\newline(0.9) & 73.8\newline(1.9) & 71.2\newline(2.1) & 70.7\newline(2.8) & 
74.1\newline(1.1) & 71.2\newline(2.1) & 65.2\newline(2.8) & 45.2\newline(9.9) & 
75.5\newline(0.9) & 72.4\newline(1.6) & 66.7\newline(2.6) & 53.0\newline(7.3) & 
74.1\newline(1.1) & 70.6\newline(1.4) & 61.3\newline(3.4) & 32.2\newline(12.3) \\

 & BKT &
77.7\newline(0.7) & 76.6\newline(1.0) & 70.5\newline(2.2) & 61.4\newline(4.7) & 
77.0\newline(0.6) & 75.5\newline(0.8) & 65.3\newline(2.4) & 55.3\newline(6.8) & 
\textbf{\color{bronze}76.0\newline(1.0)} & 72.4\newline(1.7) & 61.3\newline(3.4) & 32.2\newline(12.3) & 
76.9\newline(0.7) & \textbf{\color{bronze}75.9\newline(0.8)} & 71.2\newline(2.1) & 70.7\newline(2.8) & 
76.5\newline(0.7) & 72.8\newline(0.9) & 65.2\newline(2.8) & 45.2\newline(9.9) & 
76.8\newline(0.7) & 74.7\newline(1.0) & 66.7\newline(2.6) & 53.0\newline(7.3) & 
76.0\newline(1.0) & 72.4\newline(1.7) & 61.3\newline(3.4) & 32.2\newline(12.3) \\
\end{tabular}
\end{table*}
\setlength{\tabcolsep}{6pt}
\renewcommand{\arraystretch}{1}

\renewcommand{\arraystretch}{1.2}
\setlength{\tabcolsep}{1pt}
\begin{table*}
\centering
\tiny
\caption{Accuracy (s.d.) \% on Fashion MNIST with $\alpha=1$ and $f=2,4,6,8$. Colors rank values within sub-columns: yellow (1st), blue (2nd), and red (3rd).}
\label{tab:AccuracyFashion MNIST1}
\begin{tabular}{l|l|p{.48cm}p{.48cm}p{.48cm}p{.48cm}|p{.48cm}p{.48cm}p{.48cm}p{.48cm}|p{.48cm}p{.48cm}p{.48cm}p{.48cm}|p{.48cm}p{.48cm}p{.48cm}p{.48cm}|p{.48cm}p{.48cm}p{.48cm}p{.48cm}|p{.48cm}p{.48cm}p{.48cm}p{.48cm}|p{.48cm}p{.48cm}p{.48cm}p{.48cm}}
\midrule[1pt]
\textbf{Agg.} & \textbf{Other Strategies} & \multicolumn{4}{c|}{\textbf{ALIE}} & \multicolumn{4}{c|}{\textbf{FOE}} & \multicolumn{4}{c|}{\textbf{LF}} & \multicolumn{4}{c|}{\textbf{Mimic}} & \multicolumn{4}{c|}{\textbf{SF}} & \multicolumn{4}{c|}{\textbf{Average of Attacks}} & \multicolumn{4}{c}{\textbf{Worst Case}}\\
\midrule

\multirow{16}{*}[.035cm]{CWMed} & - &
77.5\newline(1.1) & 73.8\newline(1.7) & 63.0\newline(6.3) & 42.8\newline(10.6) & 
72.4\newline(2.5) & 67.5\newline(1.6) & 55.5\newline(7.9) & 29.1\newline(3.4) & 
73.0\newline(2.3) & 60.2\newline(3.9) & 41.3\newline(12.9) & 15.0\newline(6.6) & 
73.1\newline(3.3) & 70.9\newline(3.1) & 64.8\newline(6.8) & 63.8\newline(9.0) & 
70.2\newline(3.4) & 63.4\newline(4.4) & 53.9\newline(8.0) & 25.8\newline(4.3) & 
73.2\newline(2.5) & 67.2\newline(3.0) & 55.7\newline(8.4) & 35.3\newline(6.8) & 
70.2\newline(3.4) & 60.2\newline(3.9) & 41.3\newline(12.9) & 15.0\newline(6.6) \\

 & BKT &
78.0\newline(1.7) & 76.1\newline(1.8) & 63.0\newline(6.3) & 42.8\newline(10.6) & 
77.2\newline(1.5) & 74.3\newline(2.0) & 55.5\newline(7.9) & 29.1\newline(3.4) & 
76.4\newline(1.9) & 72.0\newline(2.7) & 41.3\newline(12.9) & 15.0\newline(6.6) & 
77.0\newline(1.8) & 73.9\newline(2.9) & 64.8\newline(6.8) & 63.8\newline(9.0) & 
76.2\newline(2.1) & 70.7\newline(3.1) & 53.9\newline(8.0) & 25.8\newline(4.3) & 
77.0\newline(1.8) & 73.4\newline(2.5) & 55.7\newline(8.4) & 35.3\newline(6.8) & 
76.2\newline(2.1) & 70.7\newline(3.1) & 41.3\newline(12.9) & 15.0\newline(6.6) \\
\end{tabular}
\end{table*}
\setlength{\tabcolsep}{6pt}
\renewcommand{\arraystretch}{1}

\renewcommand{\arraystretch}{1.2}
\setlength{\tabcolsep}{1pt}
\begin{table*}
\centering
\tiny
\caption{Accuracy (s.d.) \% on Fashion MNIST with $\alpha=0.3$ and $f=2,4,6,8$. Colors rank values within sub-columns: yellow (1st), blue (2nd), and red (3rd).}
\label{tab:AccuracyFashion MNIST0.3}
\begin{tabular}{l|l|p{.48cm}p{.48cm}p{.48cm}p{.48cm}|p{.48cm}p{.48cm}p{.48cm}p{.48cm}|p{.48cm}p{.48cm}p{.48cm}p{.48cm}|p{.48cm}p{.48cm}p{.48cm}p{.48cm}|p{.48cm}p{.48cm}p{.48cm}p{.48cm}|p{.48cm}p{.48cm}p{.48cm}p{.48cm}|p{.48cm}p{.48cm}p{.48cm}p{.48cm}}
\midrule[1pt]
\textbf{Agg.} & \textbf{Other Strategies} & \multicolumn{4}{c|}{\textbf{ALIE}} & \multicolumn{4}{c|}{\textbf{FOE}} & \multicolumn{4}{c|}{\textbf{LF}} & \multicolumn{4}{c|}{\textbf{Mimic}} & \multicolumn{4}{c|}{\textbf{SF}} & \multicolumn{4}{c|}{\textbf{Average of Attacks}} & \multicolumn{4}{c}{\textbf{Worst Case}}\\
\midrule

\multirow{16}{*}[.035cm]{CWMed} & - &
72.6\newline(2.5) & 65.3\newline(3.8) & 39.6\newline(7.2) & 19.5\newline(6.9) & 
65.1\newline(4.3) & 54.5\newline(3.5) & 39.4\newline(7.1) & 16.0\newline(5.8) & 
58.2\newline(5.3) & 41.6\newline(5.1) & 23.6\newline(8.0) & 6.0\newline(5.2) & 
65.3\newline(5.0) & 47.2\newline(13.2) & 50.0\newline(6.3) & 33.1\newline(19.3) & 
50.3\newline(4.5) & 38.2\newline(5.1) & 29.0\newline(4.1) & 16.3\newline(1.7) & 
62.3\newline(4.3) & 49.4\newline(6.1) & 36.3\newline(6.5) & 18.2\newline(7.8) & 
50.3\newline(4.5) & 38.2\newline(5.1) & 23.6\newline(8.0) & 6.0\newline(5.2) \\

 & BKT &
76.1\newline(2.9) & 69.8\newline(3.8) & 39.6\newline(7.2) & 19.5\newline(6.9) & 
75.0\newline(3.0) & 68.8\newline(5.0) & 39.4\newline(7.1) & 16.0\newline(5.8) & 
74.6\newline(2.7) & 63.7\newline(5.1) & 23.6\newline(8.0) & 6.0\newline(5.2) & 
76.2\newline(2.1) & 68.3\newline(4.9) & 50.0\newline(6.3) & 33.1\newline(19.3) & 
73.4\newline(3.8) & 55.2\newline(8.4) & 29.0\newline(4.1) & 16.3\newline(1.7) & 
75.1\newline(2.9) & 65.1\newline(5.4) & 36.3\newline(6.5) & 18.2\newline(7.8) & 
73.4\newline(3.8) & 55.2\newline(8.4) & 23.6\newline(8.0) & 6.0\newline(5.2) \\
\end{tabular}
\end{table*}
\setlength{\tabcolsep}{6pt}
\renewcommand{\arraystretch}{1}

\renewcommand{\arraystretch}{1.2}
\setlength{\tabcolsep}{1pt}
\begin{table*}
\centering
\tiny
\caption{Accuracy (s.d.) \% on Purchase100 with $\alpha=3$ and $f=2,4,6,8$. Colors rank values within sub-columns: yellow (1st), blue (2nd), and red (3rd).}
\label{tab:AccuracyPurchase1003}
\begin{tabular}{l|l|p{.48cm}p{.48cm}p{.48cm}p{.48cm}|p{.48cm}p{.48cm}p{.48cm}p{.48cm}|p{.48cm}p{.48cm}p{.48cm}p{.48cm}|p{.48cm}p{.48cm}p{.48cm}p{.48cm}|p{.48cm}p{.48cm}p{.48cm}p{.48cm}|p{.48cm}p{.48cm}p{.48cm}p{.48cm}|p{.48cm}p{.48cm}p{.48cm}p{.48cm}}
\midrule[1pt]
\textbf{Agg.} & \textbf{Other Strategies} & \multicolumn{4}{c|}{\textbf{ALIE}} & \multicolumn{4}{c|}{\textbf{FOE}} & \multicolumn{4}{c|}{\textbf{LF}} & \multicolumn{4}{c|}{\textbf{Mimic}} & \multicolumn{4}{c|}{\textbf{SF}} & \multicolumn{4}{c|}{\textbf{Average of Attacks}} & \multicolumn{4}{c}{\textbf{Worst Case}}\\
\midrule

\multirow{16}{*}[.035cm]{CWMed} & - &
\textbf{\color{bronze}72.2\newline(0.6)} & 69.3\newline(1.2) & 64.2\newline(1.6) & 40.2\newline(3.7) & 
67.2\newline(0.7) & 60.7\newline(0.6) & 50.2\newline(1.0) & 21.7\newline(0.6) & 
70.3\newline(0.6) & 68.0\newline(0.8) & 61.2\newline(0.8) & 29.9\newline(2.2) & 
68.5\newline(1.4) & 61.6\newline(2.6) & 50.4\newline(4.6) & 34.9\newline(4.9) & 
66.8\newline(0.7) & 60.6\newline(0.8) & 49.8\newline(1.3) & 16.8\newline(4.4) & 
69.0\newline(0.8) & 64.0\newline(1.2) & 55.1\newline(1.8) & 28.7\newline(3.2) & 
66.8\newline(0.7) & 60.6\newline(0.8) & 49.8\newline(1.3) & 16.8\newline(4.4) \\

 & BKT &
\textbf{\color{silver}72.4\newline(0.6)} & \textbf{\color{silver}71.7\newline(0.8)} & 64.2\newline(1.6) & 40.2\newline(3.7) & 
\textbf{\color{bronze}70.5\newline(0.6)} & 67.0\newline(0.6) & 50.2\newline(1.0) & 21.7\newline(0.6) & 
69.4\newline(0.7) & 66.4\newline(1.0) & 61.2\newline(0.8) & 29.9\newline(2.2) & 
70.5\newline(1.3) & 65.0\newline(2.2) & 50.4\newline(4.6) & 34.9\newline(4.9) & 
69.1\newline(0.7) & 62.9\newline(0.7) & 49.8\newline(1.3) & 16.8\newline(4.4) & 
70.4\newline(0.8) & 66.6\newline(1.1) & 55.1\newline(1.8) & 28.7\newline(3.2) & 
69.1\newline(0.7) & 62.9\newline(0.7) & 49.8\newline(1.3) & 16.8\newline(4.4) \\
\end{tabular}
\end{table*}
\setlength{\tabcolsep}{6pt}
\renewcommand{\arraystretch}{1}

\renewcommand{\arraystretch}{1.2}
\setlength{\tabcolsep}{1pt}
\begin{table*}
\centering
\tiny
\caption{Accuracy (s.d.) \% on Purchase100 with $\alpha=1$ and $f=2,4,6,8$. Colors rank values within sub-columns: yellow (1st), blue (2nd), and red (3rd).}
\label{tab:AccuracyPurchase1001}
\begin{tabular}{l|l|p{.48cm}p{.48cm}p{.48cm}p{.48cm}|p{.48cm}p{.48cm}p{.48cm}p{.48cm}|p{.48cm}p{.48cm}p{.48cm}p{.48cm}|p{.48cm}p{.48cm}p{.48cm}p{.48cm}|p{.48cm}p{.48cm}p{.48cm}p{.48cm}|p{.48cm}p{.48cm}p{.48cm}p{.48cm}|p{.48cm}p{.48cm}p{.48cm}p{.48cm}}
\midrule[1pt]
\textbf{Agg.} & \textbf{Other Strategies} & \multicolumn{4}{c|}{\textbf{ALIE}} & \multicolumn{4}{c|}{\textbf{FOE}} & \multicolumn{4}{c|}{\textbf{LF}} & \multicolumn{4}{c|}{\textbf{Mimic}} & \multicolumn{4}{c|}{\textbf{SF}} & \multicolumn{4}{c|}{\textbf{Average of Attacks}} & \multicolumn{4}{c}{\textbf{Worst Case}}\\
\midrule

\multirow{16}{*}[.035cm]{CWMed} & - &
69.9\newline(0.8) & 63.2\newline(2.1) & 50.0\newline(3.4) & 5.9\newline(1.3) & 
61.6\newline(0.9) & 49.0\newline(0.9) & 29.3\newline(2.1) & 12.4\newline(3.0) & 
66.6\newline(0.8) & 61.3\newline(0.9) & 41.4\newline(1.2) & 11.1\newline(1.9) & 
61.6\newline(2.8) & 50.3\newline(4.5) & 29.6\newline(4.3) & 12.4\newline(4.8) & 
61.1\newline(1.0) & 49.3\newline(1.0) & 27.9\newline(1.2) & 1.5\newline(0.6) & 
64.2\newline(1.3) & 54.6\newline(1.9) & 35.6\newline(2.4) & 8.7\newline(2.3) & 
61.1\newline(1.0) & 49.0\newline(0.9) & 27.9\newline(1.2) & 1.5\newline(0.6) \\

 & BKT &
\textbf{\color{silver}71.5\newline(0.8)} & 68.9\newline(1.8) & 50.0\newline(3.4) & 5.9\newline(1.3) & 
\textbf{\color{silver}69.2\newline(1.0)} & 63.9\newline(0.9) & 29.3\newline(2.1) & 12.4\newline(3.0) & 
68.0\newline(0.9) & 62.7\newline(0.9) & 41.4\newline(1.2) & 11.1\newline(1.9) & 
65.8\newline(1.5) & 53.7\newline(3.1) & 29.6\newline(4.3) & 12.4\newline(4.8) & 
67.1\newline(1.0) & 57.3\newline(0.8) & 27.9\newline(1.2) & 1.5\newline(0.6) & 
68.3\newline(1.0) & 61.3\newline(1.5) & 35.6\newline(2.4) & 8.7\newline(2.3) & 
65.8\newline(1.5) & 53.7\newline(3.1) & 27.9\newline(1.2) & 1.5\newline(0.6) \\
\end{tabular}
\end{table*}
\setlength{\tabcolsep}{6pt}
\renewcommand{\arraystretch}{1}

\renewcommand{\arraystretch}{1.2}
\setlength{\tabcolsep}{1pt}
\begin{table*}
\centering
\tiny
\caption{Accuracy (s.d.) \% on Purchase100 with $\alpha=0.3$ and $f=2,4,6,8$. Colors rank values within sub-columns: yellow (1st), blue (2nd), and red (3rd).}
\label{tab:AccuracyPurchase1000.3}
\begin{tabular}{l|l|p{.48cm}p{.48cm}p{.48cm}p{.48cm}|p{.48cm}p{.48cm}p{.48cm}p{.48cm}|p{.48cm}p{.48cm}p{.48cm}p{.48cm}|p{.48cm}p{.48cm}p{.48cm}p{.48cm}|p{.48cm}p{.48cm}p{.48cm}p{.48cm}|p{.48cm}p{.48cm}p{.48cm}p{.48cm}|p{.48cm}p{.48cm}p{.48cm}p{.48cm}}
\midrule[1pt]
\textbf{Agg.} & \textbf{Other Strategies} & \multicolumn{4}{c|}{\textbf{ALIE}} & \multicolumn{4}{c|}{\textbf{FOE}} & \multicolumn{4}{c|}{\textbf{LF}} & \multicolumn{4}{c|}{\textbf{Mimic}} & \multicolumn{4}{c|}{\textbf{SF}} & \multicolumn{4}{c|}{\textbf{Average of Attacks}} & \multicolumn{4}{c}{\textbf{Worst Case}}\\
\midrule

\multirow{16}{*}[.035cm]{CWMed} & - &
61.5\newline(1.4) & 42.1\newline(2.5) & 8.4\newline(2.8) & 2.5\newline(1.0) & 
47.8\newline(1.6) & 30.9\newline(3.0) & 14.8\newline(1.4) & \textbf{\color{bronze}3.4\newline(1.9)} & 
54.0\newline(2.0) & 38.4\newline(1.6) & 11.5\newline(2.2) & 2.1\newline(1.2) & 
48.4\newline(4.0) & 30.3\newline(9.5) & 5.1\newline(2.4) & 2.1\newline(1.0) & 
44.1\newline(1.4) & 21.8\newline(1.8) & 1.8\newline(0.7) & 1.3\newline(0.5) & 
51.2\newline(2.1) & 32.7\newline(3.7) & 8.3\newline(1.9) & 2.3\newline(1.1) & 
44.1\newline(1.4) & 21.8\newline(1.8) & 1.8\newline(0.7) & 1.3\newline(0.5) \\

 & BKT &
\textbf{\color{silver}69.2\newline(1.2)} & 62.2\newline(3.5) & 8.4\newline(2.8) & 2.5\newline(1.0) & 
\textbf{\color{silver}65.4\newline(1.4)} & 55.0\newline(2.2) & 14.8\newline(1.4) & \textbf{\color{bronze}3.4\newline(1.9)} & 
64.0\newline(2.1) & 52.3\newline(2.7) & 11.5\newline(2.2) & 2.1\newline(1.2) & 
57.9\newline(3.0) & 35.2\newline(8.8) & 5.1\newline(2.4) & 2.1\newline(1.0) & 
\textbf{\color{silver}62.2\newline(1.5)} & 40.9\newline(2.5) & 1.8\newline(0.7) & 1.3\newline(0.5) & 
\textbf{\color{bronze}63.7\newline(1.9)} & 49.1\newline(3.9) & 8.3\newline(1.9) & 2.3\newline(1.1) & 
57.9\newline(3.0) & 35.2\newline(8.8) & 1.8\newline(0.7) & 1.3\newline(0.5) \\
\end{tabular}
\end{table*}
\setlength{\tabcolsep}{6pt}
\renewcommand{\arraystretch}{1}

\renewcommand{\arraystretch}{1.2}
\setlength{\tabcolsep}{1pt}
\begin{table*}
\centering
\tiny
\caption{Accuracy (s.d.) \% on CIFAR10 with $\alpha=3$ and $f=2,4,6,8$. Colors rank values within sub-columns: yellow (1st), blue (2nd), and red (3rd).}
\label{tab:AccuracyCIFAR103}
\begin{tabular}{l|l|p{.48cm}p{.48cm}p{.48cm}p{.48cm}|p{.48cm}p{.48cm}p{.48cm}p{.48cm}|p{.48cm}p{.48cm}p{.48cm}p{.48cm}|p{.48cm}p{.48cm}p{.48cm}p{.48cm}|p{.48cm}p{.48cm}p{.48cm}p{.48cm}|p{.48cm}p{.48cm}p{.48cm}p{.48cm}|p{.48cm}p{.48cm}p{.48cm}p{.48cm}}
\midrule[1pt]
\textbf{Agg.} & \textbf{Other Strategies} & \multicolumn{4}{c|}{\textbf{ALIE}} & \multicolumn{4}{c|}{\textbf{FOE}} & \multicolumn{4}{c|}{\textbf{LF}} & \multicolumn{4}{c|}{\textbf{Mimic}} & \multicolumn{4}{c|}{\textbf{SF}} & \multicolumn{4}{c|}{\textbf{Average of Attacks}} & \multicolumn{4}{c}{\textbf{Worst Case}}\\
\midrule

\multirow{16}{*}[.035cm]{CWMed} & - &
69.6\newline(0.7) & 55.9\newline(1.0) & 35.8\newline(1.6) & 19.7\newline(2.8) & 
68.0\newline(0.8) & 59.6\newline(1.5) & 51.0\newline(1.5) & 28.8\newline(4.3) & 
66.7\newline(1.0) & 56.3\newline(1.6) & 32.9\newline(2.8) & 18.9\newline(2.4) & 
62.6\newline(2.6) & 45.0\newline(7.2) & 18.6\newline(4.2) & 18.5\newline(5.7) & 
63.3\newline(1.1) & 51.9\newline(3.6) & 35.9\newline(3.1) & 17.8\newline(2.8) & 
66.0\newline(1.2) & 53.7\newline(3.0) & 34.8\newline(2.6) & 20.7\newline(3.6) & 
62.6\newline(2.6) & 45.0\newline(7.2) & 18.6\newline(4.2) & 17.8\newline(2.8) \\

 & BKT &
70.6\newline(0.6) & 64.7\newline(1.0) & 35.8\newline(1.6) & 19.7\newline(2.8) & 
69.8\newline(0.7) & 68.4\newline(0.7) & 51.0\newline(1.5) & 28.8\newline(4.3) & 
65.5\newline(0.9) & 58.6\newline(2.5) & 32.9\newline(2.8) & 18.9\newline(2.4) & 
66.2\newline(1.6) & 57.7\newline(3.9) & 18.6\newline(4.2) & 18.5\newline(5.7) & 
69.0\newline(0.7) & 63.6\newline(1.0) & 35.9\newline(3.1) & 17.8\newline(2.8) & 
68.2\newline(0.9) & 62.6\newline(1.8) & 34.8\newline(2.6) & 20.7\newline(3.6) & 
65.5\newline(0.9) & 57.7\newline(3.9) & 18.6\newline(4.2) & 17.8\newline(2.8) \\
\end{tabular}
\end{table*}
\setlength{\tabcolsep}{6pt}
\renewcommand{\arraystretch}{1}

\renewcommand{\arraystretch}{1.2}
\setlength{\tabcolsep}{1pt}
\begin{table*}
\centering
\tiny
\caption{Accuracy (s.d.) \% on CIFAR10 with $\alpha=1$ and $f=2,4,6,8$. Colors rank values within sub-columns: yellow (1st), blue (2nd), and red (3rd).}
\label{tab:AccuracyCIFAR101}
\begin{tabular}{l|l|p{.48cm}p{.48cm}p{.48cm}p{.48cm}|p{.48cm}p{.48cm}p{.48cm}p{.48cm}|p{.48cm}p{.48cm}p{.48cm}p{.48cm}|p{.48cm}p{.48cm}p{.48cm}p{.48cm}|p{.48cm}p{.48cm}p{.48cm}p{.48cm}|p{.48cm}p{.48cm}p{.48cm}p{.48cm}|p{.48cm}p{.48cm}p{.48cm}p{.48cm}}
\midrule[1pt]
\textbf{Agg.} & \textbf{Other Strategies} & \multicolumn{4}{c|}{\textbf{ALIE}} & \multicolumn{4}{c|}{\textbf{FOE}} & \multicolumn{4}{c|}{\textbf{LF}} & \multicolumn{4}{c|}{\textbf{Mimic}} & \multicolumn{4}{c|}{\textbf{SF}} & \multicolumn{4}{c|}{\textbf{Average of Attacks}} & \multicolumn{4}{c}{\textbf{Worst Case}}\\
\midrule

\multirow{16}{*}[.035cm]{CWMed} & - &
63.2\newline(0.8) & 46.6\newline(4.1) & 25.8\newline(4.0) & 12.9\newline(3.2) & 
60.8\newline(1.0) & 51.1\newline(0.7) & 37.4\newline(3.5) & 18.2\newline(5.3) & 
53.9\newline(1.0) & 33.9\newline(4.8) & 22.1\newline(3.0) & 15.9\newline(2.0) & 
46.9\newline(5.1) & 15.4\newline(2.6) & 16.4\newline(4.3) & 15.2\newline(2.7) & 
50.0\newline(2.3) & 34.7\newline(4.3) & 23.3\newline(2.9) & 13.9\newline(3.3) & 
55.0\newline(2.0) & 36.3\newline(3.3) & 25.0\newline(3.5) & 15.2\newline(3.3) & 
46.9\newline(5.1) & 15.4\newline(2.6) & 16.4\newline(4.3) & 12.9\newline(3.2) \\

 & BKT &
65.5\newline(1.3) & 55.6\newline(2.5) & 25.8\newline(4.0) & 12.9\newline(3.2) & 
64.7\newline(1.5) & \textbf{\color{bronze}62.3\newline(1.7)} & 37.4\newline(3.5) & 18.2\newline(5.3) & 
60.9\newline(1.5) & 52.2\newline(2.5) & 22.1\newline(3.0) & 15.9\newline(2.0) & 
58.7\newline(2.9) & \textbf{\color{bronze}37.1\newline(6.8)} & 16.4\newline(4.3) & 15.2\newline(2.7) & 
63.0\newline(1.4) & 53.4\newline(2.5) & 23.3\newline(2.9) & 13.9\newline(3.3) & 
62.6\newline(1.7) & 52.1\newline(3.2) & 25.0\newline(3.5) & 15.2\newline(3.3) & 
58.7\newline(2.9) & 37.1\newline(6.8) & 16.4\newline(4.3) & 12.9\newline(3.2) \\
\end{tabular}
\end{table*}
\setlength{\tabcolsep}{6pt}
\renewcommand{\arraystretch}{1}

\renewcommand{\arraystretch}{1.2}
\setlength{\tabcolsep}{1pt}
\begin{table*}
\centering
\tiny
\caption{Accuracy (s.d.) \% on CIFAR10 with $\alpha=0.3$ and $f=2,4,6,8$. Colors rank values within sub-columns: yellow (1st), blue (2nd), and red (3rd).}
\label{tab:AccuracyCIFAR100.3}
\begin{tabular}{l|l|p{.48cm}p{.48cm}p{.48cm}p{.48cm}|p{.48cm}p{.48cm}p{.48cm}p{.48cm}|p{.48cm}p{.48cm}p{.48cm}p{.48cm}|p{.48cm}p{.48cm}p{.48cm}p{.48cm}|p{.48cm}p{.48cm}p{.48cm}p{.48cm}|p{.48cm}p{.48cm}p{.48cm}p{.48cm}|p{.48cm}p{.48cm}p{.48cm}p{.48cm}}
\midrule[1pt]
\textbf{Agg.} & \textbf{Other Strategies} & \multicolumn{4}{c|}{\textbf{ALIE}} & \multicolumn{4}{c|}{\textbf{FOE}} & \multicolumn{4}{c|}{\textbf{LF}} & \multicolumn{4}{c|}{\textbf{Mimic}} & \multicolumn{4}{c|}{\textbf{SF}} & \multicolumn{4}{c|}{\textbf{Average of Attacks}} & \multicolumn{4}{c}{\textbf{Worst Case}}\\
\midrule

\multirow{16}{*}[.035cm]{CWMed} & - &
44.0\newline(8.1) & 32.9\newline(2.2) & 15.8\newline(4.4) & 10.9\newline(3.2) & 
43.5\newline(7.2) & 40.1\newline(3.0) & 21.9\newline(2.2) & 11.7\newline(3.4) & 
25.4\newline(3.5) & 21.8\newline(4.5) & 22.7\newline(4.6) & 11.7\newline(1.6) & 
15.7\newline(2.9) & 12.3\newline(2.2) & 16.0\newline(3.8) & 13.6\newline(3.4) & 
29.3\newline(4.9) & 20.8\newline(4.3) & 16.4\newline(2.4) & 11.2\newline(3.4) & 
31.6\newline(5.3) & 25.6\newline(3.2) & 18.6\newline(3.5) & 11.8\newline(3.0) & 
15.7\newline(2.9) & 12.3\newline(2.2) & 15.8\newline(4.4) & 10.9\newline(3.2) \\

 & BKT &
57.1\newline(1.8) & 37.2\newline(3.3) & 15.8\newline(4.4) & 10.9\newline(3.2) & 
56.4\newline(1.9) & 47.3\newline(3.8) & 21.9\newline(2.2) & 11.7\newline(3.4) & 
51.9\newline(2.7) & 35.9\newline(3.9) & 22.7\newline(4.6) & 11.7\newline(1.6) & 
45.5\newline(4.9) & 19.5\newline(4.7) & 16.0\newline(3.8) & 13.6\newline(3.4) & 
54.2\newline(1.7) & 28.8\newline(4.0) & 16.4\newline(2.4) & 11.2\newline(3.4) & 
53.0\newline(2.6) & 33.7\newline(3.9) & 18.6\newline(3.5) & 11.8\newline(3.0) & 
45.5\newline(4.9) & 19.5\newline(4.7) & 15.8\newline(4.4) & 10.9\newline(3.2) \\
\end{tabular}
\end{table*}
\setlength{\tabcolsep}{6pt}
\renewcommand{\arraystretch}{1}

\end{document}